\documentclass[acmlarge]{acmart}
\AtBeginDocument{%
  \providecommand\BibTeX{{%
    \normalfont B\kern-0.5em{\scshape i\kern-0.25em b}\kern-0.8em\TeX}}}

\usepackage{subcaption}
\usepackage[utf8]{inputenc} 
\usepackage{multirow}
\DeclareUnicodeCharacter{FB01}{fi} 
\usepackage[export]{adjustbox}
\usepackage[
separate-uncertainty = true,
multi-part-units = repeat
]{siunitx}
\graphicspath{{Figures/}} 

\usepackage[english]{babel}
\usepackage{array}
\usepackage{ragged2e}
\usepackage{amsmath}
\usepackage{pbox}
\usepackage{makecell}

\newcommand{\matr}[1]{\mathbf{#1}}

\copyrightyear{2021}
\acmYear{2021}
\acmDOI{xxxxx/1xxx.xxxxxx6}
\acmConference[Woodstock '18]{Woodstock '18: ACM Symposium on Neural
	Gaze Detection}{June 03--05, 2018}{Woodstock, NY}
\acmBooktitle{Woodstock '18: ACM Symposium on Neural Gaze Detection,
	June 03--05, 2018, Woodstock, NY}
\acmPrice{15.00}
\acmISBN{978-1-4503-XXXX-X/18/06}

\begin{document}

\title{Ubi-SleepNet: Advanced Multimodal Fusion Techniques for Three-stage Sleep Classification Using Ubiquitous Sensing}

\author{Bing Zhai}

\email{b.zhai2@newcastle.ac.uk}
\orcid{0000-0003-1635-1406}
\affiliation{%
	\institution{School of Computing, Newcastle University}
	\city{Newcastle upon Tyne}
	\country{UK}
}

\author{Yu Guan}
\email{Yu.Guan@newcastle.ac.uk}
\orcid{0000-0002-1283-3806}
\affiliation{%
	\institution{School of Computing, Newcastle University}
	\city{Newcastle upon Tyne}
	\country{UK}
}

\author{Michael Catt}
\email{Michael.Catt@newcastle.ac.uk}
\orcid{0000-0002-1283-3806}
\affiliation{%
	\institution{Population Health Sciences Institute, Newcastle University}
	\streetaddress{}
	\city{Newcastle upon Tyne}
	\country{UK}
}

\author{Thomas Plötz}
\email{thomas.ploetz@gatech.edu}
\orcid{}
\affiliation{%
	\institution{School of Interactive Computing, Georgia Institute of Technology}
	\streetaddress{}
	\city{Atlanta}
	\country{USA}
}

\renewcommand{\shortauthors}{Zhai et al.}

\begin{abstract}
Sleep is a fundamental physiological process that is essential for sustaining a healthy body and mind. The gold standard for clinical sleep monitoring is polysomnography(PSG), based on which sleep can be categorized into five stages, including wake/rapid eye movement sleep (REM sleep)/Non-REM sleep 1 (N1)/Non-REM sleep 2 (N2)/Non-REM sleep 3 (N3). However, PSG is expensive, burdensome and not suitable for daily use. For long-term sleep monitoring, ubiquitous sensing may be a solution. Most recently, cardiac and movement sensing has become popular in classifying three-stage sleep, since both modalities can be easily acquired from research-grade or consumer-grade devices (e.g., Apple Watch). However, how best to fuse the data for greatest accuracy remains an open question. In this work, we comprehensively studied deep learning (DL)-based advanced fusion techniques consisting of three fusion strategies alongside three fusion methods for three-stage sleep classification based on two publicly available datasets. Experimental results demonstrate important evidences that three-stage sleep can be reliably classified by fusing cardiac/movement sensing modalities, which may potentially become a practical tool to conduct large-scale sleep stage assessment studies or long-term self-tracking on sleep. To accelerate the progression of sleep research in the ubiquitous/wearable computing community, we made this project open source, and the code can be found at: \href{https://github.com/bzhai/Ubi-SleepNet}{https://github.com/bzhai/Ubi-SleepNet}.

\end{abstract}



\maketitle

\section{Introduction}
Human beings spend about one-third or more of of their lives sleeping.
It is a physiological process essential to maintain life and health~\cite{Tobaldini2013HeartSleep},  and it plays a major role in repairing the body tissues, removing toxic metabolic waste products from the brain, consolidating memory, restoring energy and enhancing immune defence~\cite{Perez-Pozuelo2020TheMedicine,Besedovsky2012SleepFunction,Schwartz2009NeurophysiologyImplications,Fultz2019CoupledSleep,BenSimon2020OveranxiousUnderslept}. Long-term lack of sleep and/or poor-quality sleep  is likely to increase the risk of obesity, diabetes, and heart and blood vessel (cardiovascular) disease~\cite{Perez-Pozuelo2020TheMedicine,Kohansieh2015SleepDisease}. Accurate and long-term sleep monitoring using ubiquitous sensing technology is increasingly vital to the understanding of human health and is becoming an active area of health research. 

The gold standard for clinical sleep monitoring is polysomnography (PSG), which requires subjects to wear multi-channel sensors on the body~\cite{Kapur2017ClinicalGuideline}. \begin{figure}[t]
    \centering
    \includegraphics[width=\linewidth]{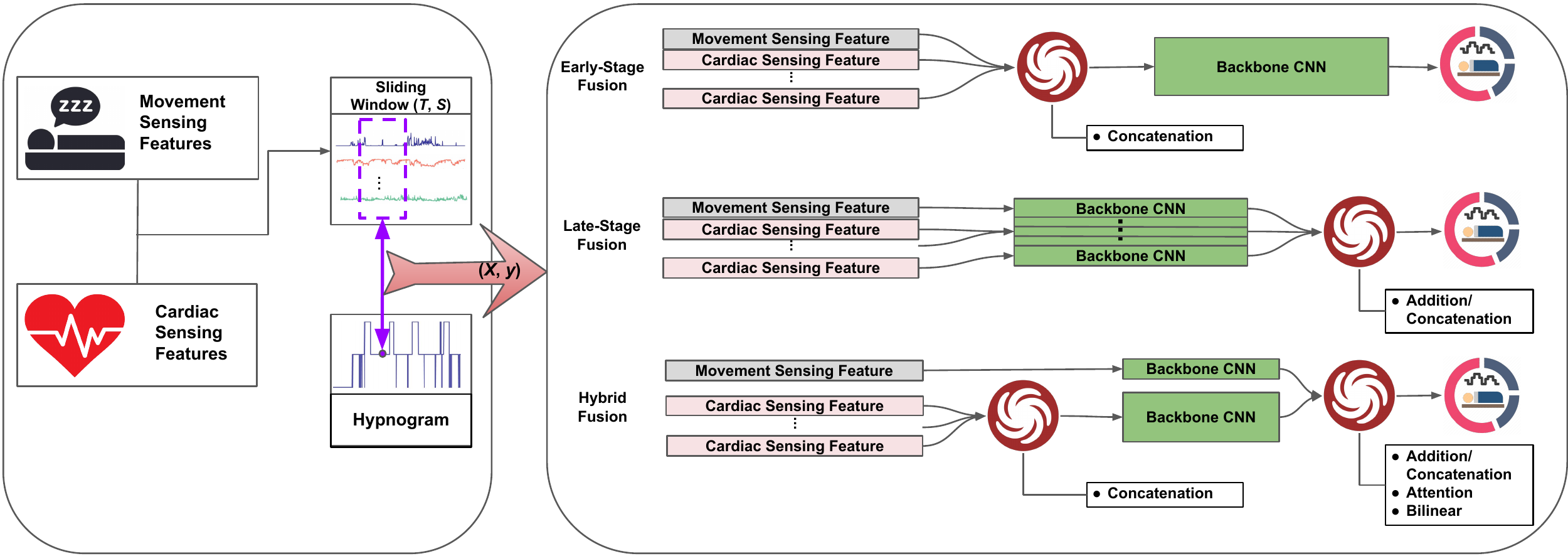}
	\caption{An overview of the three-stage sleep classification system. Features were extracted for each sleep epoch (30s). The sliding window method divides the sleep data into multiple segments with window length $T$ and stride $S$. In our study, we set $T=101$, and $S =1$. The hypnogram represents the stages of sleep over time in each sleep epoch. Three fusion strategies and three fusion methods were studied. The prediction process was performed for each sleep epoch.}
    \label{fig:system_overview}
\end{figure}
PSG recording can be classified into five stages, i.e., wake, rapid eye movement sleep (REM sleep) and three-types of non-rapid eye movement sleep (NREM sleep) including N1, N2 and N3. According to the American Academy of Sleep Medicine (AASM) rules~\cite{IBER2007TheRules}, each stage lasts 30 seconds (i.e., a sleep epoch). 
However, it is expensive and burdensome,  which is impractical for long-term sleep monitoring. 

For such monitoring, many ubiquitous sensing approaches were studied, including actigraphy~\cite{Ancoli-Israel2003TheRhythms}, smart 
watches~\cite{Walch2019SleepDevice} , WiFi~\cite{Yue2020BodyCompass:Signals}, bed sensors~\cite{Park2014BallistocardiographyEstimation} and radio signals~\cite{Hsu2017Zero-EffortSignals}, etc. Among them, in terms of reliability and usability, cardiac and movement (upper limb) sensing are considered promising modalities. They can be easily collected from lightweight research/consumer-grade devices (e.g., Apple Watch~\cite{Walch2019SleepDevice}). Based on cardiac and movement sensing, previous work studied a number of machine learning and DL approaches on sleep monitoring tasks from two-stage (wake/sleep) to five-stage (wake/REM/N1/N2/N3) sleep classification~\cite{Zhai2020MakingSensing} . 
Their initial results suggested the feasibility of using these two modalities for three-stage (wake/REM/NREM) sleep classification, and their findings corroborate sleep physiology studies~\cite{Chouchou2014HeartBrain, Montano2009HeartBehavior}, where NREM sleep was not deemed to be easily separated into N1/N2/N3 without employing EEG signals.  

The easy-to-collect nature of cardiac and movement sensing provided a salable method for large-scale and long-term sleep monitoring. Longitudinal sleep monitoring with accurate details in (three) sleep stages is meaningful to health and medical research. Deep NREM sleep (or slow wave sleep - SWS) is known to be the most “restorative” sleep stage, which controls hormonal changes that affect glucose regulation~\cite{E2008Slow-waveHumans}. Long-term reduction in NREM sleep may adversely affect glucose homeostasis and increase the risk of type 2 diabetes~\cite{S2018SleepDiabetes}. The REM sleep dysregulation has played a central role in depression and Parkinson’s studies~\cite{Agargun2003REMPatients, Kupfer1972INTERVALDEPRESSION}. For instance, reduced REM sleep latency, along with increased REM sleep duration and REM sleep density, have been considered to be an objective indicator of depressive disorder and inversely correlated to its severity~\cite{L2013REMArt, Wang2015TheDepression,Stefani2019SleepDisease}. The increased health research density in digital phenotypes by using inexpensive, mass-produced consumer wearables demand reliable algorithms that can classify sleep stages in longitudinal settings~\cite{Teo2019DigitalAging}. Beyond health and clinical monitoring, the sleep monitoring has also been welcomed by self-trackers in the past decades~\cite{Ravichandran2017MakingHealth}. 

To the best of our knowledge, only two previous studies~\cite{Walch2019SleepDevice, Zhai2020MakingSensing} have adopted cardiac and movement sensing for three-stage sleep classification based on publicly accessible sleep datasets. Both works used very basic multimodal fusion techniques (i.e., feature concatenation~\cite{Walch2019SleepDevice, Zhai2020MakingSensing}) in  neural networks, which tested the feasibility of classifying three-stage sleep. However, the models used in these benchmark studies did not achieve good performance, due to overestimating NREM sleep time and underestimating wake time. The simple fusion technique may not fully utilize the advantages of multimodal data, especially in heterogeneous multimodal data scenarios. For instance, movement sensing can achieve better classification performance in sleep/wake tasks compared with the use of cardiac sensing alone~\cite{Zhai2020MakingSensing}. But movement sensing alone is incapable of discerning three-stage sleep. Given that, it is desirable to explore the advanced fusion techniques to further boost the performance. 

In this work, we firstly systematically studied three fusion strategies for three-stage sleep classification, including early-stage fusion, late-stage fusion and hybrid fusion, to answer the question, \textit {"At what stage should the cardiac and movement sensing representation be merged?"}  

Secondly, we employed three fusion methods (simple operations, attention mechanism, tensor methods) to answer the question, \textit {"How to better combine cardiac and movement sensing representations?"}  The simple baseline operations (concatenation and addition) as well as the advanced fusion methods (the attention mechanism-based method~\cite{Yang2016StackedAnswering}  and bi-linear pooling-based method~\cite{Lu2016HierarchicalAnswering}) were studied. The pipeline of our system is demonstrated in Figure~\ref{fig:system_overview}.

These fusion techniques were comprehensively evaluated on two public datasets which are the Apple Watch dataset~\cite{Walch2019SleepDevice} and the Multi-Ethnic Study of Atherosclerosis (MESA) dataset~\cite{Zhang2018TheCommons,Zhai2020MakingSensing,Chen2015Racial/ethnicMESA}. The Apple Watch dataset includes cardiac and movement signals collected from consumer-grade devices from a cohort of 31 young and healthy adults. For the MESA dataset, only the cardiac and movement sensing signals were used, which can be acquired from research-grade devices. The dataset consists of 1743 subjects from the aging population.

For these two representative datasets, our results suggested that three-stage sleep classification can be reliably achieved by employing advanced fusion techniques on the cardiac and movement sensing data, which can be easily acquired from consumer/research-grade devices. Several models developed in our study achieved the state-of-the art performance for three-stage sleep classification. We also evaluated the module parameter size and its corresponding inference time, which may play a vital role in ubiquitous computing applications. 

Moreover, we also investigated a visualization method to explore the decision-making process of the multimodal fusion model for three-stage sleep classification. The exploratory user research demonstrated that the gradient class activation map (Grad-CAM)~\cite{Selvaraju2020Grad-CAM:Localization} based sleep data visualization can be understood and used by humans, which facilitates the transparency of using DL in sleep health research.

This work contributes to the long-term non-intrusive three-stage sleep monitoring solution that may be deployed with mass-produced and inexpensive consumer-grade wearables, which may potentially be used for large-scale population-based sleep health studies and long-term sleep self-tracking.

\section{Sleep Monitoring and Ubiquitous Sensing }

\subsection{Clinical Sleep Monitoring and Sleep Physiology}
\subsubsection{Polysomnography and Sleep Stages}
Traditionally, gold-standard human sleep assessment was conducted in laboratory settings using polysomnography (PSG) which commonly involved electroencephalography (EEG), electromyography (EMG) and electrooculography (EOG). Together they facilitated the measurement of brain activity, alongside both muscle and eye movement~\cite{Berry2012FundamentalsMedicine}. 
PSG usually requires a sleep laboratory and a sleep technician in a controlled environment. Multiple skin electrodes will be placed on the subject’s body and the set-up procedure will take 45-60 minutes~\cite{Peters2021OvernightResults}.
It is impractical to measure sleep using this method for more than two consecutive nights as participants feel burdened.

Sleep stages consist of wakefulness (Wake), REM sleep and NREM sleep, three vigilance states for humans~\cite{IBER2007TheRules}. 
NREM sleep can be further subdivided into three stages: light sleep stage 1 (N1), light sleep stage 2 (N1) and deep sleep (N3)~\cite{Berry2017AASM2.4}. The alternate appearance of NREM sleep and REM sleep constitutes a sleep cycle. A healthy person typically has 4-5 sleep cycles in one night~\cite{Berry2012FundamentalsMedicine}. As the sleep cycle increases, the proportion of REM sleep increases, and the proportion of NREM sleep decreases. Five sleep stages can be distinguished by PSG, i.e., by analyzing the characteristics of EEG, EMG and EOG.
\subsubsection{Actigraphy and Sleep-Wake Monitoring}
Actigraphy is a valid method for detecting sleep/wake and is commonly used for ambulatory monitoring of sleep time or rhythms~\cite{Ancoli-Israel2003TheRhythms, Kapur2017ClinicalGuideline}. 
The actigraphy equipment is a type of wearable wristband that consists of various sensors that can monitor the light-off time and the movement of the limbs (using an accelerometer). However, it is limited to monitoring sleep-wake as the actigraphy data may not contain sufficient information to discern sleep stages.

\subsubsection {Cardiac Activities and Sleep Physiology} 
Cardiovascular autonomous control plays an essential role in sleep, and it will be different when transitioning to different sleep stages. The modulation of the autonomic nervous system (ANS) regulates cardiovascular functions during sleep onset and sleep stages~\cite{Malliani1991CardiovascularDomain, Montano2009HeartBehavior}. Heart rate variability (HRV) analysis is a classical tool for the ANS analysis. Research on HRV in sleep stages noted that REM sleep was characterized by a likely sympathetic predominance, while NREM sleep followed an opposite trend~\cite{Cabiddu2012ModulationRespiration, Mendez2006Time-varyingStages, Monti2002AutonomicSubjects}. 
The transition between wake, NREM and REM sleep is accompanied by changes of several HRV characters, such as the HR, Low-Frequency (LF) power, High-Frequency (HF) power and LF/HF ratio~\cite{Cabiddu2012ModulationRespiration, Chouchou2014HeartBrain, Mendez2006Time-varyingStages}.

Not all sleep stages are associated with brain activity. A study conducted by Desseilles et al.~\cite{Chouchou2014HeartBrain} through HRV and brain imaging analysis found close connectivity between autonomic cardiac modulations and the activity of certain brain areas during REM sleep. There is no conclusive connectivity between the brain and cardiac activity during NREM sleep. Therefore,  it may be not be easy to discern each NREM sleep stage accurately, without EEG signals.

\subsection{Ubiquitous Sensing Techniques for Sleep Monitoring }

\subsubsection{Wireless Sleep Monitoring}In recent years, the wireless technologies showed the potential of sleep monitoring, e.g., ballistocardiograph (BCG, with an example consumer product: Beddit \textsuperscript{TM}), which can monitor heart rate wirelessly~\cite{Giovangrandi2011BallistocardiographyRevisiting}. However, wireless-based approaches face some challenges when deploying them in clinical research owing to, a) the non-standardised measurement methods; b) the lack of precise understanding of the physiological origins that influences the signal waveform; c) comparatively low reliability and specificity of these signals to the existing clinical methods (for example, WiFi signals may be scattered by multiple subjects), which may hamper its wide applications in health and medical research~\cite{Giovangrandi2011BallistocardiographyRevisiting}.

~\subsubsection{Mobile and Miniaturized EEG}
The development of mobile EEG systems (e.g., Emotiv\textsuperscript{TM}) reduced the time spent on the electrodes installation process and improved the motion tolerance during the recording procedure~\cite{DeVos2014MobileCognition}.The main disadvantage of these devices is that the electrodes embedded in the cap are visible, and the form factor limitations prevent comfortable, continuous, and long-term sleep monitoring. Furthermore, once the electrode gel dries, the signal quality decreases, and the gel leaves residues~\cite{ Bleichner2015ExploringSee}, which may impact the wearing experience. The lightweight headband EEG (e.g. Sleep Profiler\textsuperscript{TM} and Dreem\textsuperscript{TM}) can monitor sleep in a natural environment. But it takes extra effort each time of wearing to adjust the equipment position to reduce the skin impedance to an acceptable level~\cite{Looney2012TheMonitoring}. In addition, compared to smartwatches, these devices are usually expensive. The recent development of ear EEG improved the wearing comfort and reduced electrode set-up time. Several studies demonstrated reasonable results of sleep stage classification using these prototypes~\cite{ Mikkelsen2019AccurateEar-EEG, Mikkelsen2017AutomaticEar-EEG}. However, these ear EEG devices commonly adopt around-ear or/and in-ear style and are made of silicone materials, which offer a bearable wearing experience, making them less popular than the mass-produced wearables~\cite{ Mikkelsen2019AccurateEar-EEG}.

\subsubsection{Consumer and Research-grade Wearables for Sleep Monitoring}
Many leading consumer products such as Fitbit\textsuperscript{TM}  and Xiaomi\textsuperscript{TM} band provide sleep stages tracking services. However, these consumer products commonly lack minimal validation, with poor algorithm transparency on data processing/sleep stage classification, resulting in these devices being precluded in clinical, research, or occupational settings~\cite{Khosla2018ConsumerStatement}. Nevertheless, another consumer product, Apple Watch, provides access to the accelerometer data and heart rate data, making it feasible to develop an algorithm for sleep health studies and self-trackers.

Consumer-grade wearable devices with diverse modalities offer a potential solution to ambulatory sleep tracking. Such sensors provide valuable, inexpensive, unobtrusive measurement tools to collect biological signals. Many of these wearables can communicate with smartphones, facilitating data collection and storage during large-scale population studies. Therefore, exploring the use of consumer-grade wearables in sleep and health studies becomes prevalent as the HR/HRV data and movement sensing data are generally available on these wearables~\cite{Walch2019SleepDevice}. 

The sleep stage classification based on ECG/PPG signals has also been investigated by~\cite{Fonseca2018AClassification, Fonseca2017CardiorespiratoryFields}. The results demonstrated promising performance. However, PPG data is generally unavailable on many consumer wearables, and ECG requires the skin electrodes to be placed near the heart. Collecting these raw signals may require research-grade wearables (e.g., Empatica\texttrademark\ E4 ), which demand additional financial costs for daily sleep monitoring.

Several previously published studies demonstrated that using HR/HRV features and movement sensing together could discern three sleep stages and achieved promising results~\cite{Hayano2017SleepSignals, Walch2019SleepDevice, Zhai2020MakingSensing}. Heterogeneous modalities may carry supplementary information for sleep stage classification. There is still much to be understood regarding how to construct this fusion architecture and which fusion method will be the most effective for sleep-stage classification. Our work adds to this knowledge. Exploring multimodal fusion strategies and methods to better integrate different physiological signals is of great significance for health research and self-monitoring of sleep using ubiquitous computing technology.

\section{Advanced Fusion Techniques for Three-stage Sleep Classification}
In this section, we will first discuss the current progress of multimodal fusion strategies and methods and their applications in sleep monitoring. We then present our study structure, followed by a technical description of three fusion strategies (early-stage, late-stage and hybrid fusion) and three methods (simple operation, attention mechanism and tensor-based method)

\subsection{Overview of Multimodal Fusion}
 Multimodal fusion in machine learning has been extensively studied in pattern recognition applications, such as in image and video captioning\cite{You2016ImageAttention}, visual question answering~\cite{Yang2016StackedAnswering}, audio-visual speech recognition~\cite{Afouras2018DeepRecognition} and affect recognition~\cite{Kapoor2005MultimodalEnvironments}. In the field of ubiquitous computing, multimodal fusion has also been adopted for human activity recognition~\cite{Liu2020GlobalFusion:Fusion}, sleep stage classification~\cite{Zhai2020MakingSensing}, fatigue assessment ~\cite{Bai2020FatigueSensors} and person identification~\cite{Chen2017Rapid}. The simple concatenation method was commonly adopted in these studies to combine the raw inputs or combine the representations obtained from the pre-trained model of each modality~\cite{radu2018multimodal}. Other researchers explored more advanced fusion methods, such as the attention-based fusion scheme for human activity recognition~\cite{Liu2020GlobalFusion:Fusion} 
 
For monitoring sleep, several previous works achieved promising results for sleep stage classification by concatenating multimodal intermediate features and feeding them into DL models~\cite{Zhai2020MakingSensing, Hayano2017SleepSignals}. However, these studies focus on the choice of modalities rather than the fusion techniques. Different modalities may contain complementary information. It is difficult to explicitly identify the best suitable cross-modal fusion architectures.

In terms of the movement sensing and cardiac sensing, they are different in signal-to-noise ratio, data generation process and measurement frequency. Moreover, the activity count is better in sleep/wakefulness classification, but it is difficult to discern different sleep stages~\cite{Ancoli-Israel2003TheRhythms}. For healthy adults, the difference in heart rate variability between REM sleep and wake is less than the difference in  NREM and REM sleep~\cite{Chouchou2014HeartBrain}.

The choice of fusion strategy and fusion method may thus influence the model classification performance. In recent years, the DL-based computational models have outperformed shallow machine learning models for sleep stage classifications, not only on unimodal data, but also on the multimodal data~\cite{Zhai2020MakingSensing,Palotti2019BenchmarkTechniques, Phan2019SeqSleepNet:Staging}. Therefore, this work will only focus on the multimodal fusion techniques based on DL networks.

\subsection{Problem Statement}
Based on the movement sensing and cardiac sensing data, the goal of this work is to comprehensively study how to use the advanced fusion techniques to reliably classify three-stage sleep. As demonstrated in Fig. ~\ref{fig:system_overview}, we adopt a sliding window method with window size $T$ and stride $S$ to segment sleep recordings into frames.  In each frame, we could either extract the handcraft features (e.g. heart rate  features that were deemed to be intermediate / mid-level features) from each sleep epoch that can provide physiologically meaningful features to the model ~\cite{Zhai2020MakingSensing,Fonseca2017CardiorespiratoryFields}, or we could use neural networks to extract the deep features. The time step $t$ represents one sleep epoch (i.e. every 30 seconds). Given that, we aim to map the data in a sliding window to a sleep stage that corresponds to the center point of the window (e.g., the purple point in the hypnogram in Fig. ~\ref{fig:system_overview}).  

Suppose the $i$th frame-wise time-series input data for cardiac sensing can be denoted as $\mathbf{X}_{car}^{(i)} \in \mathbb{R}^{C_{car} \times T}$, where $C_{car}$ denotes the number of features/input channels and $T$ denotes the sliding window length. For movement sensing, the input data can be denoted as $\mathbf{X}_{mov}^{(i)} \in \mathbb{R}^{C_{mov} \times T}$. The details of feature extraction will be introduced in Section ~\ref{sec:methods}. The goal of deep multimodal fusion is to determine a multilayer neural network $f(\cdot)$ whose output $\hat{y}^{(i)}$ is expected to be the same as the target $y^{(i)}$ as much as possible for each sample $(\mathbf{X}_{mov}^{(i)}, \mathbf{X}_{car}^{(i)})$. This can be implemented by minimizing the empirical loss $\mathcal{L}$ for classification denoted as:
\begin{equation} \label{eq: loss_func}
    \min_{f} \frac{1}{N}\sum_{i=1}^{N}\mathcal{L}\left(\hat{y}^{(i)} = f(\mathbf{X}_{mov}^{(i)}, \mathbf{X}_{car}^{(i)}), y^{(i)}\right)
\end{equation}

\subsection{Fusion Strategy}
Traditional fusion strategies include feature level fusion (e.g., ~\cite{Rattani2007FeatureBiometrics,Wang2019MultifocusDomain,Wang2021MedicalTransform}), score-level fusion (e.g., ~\cite{Guan2017EnsemblesWearables}) or decision-level fusion (e.g., ~\cite{Guan2015OnMethod}). In the end-to-end DL era, the boundary between multimodal representation and fusion has been blurred. Representation learning is interlaced with classification (or regression) objectives. Nevertheless, the fusion strategy for DL models may still be carried out in three stages, such as early fusion, late fusion and hybrid fusion~\cite{Zhang2020MultimodalApplications}. 

Fusion at different stages may influence the results of representation learning. For example, the early and late fusion may inhibit intra-modal or inter-modal interaction~\cite{Zhang2020MultimodalApplications}.   Neverova et al. noted that highly correlated modalities should be fused together~\cite{Neverova2016ModDrop:Recognition}. Hazirbas et al. demonstrated that the performance of fusion is highly affected by the choice of which layer to fuse~\cite{Hazirbas2017FuseNet:Architecture}. For sleep stage classification , the way to fuse heterogeneous intermediate features is worthy of exploration. In this study, we want to gain a comprehensive understanding at what stage we should fuse these inputs to achieve the most performance improvements on the three-stage sleep classification task. We adopted three commonly used fusion strategies, including early-stage fusion, late-stage fusion and hybrid fusion, as shown in Figure~\ref{fig:system_overview}.

\subsubsection{Early-stage Fusion}
In the early-stage fusion, data from different modalities (e.g., intermediate features) are concatenated (stacked) in the input stage. It is popular because of its simplicity, yet it is sub-optimal~\cite{Phan2021XSleepNet:Staging}.  Early-stage fusion firstly concatenates the cardiac (denoted as subscript $car$) and movement (denoted as subscript $act$) sensing data then feeds them into neural networks $h$ to make a corresponding prediction.
\begin{equation}
    \hat{y}^{(i)} = h(\text{Concatenate}(\mathbf{X}^{(i)}_{car}, \mathbf{X}^{(i)}_{mov})).
\end{equation}
where Concatenate is the matrix concatenation function.

\subsubsection{Late-stage Fusion}
Late-stage fusion is another prevalent way to fuse (high-level) representation from multiple sources. This fusion strategy allows high-level representations to have better intra-modal coherence. Late-stage fusion processes each modality's $c$th channel input data with a network $q$ and then combines all their high-level representations via an aggregation operation followed by the classification layers. It is denoted as:
\begin{equation}
    \hat{y}^{(i)} = \varphi(\text{Agg}(q(\mathbf{x}_{mov,1}^{(i)}), \cdots, q(\mathbf{x}_{mov, C_{mov}}^{(i)}),  q(\mathbf{x}_{car,1}^{(i)}), \cdots, q(\mathbf{x}_{car,C_{car}}^{(i)})))
\end{equation}
where Agg is the aggregation function and $\varphi$ denotes the classifier (e.g. fully connected layers), and  $\mathbf{x}^{(i)} \in \mathbb{R}^{1 \times T}$ and $T$ is the window length. The cardiac intermediate features are denoted as  $\mathbf{X}_{car}^{(i)} = [\mathbf{x}^{(i)}_{car,1}, \mathbf{x}^{(i)}_{car,2}, \cdots,  \mathbf{x}^{(i)}_{car,C_{car}}]$.  In this study, the aggregation function represents various fusion methods that will be introduced in the next section. $q$ denotes neural networks that learn the latent representation (e.g., for CNNs, it is the feature maps) of the $c$th intermediate feature, where $C_{car}$ and $C_{mov}$ are the numbers of the intermediate features for cardiac sensing and movement sensing respectively.

\subsubsection{Hybrid Fusion}
With hybrid fusion, the fusion may occur at multiple stages/layers of the DL models~\cite{radu2018multimodal,Zeng2021Multi-levelCounting}. It's commonly understood that the DL model hierarchically encodes features at different levels, starting from low-level to higher-level features as the layers go deeper~\cite{Heaton2018IanLearning}. In this study, we would not cover all possible combinations of fusion architecture. Therefore, following previous work ~\cite{radu2018multimodal}, we consider a simple scenario, which is firstly to fuse different input channel data belonging to the same modality (sharing a representation learning network) and then to fuse the high-level features from both modalities at the later stage. Formally, the hybrid-fusion strategy can be written as:
\begin{equation}
    \hat{y}^{(i)} =  \varphi(\text{Agg}(g_{mov}(\mathbf{X}_{mov}^{(i)}), g_{car}(\mathbf{X}_{car}^{(i)})))
\end{equation}
where $\varphi$ is the classifier (e.g., fully connected neural networks) and $g$ denotes the modality-specific networks (e.g., CNNs) that can learn representation from a specific modality such that the $g_{mov}$ does not share network parameters with $g_{car}$. Agg is the aggregation function that can be implemented as concatenation[51],  attention mechanism~\cite{Yang2016StackedAnswering} and tensor-based method~\cite{Fukui2016MultimodalGrounding}.

\subsection{Fusion Method}

Based on their complexity, fusion methods can be divided into three types: simple operations, attention-based methods and tensor-based methods. For feature vectors from different modalities, the concatenation and addition are two commonly used simple operations~\cite{Zhang2020MultimodalApplications}.  The attention mechanism is widely used for multimodal fusion. This usually refers to dynamically calculating a weight vector for each time step (or spatial position) and weighting a set of feature vectors~\cite{Bahdanau2015NeuralTranslate, Duan2020SOFA-Net:Counting}. 
For tensor-based methods, bilinear pooling is a method of fusing two unimodal representations to a joint presentation by calculating their outer product. This method can capture the multiplicative interaction between all elements in two vectors~\cite{Yu2017Multi-modalAnswering}.

For the early-stage fusion, we only adopted the concatenation as the only fusion method in this study. For the late-stage fusion, we selected two commonly used simple methods, which are concatenation and element-wise addition. Hybrid fusion provides aggregated representations for each modality, which facilitates flexible fusion methods. Apart from the simple operation methods, we also evaluated the attention mechanism and the tensor-based method. The choice of fusion method may be influenced by the application context. 
\subsubsection{Concatenation}
For \textbf{early-stage fusion}, the concatenation method concatenates inputs of all modalities into one matrix, which can be denoted as:  
\begin{equation} \label{eq: early_concatenation}
\matr{K}_{early}^{(i)} = \text{Concatenate}(\mathbf{X}_{car}^{(i)}, \mathbf{X}_{mov}^{(i)})
\end{equation}
where $\matr{K}_{early}^{(i)} \in \mathbb{R}^{(C_{mov} + C_{car}) \times T}$ .  $\matr{X}_{car}^{(i)} \in \mathbb{R}^{C_{car} \times T}$ is the intermediate feature matrix of cardiac sensing, $\mathbf{X}_{mov}^{(i)} \in \mathbb{R}^{C_{mov} \times T}$  is the intermediate feature matrix of movement sensing, and the $C_{car}$ represents the number of intermediate feature inputs of cardiac sensing.

For \textbf{late-stage fusion}, suppose we have the cardiac latent representation denoted as $\matr{X}_{car, c}^{\prime (i)} \in \mathbb{R}^{U \times L}$, which is learned from a neural network $g$ via $\matr{X}_{car,c}^{\prime (i)} = g(\mathbf{x}_{car,c}^{(i)})$. The movement representation matrix is computed in the same way, which can be formally denoted as $\matr{X}_{mov}^{\prime (i)} =  g(\mathbf{x}_{mov}^{(i)})$ where the $\matr{X}^{\prime (i)}_{mov} \in \mathbb{R}^{U \times L}$ is the latent representation of movement sensing. $L$ is the temporal length and $U$ is the representation's dimension. For example, in a convolutional neural network, $U$ is the number of feature maps. At the late stage, as the feature maps of each input intermediate feature were kept separately,  the concatenation operation concatenates these representations together, as follows: 
\begin{equation}\label{eq:concatenation}
    \matr{K}^{(i)}_{late} = \text{Concatenate}(\matr{X}^{\prime (i)}_ {mov, 1}, \cdots,\matr{X}^{\prime (i)}_{mov, C_{mov}},  \matr{X}^{\prime (i)}_{car,1},\cdots, \matr{X}^{\prime (i)}_{car,C_{car}})
\end{equation}
In this study, for the activity counts (handcraft feature) and cardiac features, the late-stage fusion's representation is denoted as $\matr{K}^{(i)}_{late} \in \mathbb{R}^{(C_{mov}+C_{car}) \times U \times L}$

For the \textbf{hybrid fusion}, the high-level representation of each modality is obtained from their own sub-network. The movement sensing representation is denoted as $\matr{X}_{mov}^{\prime\prime (i)} =  g_{mov}(\matr{X}_{mov}^{(i)})$ and the cardiac sensing is formally denoted as $\matr{X}_{car}^{\prime\prime (i)} =  g_{car}(\matr{X}_{car}^{(i)})$. The concatenation method for the hybrid fusion can be written as:
\begin{equation}
    \matr{K}^{(i)}_{hybrid} = \text{Concatenate}(\matr{X}^{\prime \prime(i)}_{car},\matr{X}^{\prime\prime(i)}_{mov})
\end{equation}
where $\matr{K}^{(i)}_{hybrid} \in \mathbb{R}^{2U \times L}$

\subsubsection{Addition}The second simple operation is the element-wise addition denoted as $\oplus$. For the \textbf{late-stage fusion}, the addition operation is to integrate the high-level representation of each channel from each modality. The method is formally denoted: 
\begin{equation}\label{eq:addition}
\matr{Q}_{late}^{(i)} = \matr{X}_{mov,1}^{\prime (i)} \oplus, \cdots, \matr{X}_{mov,C_{mov}}^{\prime (i)} \oplus \matr{X}_{car,1}^{\prime (i)} ,\cdots, \oplus \matr{X}_{car,C_{car}}^{\prime (i)}
\end{equation}
where $\matr{Q}^{(i)}_{late} \in \mathbb{R}^{U \times L}$.

For the \textbf{hybrid fusion}, the addition method will aggregate the high-level representation of each modality. Formally, it can be denoted as:
\begin{equation}
    \matr{Q}_{hybrid}^{(i)} = \matr{X}_{car}^{\prime\prime (i)} \oplus \matr{X}_{mov}^{\prime \prime(i)}
\end{equation}
where $\matr{Q}^{(i)}_{hybrid} \in \mathbb{R}^{U \times T}$.

\subsubsection{Attention Mechanism}
Attention methods have been broadly adopted in multimodal fusion tasks. For example, in VQA tasks the method used is to fuse the visual representations with the language representation~\cite{Yang2016StackedAnswering}. In our study we derived the attention model that could use the attention vectors to weight one modality based on the context of another modality. The meaning behind this is to filter the most significant information from a unimodal, which is jointly relevant for three-stage sleep classification. Therefore, we designed two attention fusion methods. The first one is Attention-on-Movement (Attention-on-Mov) and the second one is Attention-on-Cardiac (Attention-on-Car). 
Given the cardiac representation matrix $\matr{X}_{car}^{\prime (i)}$ and the movement representation matrix $\matr{X}_{mov}^{\prime (i)}$, we firstly feed them through a single-layer neural network and then apply the softmax function to generate the attention distribution over the temporal dimension, which is denoted as: 
\begin{equation} \label{eq: fused_matrix}
\matr{H}_{att}^{(i)} = \text{tanh}(\matr{W}_{car}\matr{X}_{car}^{\prime {(i)}} \oplus \matr{W}_{mov}\matr{X}_{mov}^{\prime {(i)}} + b_h)
\end{equation}
\begin{equation} \label{eq:attention_weights}
\matr{P}_{att}^{(i)} = \text{softmax}(\matr{W}_{att} \matr{H}_{att}^{(i)} + b_{att} )
\end{equation}
where $\matr{X}_{mov}^{\prime (i)} \in \mathbb{R}^{U \times L}$.  Suppose we have linear transformation matrices that include $\matr{W}_{mov}, \matr{W}_{car} \in \mathbb{R}^{D \times U}$ and $\matr{W}_{att} \in \mathbb{R}^{L \times D}$, then $\matr{H}_{att}^{(i)} \in \mathbb{R}^{D \times L}$ and $\matr{P}_{att}^{(i)} \in \mathbb{R}^{L \times L}$, where $D$ is the dimension of attention embedding space. The attention weight matrix is denoted as  $ \matr{P}_{att}^{(i)} = [\matr{p}_{att,1}^{(i)},\cdots, \matr{p}_{att, L}^{(i)}]$ and each temporal step has an attention vector $\matr{p}_{att,l}^{(i)}$, where $\sum\limits\matr{p}_{att,l}^{(i)} = 1$. The subscript $att$ stands for attention and $l$ is the temporal step index.

We assume that applying attention weights on different modalities will have an impact on the results. Therefore, two scenarios were studied in this work. The first method is to weight cardiac sensing representation based on the attention distribution and concatenate them to build the joint feature representation matrix. It can be written as:
\begin{equation}\label{eq:attention_on_car}
\matr{V}_{car}^{(i)}= \matr{X}_{car}^{\prime(i)}\matr{P}_{att}^{(i)}
\end{equation}
\begin{equation}\label{eq:joint_feature_from_car}
\matr{K}_{car}^{(i)} = \text{Concatenate}(\matr{V}_{car}^{(i)} ,\matr{X}_{mov}^{\prime(i)})
\end{equation}
We refer to this method as Attention-on-Car and $\matr{K}_{car}^{(i)} \in \mathbb{R}^{2U \times L}$

The second method is to weight the latent feature of movement sensing using the attention distribution, then concatenate them to build the joint representation matrix, which can be denoted as:
\begin{equation}
\matr{V}_{mov}^{(i)} = \matr{X}_{mov}^{\prime(i)}\matr{P}_{att}^{(i)}
\end{equation}
\begin{equation}
\matr{K}_{mov}^{(i)} = \text{concatenate}(\matr{V}_{mov}^{(i)} , \matr{X}_{car}^{\prime(i)})
\end{equation}
We refer to this method as Attention-on-Mov and $\matr{K}_{mov}^{(i)} \in \mathbb{R}^{2U \times L}$ is the merged joint representation.

\subsubsection{Bilinear Pooling Method}

Bilinear pooling is a method to compute the matrices outer product that can facilitate multiplication interaction between all elements in both matrices. It's a method often used to fuse visual feature vectors with textual feature vectors to create a joint representation space, even though their distribution may vary dramatically\cite{Lu2016HierarchicalAnswering, Fukui2016MultimodalGrounding}. 
During the NREM sleep period, our cardiac system is co-modulated by peripheral and sympathetic neural systems. The heart rate is generally below the average of wake and REM sleep period and accompanied with tiny tremors in limb movement. We hypothesized that the bilinear model may be able to capture such tiny differences between REM and NREM sleep. Given its superior representation learning capacity, it has achieved remarkable performance in fine-grained image classification tasks~\cite{Lin2015BilinearRecognition}. Bilinear model calculates the outer product of two matrices. In this work, suppose we have two feature representation matrices $\matr{X_{car}^{\prime(i)}}$ and $\matr{X}_{mov}^{\prime(i)}$, and the bilinear representation can be written as: 
\begin{equation} \label{eq:bilinear}
\Vec{k}_{bi}^{(i)} = \text{vec} (\matr{X}_{car}^{\prime(i)} \otimes \matr{X}_{mov}^{\prime(i)})
\end{equation}
The symbol of $\otimes $ denotes Kronecker product of two matrices, and the vec denotes the matrix vectorization. After the vectorization, we then perform an element-wise signed square-root as denoted: 
 \begin{equation}
      \matr{k}_{bi}^{(i)} \leftarrow \text{sign}(k_{bi}^{(i)}) \sqrt{|k_{bi}^{(i)}|}
 \end{equation}
and then apply $l_2$ normalization on the vector $\matr{k}_{bi}^{(i)}$. We pass the normalized vector to a linear function that can reduce the feature dimensions before feeding them to the classifier. 

%
\section{Experiment Design}
\label{sec:methods}
In this section, we describe the experimental design of advanced multimodal fusion strategies and methods for the three-stage sleep classification using wearable devices. We firstly introduce two open-access datasets used in the study, including the data collection, data pre-processing and feature extraction. Secondly, we illustrate four backbone networks used with advanced multimodal fusion techniques. Finally, we list the evaluation metrics used in the study.

\subsection{Dataset Description}
\subsubsection{Apple Watch Sleep Dataset} The first dataset used in our study is the Apple Watch Sleep Study\footnote{https://physionet.org/content/sleep-accel/1.0.0/}, which is an open-access dataset collected at the University of Michigan between 2017 and 2019~\cite{Walch2019SleepDevice, Walch2019MotionV1.0.0}. The dataset consists of 31 healthy subjects with no known sleep disorders or cardiovascular diseases and neurological or psychiatric impairment disorders\cite{Walch2019SleepDevice}. All subjects wore Apple Watch (Apple Inc. series 2 and 3) and performed continuous recording for 7 to 14 days, and then joined the PSG study in the sleep laboratory on the last day~\cite{Walch2019SleepDevice}. During the PSG study, all subjects wore Apple Watch, which recorded heart rate and triaxial acceleration~\cite{Walch2019SleepDevice}. The acceleration and heart rate were measured by Apple Watch and recorded by a custom-developed watch application using the built-in functions of the iOS Watch kit and HealthKit by creating a “Workout Session” in app~\cite{Walch2019MotionV1.0.0}. The PSG recordings were annotated according to the AASM rules~\cite{Walch2019SleepDevice}. 

The heart rate is measured by the PPG sensor of the Apple Watch and recorded as beats per minute (BPM), and a sample is returned from the Apple API every few seconds. The heart rate data is timestamped and the interval is between 2s and 5s. After the data cleaning process, we performed the feature engineering process on triaxial acceleration data; following~\cite{TeLindert2013SleepMEMS, Walch2019SleepDevice}, we used activity counts as the movement feature. The final activity counts were added for each sleep epoch. Since the heart rate collected from Apple Watch is calculated in two to five seconds, we may treat them as a “pseudo” instantaneous heart rate (IHR). In each sleep epoch, we calculated the summary statistics of the heart rate data (called HR statistics or HRS for short), which includes the mean, standard deviation, minimum, maximum, skewness and kurtosis of the heart rate. Together with the activity counts, we constructed a seven-dimensional vector for each sleep epoch from these intermediate summary features and called it the Apple ACT-HRS feature set.  

\subsubsection{MESA Dataset}
The Multi-Ethnic Study of Atherosclerosis (MESA) is a multi-site prospective study that includes 6,814 men and women. The ethnic groups include White, Black/African American, Hispanic, or Chinese, and the subjects are between the ages of 45 and 84~\cite{Zhang2018TheCommons,Chen2015Racial/ethnicMESA}. The study was designed prospectively to evaluate risk factors of cardiovascular disease. The study had 2,237 participants enrolled in the sleep exam, which includes seven days of wrist-worn actigraphy; they underwent concurrent PSG for one night (wrist-worn actigraphy collected concurrently)~\cite{Zhang2018TheCommons}. The subjects who reported regular night-time use of nocturnal oxygen or positive airway pressure devices were excluded from the sleep exam~\cite{Zhang2018TheCommons}. The actigraphy recorded the activity counts in 1/30Hz and ECG is recorded at 100Hz.

We used the method and data processing protocol provided by the benchmark study~\cite{Zhai2020MakingSensing}  to synchronize PSG, ECG and activity counts of each subject.  After the data pre-processing,  1,743 of 2,237 participants satisfied the data quality condition. Full details of the study setup, protocol and sampling rates are available in~\cite{Zhai2020MakingSensing,Zhang2018TheCommons,Chen2015Racial/ethnicMESA}. According to the feature set used in previous research~\cite{Zhai2020MakingSensing, Walch2019SleepDevice}, we used the same features including activity counts and eight HRV features derived from the NN interval data in each sleep epoch~\cite{Zhai2020MakingSensing}. The feature set consists of the Mean NNi, Standard Derivation of RR interval (SDNN), RR interval differences (SDSD), Very Low Frequency, Low Frequency, High Frequency Bands, Low Frequency to High Frequency Ratio and Total Power. These features have been investigated in several sleep physiology studies~\cite{Boudreau2013CircadianStages, Fonseca2017CardiorespiratoryFields, Tobaldini2013HeartSleep, Montano2009HeartBehavior,Chouchou2014HeartBrain}.  For each sleep epoch, we constructed the intermediate feature vector based on eight HRV features and the activity counts (a scalar value per sleep epoch), which we named the MESA ACT-HRV feature set.

In addition, we converted the NN intervals into IHR data, and calculated the statistical features of IHR and combined activity counts as the second intermediate feature set. The purpose is to study the feature effects on the choices of fusion strategies and methods. We named it the MESA ACT-HRS feature set.

\subsection{Evaluation Metrics}
\label{sec:methods_metrics}

For performance evaluation,  accuracy, Cohen's $\kappa$, mean F1 and time deviation(~\cite{Zhai2020MakingSensing}) were used. The time deviation that was used in the benchmark study \cite{Zhai2020MakingSensing} is denoted as  ($TD_k=  \frac{1}{N}\sum_{i=1}^{N}(Pred_{c}^i - GT_{c}^i)$). For a sleep stage $c$, the $Pred_{c}$ refers to the predicted minutes and $GT_{c}$ refers to the ground truth sleep minutes. The superscript $i$ represents the $i$th subject. The time deviation summarizes the mean bias of the total minutes of each sleep stage predicted by the classifier in the population. To understand the impact of individual differences in performance evaluation, this study adopted the subject-level evaluation. We calculated the metrics of each subject individually and obtained the mean value and 95\% confidence interval of each metric for the population.

\subsection{Experimental Procedure}
\label{sec:experimental_design}
Following previous work, we adopted a highly overlapping sliding window method with $S=1$ to segment the input time-series data. In~\cite{Zhai2020MakingSensing}, the hyperparameter tuning results showed that the window length can impact the prediction performance. For convolutional neural networks, a longer window produced better results compared with a shorter window. 

For each sleep epoch, we selected 50 adjacent (forward and backward) sleep epochs' data to construct the inputs with a window length of 101. The details are shown in Figure~\ref{fig:system_overview}. For the sliding window at the beginning and end of the recording, we filled these empty sleep epoch inputs with a value of -1. 
For the training, validation and testing, our experimental settings are as follows:
\begin{itemize}
    \item \textbf{Apple Watch Sleep Dataset} Following the experiment setting of previous work~\cite{Walch2019SleepDevice}, instead of using leave-one-subject-out-cross-validation, we adopted leave-two-subjects-out cross validation. Each fold had two subjects for testing, except for the last fold, which only contained one subject (total 31 subjects and 16 folds). In each fold, we then randomly split the subjects in training dataset into a validation dataset (20\%) and a training dataset (80\%). The validation set was used to select the best model for the test dataset. 
    \item \textbf{MESA Sleep Dataset} The dataset contains 1743 valid sleep records of subjects. We employed the hold-out method to divide the entire dataset into a test set of 348 subjects (20\%) and a training set of 1,395 subjects (80\%) following the previous study~\cite{Zhai2020MakingSensing}. The training set was further randomly split into a validation set (20\%) and a training set (80\%)~\cite{Zhai2020MakingSensing}. Again, the validation set was used to select the best model for the test dataset. 
\end{itemize}
All experiments conducted in this paper adopted the above setting for each dataset respectively.

In previous work~\cite{Zhai2020MakingSensing}, it was found that the performance improvement of three-stage sleep classification was more related to increasing the number of LSTM networks instead of increasing the number of CNN layers for the three-stage sleep classification task. Therefore, in this study, we focused on the design of CNN architecture. All experiments in this work adopted the Adam gradient update rule \cite{Kingma2015Adam:Optimization} with learning rate $\alpha=10^{-4}, \beta_1 = 0.9$, and $\beta_2 = 0.99$. No early-stopping or weight decay was adopted in training processing. The batch size was set to 1024 except for the experiments containing the bilinear method which were set to 512. For the attention method, we set the attention embedding dimension to 256. For the bilinear method, we reduced the size of the feature dimension to 1024 using a linear layer. The training epoch corresponding to both datasets was set to 20.

\subsection{Implementation Details}

\subsubsection{Hyperparameter Tuning and Backbone Networks}
\begin{figure}[t!]
    \centering
    \includegraphics[width=0.9\linewidth]{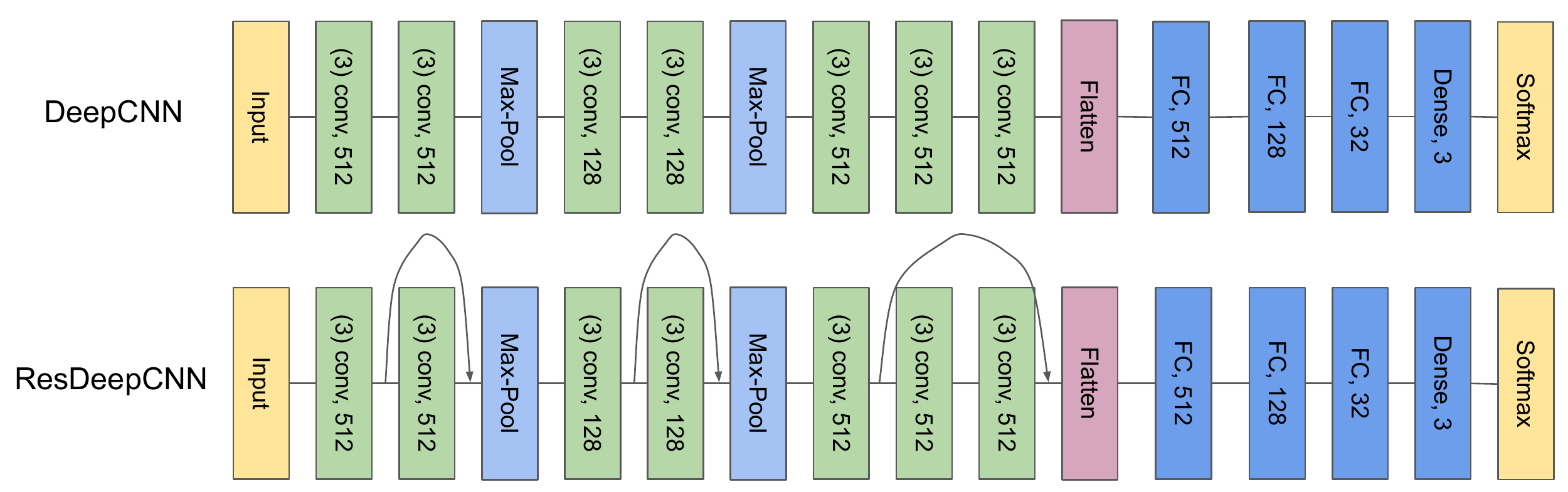}
	\caption{Backbone network used in this study.The DeepCNN network was selected from the hyperparameter search results. We added the skip connection inside each convolutional block and we referred to it as ResDeepCNN. The stride and padding value were set to 1 for all convolutional (conv) layers, and the kernel size was set to 3. The kernel size and stride were set to 2 for all max pooling (Max-Pool) layers. The dropout was applied after each fully connected (FC) layer, and the dropout rate was set to 0.25.}
    \label{fig:network}
\end{figure}

The fusion strategies and fusion methods may not benefit from a single layer convolutional neural network. To find a feasible backbone deep CNN that was capable of serving our study, inspired by~\cite{Simonyan2015VeryRecognition}, we designed a backbone network and conducted a hyperparameter search on 3-5 convolutional layer blocks (corresponding to 7-13 convolutional layers). From the hyperparameter tuning results, the network from the highest F1 validation score group was selected. We further gradually reduced the number of hidden units in fully connected layers and the experimental results showed slight improvements on model performance. We called this DeepCNN. To better understand the impact of modality fusion strategies and methods in different CNN architectures, inspired by~\cite{He2016IdentityNetworks, Orhan2017SkipSingularities}, we further added a skip connection in each convolutional block and called it ResDeepCNN, as the skip connection became an indispensable component in a variety of neural architectures that could boost representation learning.  Figure\ref{fig:network} lists the details of two network structures. As our study focuses on the fusion strategy and methods, the backbone network was merely designed to conduct the feasible experiments. More details about the hyperparameter search can be seen in Appendix~\ref{apx:hyper-parameter tuning}

\subsubsection{Backbone Network Setting}
For the early-stage fusion and hybrid fusion, DeepCNN and ResDeepCNN were the main networks for the experiments. We slightly adapted the DeepCNN and the ResDeepCNN for the late-stage fusion experiments according to~\cite{Yang2015DeepRecognition} , which allowed each input channel to share the convolutional kernels but kept the feature representation separate. This means that, for a convolutional layer, the feature maps extracted from each input channel would not be fused with the feature maps of other channels. Instead, each input channel's feature maps would be fused before the classification module (fully-connected layers). For instance, for DeepCNN in the Apple Watch dataset, if the input was an intermediate feature matrix that contained cardiac and movement sensing and was denoted as $\mathbf{S}^{(i)}_{0} \in \mathbb{R}^{7 \times 101}$ ( one movement feature and six HRS features), the feature map function $\mathcal{F}_{l+1} : \mathbf{S}^{(i)}_{l} \mapsto \mathbf{S}_{l+1}^{(i)} $ was realized by a convolutional layer, where $l$ denotes the $l$th convolutional layer. The output of the first convolutional layer was the feature map denoted as $\mathbf{S}_1^{(i)} = \mathcal{F}_1(\mathbf{S}^{(i)}_{0}) \in \mathbb{R}^{C_{1} \times 7 \times 101}$, where $C_{1}$ was the number of feature maps of the first CNN layer. In this way, the intermediate feature of each input channel was kept separate.  

\section{Results}
This section empirically compares each combination of multimodal fusion strategies and methods based on two scenarios. The first scenario is the MESA dataset which contains multimodal data that can be extracted from research grade-wearable devices. The second scenario is the Apple Watch dataset derived from the consumer-grade smartwatches (Apple Watch Series 2 and 3) with the sleep stages annotated using the gold-standard PSG study.

We reported the performance in the order of three fusion strategies which included early-stage fusion, late-stage fusion, and hybrid fusion, and three fusion methods, including simple operation (concatenation and addition), attention mechanism, and bi-linear pooling method. We also investigated the effects of different window lengths (51 and 21), and the corresponding results can be seen in Appendix~\ref{apx:results_of_21_51}. The experiments using raw accelerometer data and HR statistical features can be seen in Appendix~\ref{apx:raw_acc_feature_hrs_experiment}.

For consistency, all our fusion strategies and methods in each dataset were evaluated on the subject-level during the sleep recording period. Accuracy, Cohen's $\kappa$, the mean F1 score and time deviation (minutes) were calculated based on the predictions during the sleep recording period. In the end, we compared the model parameter size and inference time for each strategy and method. These factors are important for model selection in the context of ubiquitous computing.

\subsection{Apple Watch Dataset}
\subsubsection{Activity Counts and HRS Features}
The first experiment was performed based on activity counts and HRS feature set (ACT-HRS) derived from the consumer wearables.  
\begin{table}[hbtp]
  \centering
  \caption{Three-stage sleep classification results (mean ± standard error at 95\% confidence interval) for each combination of fusion strategies and methods with the Apple Watch dataset using ACT-HRS feature based on a window length of 101.}
    \resizebox{0.95\textwidth}{!}{%
\begin{tabular}{c|c|c|c|c|c|c|c|c}
\toprule
\multicolumn{3}{c|}{\textbf{Fusion Specifics}} & \multicolumn{3}{c|}{\textbf{Performance Metrics}} & \multicolumn{3}{c}{\textbf{Time Deviation(min.)}} \\
\midrule
\textbf{Fusion Strategy} & \textbf{Network} & \textbf{Fusion Method} & \textbf{Accuracy(\%)} & \textbf{Cohen's $\kappa$} & \textbf{Mean F1(\%)} & \textbf{Non-REM sleep} & \textbf{REM sleep} & \textbf{Wake} \\
\midrule
\multicolumn{1}{c|}{\multirow{2}[4]{*}{Early-Stage Fusion}} & DeepCNN & Concatenation & 72.3 $\pm$ 2.5 & 40.0 $\pm$ 5.8 & 59.0 $\pm$ 3.6 & -0.6 $\pm$ 22.3 & 11.8 $\pm$ 22.5 & -11.2 $\pm$ 6.9 \\
\cmidrule{2-9}  & ResDeepCNN & Concatenation & 76.0 $\pm$ 2.4 & 45.7 $\pm$ 6.1 & 63.4 $\pm$ 3.5 & 12.3 $\pm$ 17.2 & -4.0 $\pm$ 17.1 & -8.2 $\pm$ 7.3 \\
\midrule
\multicolumn{1}{c|}{\multirow{4}[4]{*}{Late-Stage Fusion}} & \multirow{2}[2]{*}{DeepCNN} & Concatenation & 76.2 $\pm$ 2.7 & 47.0 $\pm$ 7.1 & 63.7 $\pm$ 4.5 & 8.4 $\pm$ 16.5 & 1.3 $\pm$ 19.4 & -9.7 $\pm$ 7.1 \\
  &   & Addition & \boldmath{}\textbf{78.2 $\pm$ 2.1}\unboldmath{} & 49.9 $\pm$ 6.9 & 65.2 $\pm$ 3.8 & 11.4 $\pm$ 10.4 & -0.6 $\pm$ 11.2 & -10.8 $\pm$ 6.6 \\
\cmidrule{2-9}  & \multirow{2}[2]{*}{ResDeepCNN} & Concatenation & 75.5 $\pm$ 2.9 & 47.0 $\pm$ 6.5 & 63.4 $\pm$ 4.2 & 10.4 $\pm$ 18.5 & -2.1 $\pm$ 19.6 & -8.3 $\pm$ 7.0 \\
  &   & Addition & 78.2 $\pm$ 2.3 & \boldmath{}\textbf{52.0 $\pm$ 5.8}\unboldmath{} & \boldmath{}\textbf{66.5 $\pm$ 3.8}\unboldmath{} & 1.9 $\pm$ 11.7 & 4.7 $\pm$ 12.6 & -6.5 $\pm$ 7.3 \\
\midrule
\multicolumn{1}{c|}{\multirow{10}[4]{*}{Hybrid Fusion}} & \multirow{5}[2]{*}{DeepCNN} & Concatenation & 72.9 $\pm$ 3.0 & 40.1 $\pm$ 6.6 & 60.6 $\pm$ 3.8 & 10.7 $\pm$ 20.1 & 0.6 $\pm$ 19.5 & -11.4 $\pm$ 6.4 \\
  &   & Addition & 72.1 $\pm$ 2.9 & 38.6 $\pm$ 6.5 & 58.9 $\pm$ 3.9 & 8.8 $\pm$ 20.4 & 1.3 $\pm$ 19.9 & -10.2 $\pm$ 7.3 \\
  &   & Attention-on-Mov & 73.5 $\pm$ 2.9 & 41.3 $\pm$ 6.2 & 60.4 $\pm$ 3.9 & 1.2 $\pm$ 18.7 & 5.0 $\pm$ 18.9 & -6.2 $\pm$ 7.7 \\
  &   & Attention-on-Car & 71.1 $\pm$ 3.4 & 37.4 $\pm$ 6.7 & 58.1 $\pm$ 4.1 & 7.5 $\pm$ 24.3 & 0.7 $\pm$ 23.1 & -8.2 $\pm$ 7.5 \\
  &   & Bilinear & 69.5 $\pm$ 3.5 & 29.5 $\pm$ 7.5 & 53.0 $\pm$ 4.2 & 1.2 $\pm$ 25.7 & 10.0 $\pm$ 25.0 & -11.3 $\pm$ 8.5 \\
\cmidrule{2-9}  & \multirow{5}[2]{*}{ResDeepCNN} & Concatenation & 74.4 $\pm$ 2.4 & 44.2 $\pm$ 5.7 & 62.0 $\pm$ 3.3 & 2.5 $\pm$ 16.7 & 5.3 $\pm$ 16.9 & -7.8 $\pm$ 7.0 \\
  &   & Addition & 74.9 $\pm$ 2.3 & 44.3 $\pm$ 5.4 & 62.3 $\pm$ 3.7 & 12.8 $\pm$ 14.2 & -7.6 $\pm$ 15.6 & -5.2 $\pm$ 8.2 \\
  &   & Attention-on-Mov & 75.2 $\pm$ 2.6 & 45.0 $\pm$ 5.6 & 63.1 $\pm$ 3.4 & 10.4 $\pm$ 19.4 & -2.0 $\pm$ 19.5 & -8.4 $\pm$ 7.2 \\
  &   & Attention-on-Car & 72.2 $\pm$ 3.0 & 41.5 $\pm$ 6.8 & 59.7 $\pm$ 3.8 & -2.9 $\pm$ 25.3 & 12.0 $\pm$ 26.4 & -9.1 $\pm$ 6.7 \\
  &   & Bilinear & 70.8 $\pm$ 3.5 & 38.4 $\pm$ 7.5 & 58.1 $\pm$ 4.4 & -2.8 $\pm$ 22.9 & 6.4 $\pm$ 24.8 & -3.5 $\pm$ 8.0 \\
\bottomrule
\end{tabular}%

}
\label{tab:apple_hrs_rp}%
\end{table}%

Table~\ref{tab:apple_hrs_rp} lists the subject-level evaluation results of the Apple Watch dataset based on the window length of 101 during the sleep recording period. Since Apple Watch sampled the heart rate data with the unknown resolution and method, we only performed the experiments based on the ACT-HRS feature setting.

Overall, the ResDeepCNN achieved the highest mean F1 score of 66.5\%, the Cohen's $\kappa$ of 52 and accuracy of 78.2\% using the addition method in late-stage fusion. The same methods used on DeepCNN were higher than these in early-stage fusion too. 

In the hybrid fusion strategy, using the Attention-on-Mov method achieved the highest scores irrespective of backbone networks. We observed the same pattern in the experiments using the window lengths 51 and 21. For window lengths 51 and 21, the highest performed models in each category were lower than the highest performed models using the window length of 101. We listed these results for window lengths 51 and 21 in Appendix~\ref{apx:results_of_21_51}. 

\subsection{MESA Sleep Dataset Results}

The second experiment was conducted on the MESA dataset. It was by far the largest sleep dataset that contained activity counts and instantaneous heart rate, which might be extracted from research-grade wearable devices. Again, we performed experiments on two different feature sets. The first feature set included activity counts and HRV features (ACT-HRV)~\cite{Zhai2020MakingSensing}, while the second feature set was ACT-HRS derived using the same feature extraction method in the Apple Watch dataset. 

\subsubsection{MESA Activity Counts and HRV Features}
The reason for using the HRV features was that they had sleep physiological meaning. Table \ref{tab:mesa_act_hrv_rp} shows the subject-level evaluation results based on the window length of 101. For the early-stage and late-stage fusion, the results of two backbone networks were comparable, which showed the skipping connections did not improve the classification performance.

\begin{table}[htp]
  \centering
  \caption{
 {Three-stage sleep classification results (mean $\pm$ standard error at 95\% confidence interval) for each combination of fusion strategies and methods with the MESA test dataset using the ACT-HRV feature set based on a window length of 101.}
}
  \resizebox{0.95\textwidth}{!}{%

\begin{tabular}{c|c|c|c|c|c|c|c|c}
\toprule
\multicolumn{3}{c|}{\textbf{Fusion Specifics}} & \multicolumn{3}{c|}{\textbf{Performance Metrics}} & \multicolumn{3}{c}{\textbf{Time Deviation(min.)}} \\
\midrule
\textbf{Fusion Strategy} & \textbf{Network} & \textbf{Fusion Method} & \textbf{Accuracy(\%)} & \textbf{Cohen's $\kappa$} & \textbf{Mean F1(\%)} & \textbf{Non-REM sleep} & \textbf{REM sleep} & \textbf{Wake} \\
\midrule
\multicolumn{1}{c|}{\multirow{2}[4]{*}{Early-Stage Fusion}} & DeepCNN & Concatenation & 78.6 $\pm$ 0.9 & 62.8 $\pm$ 1.8 & 71.1 $\pm$ 1.3 & 27.1 $\pm$ 6.9 & -6.5 $\pm$ 3.6 & -20.7 $\pm$ 6.4 \\
\cmidrule{2-9}  & ResDeepCNN & Concatenation & 78.0 $\pm$ 1.1 & 60.2 $\pm$ 2.0 & 71.1 $\pm$ 1.4 & 54.1 $\pm$ 7.0 & 1.2 $\pm$ 3.8 & -55.3 $\pm$ 6.6 \\
\midrule
\multicolumn{1}{c|}{\multirow{4}[4]{*}{Late-Stage Fusion}} & \multirow{2}[2]{*}{DeepCNN} & Concatenation & 79.6 $\pm$ 0.9 & 64.3 $\pm$ 1.8 & 72.5 $\pm$ 1.3 & 12.9 $\pm$ 6.6 & 0.1 $\pm$ 3.5 & -13.0 $\pm$ 6.2 \\
  &   & Addition & 78.5 $\pm$ 0.9 & 62.3 $\pm$ 1.8 & 71.1 $\pm$ 1.3 & 25.7 $\pm$ 6.4 & -6.8 $\pm$ 3.3 & -18.9 $\pm$ 6.4 \\
\cmidrule{2-9}  & \multirow{2}[2]{*}{ResDeepCNN} & Concatenation & 79.3 $\pm$ 0.9 & 64.4 $\pm$ 1.7 & 72.6 $\pm$ 1.2 & 8.9 $\pm$ 6.4 & 4.2 $\pm$ 3.4 & -13.1 $\pm$ 6.1 \\
  &   & Addition & 78.6 $\pm$ 1.0 & 62.8 $\pm$ 1.9 & 71.4 $\pm$ 1.3 & 2.4 $\pm$ 6.9 & -3.0 $\pm$ 3.5 & 0.6 $\pm$ 6.7 \\
\midrule
\multicolumn{1}{c|}{\multirow{10}[4]{*}{Hybrid Fusion}} & \multirow{5}[2]{*}{DeepCNN} & Concatenation & 77.6 $\pm$ 1.1 & 62.7 $\pm$ 1.8 & 71.4 $\pm$ 1.3 & -12.7 $\pm$ 7.3 & 7.2 $\pm$ 3.9 & 5.5 $\pm$ 7.0 \\
  &   & Addition & 79.0 $\pm$ 0.9 & 62.9 $\pm$ 1.7 & 70.7 $\pm$ 1.3 & 53.6 $\pm$ 6.9 & -15.0 $\pm$ 3.3 & -38.6 $\pm$ 6.3 \\
  &   & Attention-on-Mov & 79.0 $\pm$ 1.0 & 63.9 $\pm$ 1.8 & 72.1 $\pm$ 1.3 & 2.3 $\pm$ 7.0 & -4.9 $\pm$ 3.5 & 2.6 $\pm$ 6.9 \\
  &   & Attention-on-Car & 78.1 $\pm$ 1.0 & 63.0 $\pm$ 1.7 & 71.6 $\pm$ 1.3 & -2.1 $\pm$ 6.8 & 6.5 $\pm$ 4.0 & -4.4 $\pm$ 6.3 \\
  &   & Bilinear & 75.7 $\pm$ 0.9 & 58.6 $\pm$ 1.8 & 68.8 $\pm$ 1.2 & 3.7 $\pm$ 6.8 & 16.6 $\pm$ 4.0 & -20.3 $\pm$ 6.3 \\
\cmidrule{2-9}  & \multirow{5}[2]{*}{ResDeepCNN} & Concatenation & 79.7 $\pm$ 0.9 & 65.3 $\pm$ 1.7 & 72.7 $\pm$ 1.3 & 6.4 $\pm$ 6.7 & -7.7 $\pm$ 3.4 & 1.3 $\pm$ 6.7 \\
  &   & Addition & \boldmath{}\textbf{79.8 $\pm$ 0.9}\unboldmath{} & 64.1 $\pm$ 1.7 & 72.7 $\pm$ 1.3 & 24.1 $\pm$ 6.9 & -9.7 $\pm$ 3.2 & -14.4 $\pm$ 6.5 \\
  &   & Attention-on-Mov & 79.6 $\pm$ 1.0 & \boldmath{}\textbf{65.5 $\pm$ 1.8}\unboldmath{} & \boldmath{}\textbf{73.3 $\pm$ 1.3}\unboldmath{} & -0.9 $\pm$ 6.4 & 6.1 $\pm$ 3.7 & -5.1 $\pm$ 6.3 \\
  &   & Attention-on-Car & 78.5 $\pm$ 1.0 & 62.7 $\pm$ 1.7 & 70.5 $\pm$ 1.3 & 37.6 $\pm$ 7.0 & -9.5 $\pm$ 4.0 & -28.1 $\pm$ 6.2 \\
  &   & Bilinear & 75.7 $\pm$ 0.9 & 58.6 $\pm$ 1.8 & 68.8 $\pm$ 1.2 & 3.7 $\pm$ 6.8 & 16.6 $\pm$ 4.0 & -20.3 $\pm$ 6.3 \\
\bottomrule
\end{tabular}%

}

  \label{tab:mesa_act_hrv_rp}%
\end{table}%

For the hybrid fusion strategy, the ResDeepCNN achieved the highest accuracy, the Cohen's $\kappa$ and the mean F1 score of 79.6\%, 65.5 and 73.3\%, respectively, using the Attention-on-Mov method. The results were statistically significant ($p<0.05$) and higher than the models in the early-stage fusion. Those metrics were higher than the Attention-on-Car models too. Similar to the Apple Watch dataset, the performance of models based on window lengths 51 and 21 tend to be worse than experiments performed with window length 101. We list these results in Appendix~\ref{apx:results_of_21_51}. 

In terms of time deviation, DeepCNN achieved the optimal time deviation using the Attention-on-Mov method. The mean value of NREM sleep time deviation was 0.9, and the mean value of REM sleep time deviation was 6.1.

\subsubsection{MESA Activity Counts and HRS Features}
We derived the heart rate statistical features from the instantaneous heart rate (IHR) data in the MESA dataset. The purpose was to understand whether the type of intermediate feature would cause a difference in results.

The subject-level evaluation is shown in Table~\ref{tab:mesa_hrs_rp}. In the early-stage fusion, similar to the ACT-HRV feature setting, the results of two backbone networks were comparable. The ResDeepCNN, using the Attention-on-Mov fusion method, achieved the highest accuracy, the Cohen's $\kappa$, and the mean F1 score of 80.3\%, 65.6, and 72.9\%, respectively. However, the Attention-on-Mov model based on the ACT-HRS feature set highly overestimated the NREM sleep time and underestimated the wake minutes. Again, the models with window lengths 51 and 21 achieved lower performance than 101. Therefore, we listed these results in Appendix~\ref{apx:results_of_21_51} 

\begin{table}[htbp]
  \centering
  \caption{
  Three-stage sleep classification results (mean $\pm$ standard error at 95\% confidence interval) for each combination of fusion strategies and methods with the MESA test dataset using the ACT-HRS feature set based on a window length of 101.
}
  \resizebox{0.95\textwidth}{!}{%

\begin{tabular}{c|c|c|c|c|c|c|c|c}
\toprule
\multicolumn{3}{c|}{\textbf{Fusion Specifics}} & \multicolumn{3}{c|}{\textbf{Performance Metrics}} & \multicolumn{3}{c}{\textbf{Time Deviation(min.)}} \\
\midrule
\textbf{Fusion Strategy} & \textbf{Network} & \textbf{Fusion Method} & \textbf{Accuracy(\%)} & \textbf{Cohen's $\kappa$} & \textbf{Mean F1(\%)} & \textbf{Non-REM sleep} & \textbf{REM sleep} & \textbf{Wake} \\
\midrule
\multicolumn{1}{c|}{\multirow{2}[4]{*}{Early-Stage Fusion}} & DeepCNN & Concatenation & 78.0 $\pm$ 1.0 & 63.4 $\pm$ 1.8 & 72.0 $\pm$ 1.2 & 2.7 $\pm$ 7.0 & 15.6 $\pm$ 4.0 & -18.2 $\pm$ 6.4 \\
\cmidrule{2-9}  & ResDeepCNN & Concatenation & 76.9 $\pm$ 1.1 & 61.9 $\pm$ 1.9 & 71.1 $\pm$ 1.3 & -36.7 $\pm$ 7.8 & 12.1 $\pm$ 4.2 & 24.5 $\pm$ 7.5 \\
\midrule
\multicolumn{1}{c|}{\multirow{4}[4]{*}{Late-Stage Fusion}} & \multirow{2}[2]{*}{DeepCNN} & Concatenation & 79.1 $\pm$ 1.0 & 64.7 $\pm$ 1.7 & 72.8 $\pm$ 1.3 & 0.6 $\pm$ 6.9 & 7.5 $\pm$ 3.7 & -8.1 $\pm$ 6.4 \\
  &   & Addition & 78.1 $\pm$ 0.9 & 62.8 $\pm$ 1.6 & 70.6 $\pm$ 1.2 & 23.1 $\pm$ 6.6 & -4.5 $\pm$ 3.7 & -18.5 $\pm$ 6.3 \\
\cmidrule{2-9}  & \multirow{2}[2]{*}{ResDeepCNN} & Concatenation & 77.8 $\pm$ 1.0 & 62.5 $\pm$ 1.7 & 71.0 $\pm$ 1.3 & -1.1 $\pm$ 7.3 & -0.7 $\pm$ 3.5 & 1.8 $\pm$ 6.8 \\
  &   & Addition & 77.7 $\pm$ 1.0 & 62.6 $\pm$ 1.7 & 70.7 $\pm$ 1.2 & 24.3 $\pm$ 7.0 & 3.7 $\pm$ 3.9 & -28.0 $\pm$ 6.4 \\
\midrule
\multicolumn{1}{c|}{\multirow{10}[4]{*}{Hybrid Fusion}} & \multirow{5}[2]{*}{DeepCNN} & Concatenation & 78.2 $\pm$ 0.9 & 64.4 $\pm$ 1.7 & 70.2 $\pm$ 1.2 & 18.5 $\pm$ 7.1 & -14.6 $\pm$ 3.3 & -3.9 $\pm$ 6.5 \\
  &   & Addition & 78.1 $\pm$ 0.9 & 62.2 $\pm$ 1.8 & 71.2 $\pm$ 1.2 & 14.7 $\pm$ 7.4 & 1.4 $\pm$ 3.6 & -16.1 $\pm$ 6.8 \\
  &   & Attention-on-Mov & 79.2 $\pm$ 0.9 & 63.8 $\pm$ 1.8 & 71.8 $\pm$ 1.3 & 28.1 $\pm$ 7.3 & -4.3 $\pm$ 3.5 & -23.9 $\pm$ 6.5 \\
  &   & Attention-on-Car & 76.6 $\pm$ 1.0 & 61.4 $\pm$ 1.7 & 70.4 $\pm$ 1.2 & -7.4 $\pm$ 7.9 & 17.7 $\pm$ 4.7 & -10.3 $\pm$ 6.7 \\
  &   & Bilinear & 75.6 $\pm$ 0.9 & 58.0 $\pm$ 1.8 & 67.7 $\pm$ 1.2 & 6.3 $\pm$ 7.6 & -8.1 $\pm$ 3.7 & 1.8 $\pm$ 7.1 \\
\cmidrule{2-9}  & \multirow{5}[2]{*}{ResDeepCNN} & Concatenation & 79.4 $\pm$ 1.0 & 64.4 $\pm$ 1.7 & 72.7 $\pm$ 1.2 & 31.7 $\pm$ 6.9 & 4.9 $\pm$ 3.6 & -36.6 $\pm$ 6.3 \\
  &   & Addition & 78.9 $\pm$ 0.9 & 63.6 $\pm$ 1.8 & 72.2 $\pm$ 1.2 & 24.6 $\pm$ 7.0 & 4.9 $\pm$ 3.5 & -29.5 $\pm$ 6.5 \\
  &   & Attention-on-Mov & \boldmath{}\textbf{80.3 $\pm$ 0.9}\unboldmath{} & \boldmath{}\textbf{65.6 $\pm$ 1.7}\unboldmath{} & \boldmath{}\textbf{72.9 $\pm$ 1.3}\unboldmath{} & 35.5 $\pm$ 6.9 & 0.8 $\pm$ 3.6 & -36.3 $\pm$ 6.2 \\
  &   & Attention-on-Car & 79.3 $\pm$ 0.9 & 62.8 $\pm$ 1.7 & 71.1 $\pm$ 1.2 & 29.1 $\pm$ 7.1 & -2.4 $\pm$ 3.7 & -26.7 $\pm$ 6.3 \\
  &   & Bilinear & 74.1 $\pm$ 0.9 & 56.8 $\pm$ 1.7 & 66.9 $\pm$ 1.2 & 6.2 $\pm$ 7.1 & 11.7 $\pm$ 4.2 & -17.9 $\pm$ 6.5 \\
\bottomrule
\end{tabular}%

}
  \label{tab:mesa_hrs_rp}%
\end{table}%

\subsection{Inference Efficiency}

\begin{table}[htbp]
  \caption{The number of model parameters and inference time of each combination of fusion strategies and methods evaluated in millions of parameters and milliseconds respectively with the  \textit{Apple Watch} dataset, using the \textit{ACT-HRS} feature sets based on a window length of 101.}
\centering
\resizebox{0.7\textwidth}{!}{%
\begin{tabular}{c|c|c|c|c}
\toprule
\textbf{Fusion Strategy} & \textbf{Network} & \textbf{Fusion Method} & \textbf{Total Parameters (M)} & \textbf{Inference Time (ms per sample)} \\
\midrule
\multicolumn{1}{c|}{\multirow{2}[4]{*}{Early-Stage}} & DeepCNN & Concatenation & \textbf{9.44} & \boldmath{}\textbf{3.52$\pm$0.08}\unboldmath{} \\
\cmidrule{2-5}  & ResDeepCNN & Concatenation & 9.44 & 3.54$\pm$0.08 \\
\midrule
\multicolumn{1}{c|}{\multirow{4}[4]{*}{Late-Stage Fusion}} & \multirow{2}[2]{*}{DeepCNN} & Concatenation & 48.75 & 22.9$\pm$0.17 \\
  &   & Addition & 9.43 & 32.25$\pm$5.53 \\
\cmidrule{2-5}  & \multirow{2}[2]{*}{ResDeepCNN} & Concatenation & 48.75 & 31.16$\pm$5.06 \\
  &   & Addition & 9.43 & 22.11$\pm$0.25 \\
\midrule
\multicolumn{1}{c|}{\multirow{8}[4]{*}{Hybrid Fusion}} & \multirow{4}[2]{*}{DeepCNN} & Concatenation & 18.80 & 7.13$\pm$0.12 \\
  &   & Addition & 12.24 & 7.01$\pm$0.12 \\
  &   & Attention-on-Act & 19.07 & 7.32$\pm$0.12 \\
  &   & Bilinear & 274.65 & 10.09$\pm$0.11 \\
\cmidrule{2-5}  & \multirow{4}[2]{*}{ResDeepCNN} & Concatenation & 18.80 & 7.02$\pm$0.16 \\
  &   & Addition & 12.24 & 7.0$\pm$0.14 \\
  &   & Attention-on-Act & 19.07 & 7.27$\pm$0.17 \\
  &   & Bilinear & 274.65 & 10.22$\pm$0.17 \\
\bottomrule
\end{tabular}%

    }

  \label{tab:inference_time}%
\end{table}%

In the mobile computing scenario of three-sleep stage classification, the model based on the deep learning architecture may require sufficient computing resources. This  may be a challenge for many inexpensive or low-end smartwatches and smartphones. Table~\ref{tab:inference_time} shows the model parameter size and inference time of each combination of fusion strategies and methods. In addition, we counted the number of trainable parameters and calculated the time required for forward propagation (running on CPU). All experiments were conducted using Pytorch 1.6 and the hardware platform consisted of 8 cores AMD-7 3700X with 4.4GHz and 64GB DDR4 memory. We independently ran each model 10 times on the Apple Watch dataset. Each time, we inferred 500 samples (sleep epochs) using the Pytorch profiling module to calculate the statistical summary of the inference time.

Overall, the models using the addition method in late-stage fusion, hybrid fusion, and concatenation in early-stage fusion had the least model parameters. As a result, the models in early-stage fusion achieved the shortest inference time. The addition method in the late-stage fusion had the same number of parameters as the models in early-stage fusion, but the inference time was increased by 7-10 times. This was because the late-stage fusion calculated the feature maps of each input channel separately and fused them before the classifier module (fully connected layers). Consequently, the feature matrix extracted by the convolutional module was a 3D tensor (e.g., the number of input feature dimensions $\times$  number of feature maps $\times$ temporal steps) in the late-stage fusion. In contrast, the early-stage fusion generated a 2D tensor (e.g., number of feature maps $\times$ temporal steps). The convolution operation required additional time to calculate the feature maps of each input channel.

The bilinear model had the largest model parameters. Most model parameters belonged to the feature representation module, which contained a fully connected layer to reduce the dimension of feature representation at an order of two magnitudes. Since the calculation speed of the fully connected (FC) layer was much faster than the convolutional layer, the inference time did not increased as much as the model parameter size. 

\section{Exploration of using Grad-CAM on Sleep Sensing Data}

One of the major drawbacks to the consumer sleep monitoring devices was inaccurate results with an unknown decision making process, which would harm the user's confidence and trust in the devices~\cite{Ravichandran2017MakingHealth}. Explanation of the decision-making process lies at the heart of a responsible research in applied machine learning~\cite{Guidotti2018AModels}. Before building confidence and trust, the first step is communication and explanation~\cite{Lim2009WhySystems}. Unlike multimedia data, the time-series data requires the user to associate a meaning to the values in the temporal dimension~\cite{Guidotti2018AModels}.

To investigate in what kind of visualization of the decision-making process of multimodal fusion can be understood by humans, the gradient class activation map (Grad-CAM)~\cite{Selvaraju2020Grad-CAM:Localization} was adopted to visualize the important areas that matter to a specific class prediction from a qualitative perspective. In sleep physiology, the HRV features such as HF and LF have been proven to be different based on different sleep stages, e.g., NREM sleep is associated with a lower overall HRV, and REM sleep is accompanied by increased variability~\cite{Vanoli1995a}. Wake is associated most often with changes in activity counts. Based on these phenomena, we selected subjects from the MESA test dataset with F1 scores greater than 90\%. To obtain clear graphics, a post-processing method was used on the Grad-CAM output, which simplified the time-series data by setting the heat map value to 1 if the CAM value was greater than a threshold of 0.8, otherwise it was set to 0.2. 

Not all sleep epochs generated have consistent patterns, as the neural network is known to be able to learn background information that is relevant to the classification~\cite{Guidotti2018AModels, Adadi2018PeekingXAI}. Then we chose sleep epochs with good quality~\footnote{ a) stay in a sleep stage for at least 5 minutes. b) The fluctuation patterns of the highlighted areas should be similar to that described in the studies of sleep physiology.}.  We then presented the visualization results to sleep experts, and the feedback showed that sleep technologists tend to use fewer PSG channels to reduce the overload of irrelevant information during sleep stage annotation.

\begin{figure}[h]
\centering
\subcaptionbox{Wake}{\includegraphics[width=0.3\textwidth]{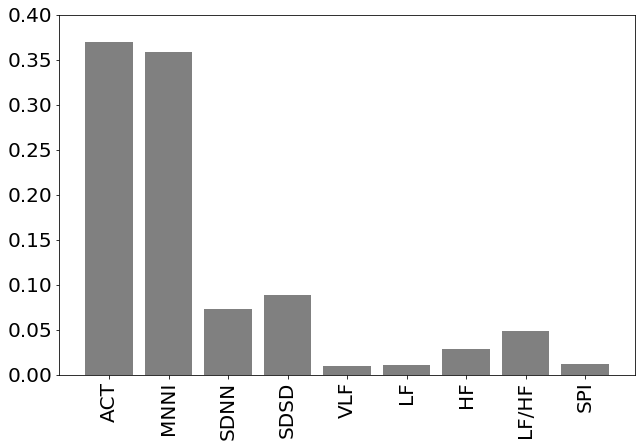}}%
\hfill 
\subcaptionbox{NREMS}{\includegraphics[width=0.3\textwidth]{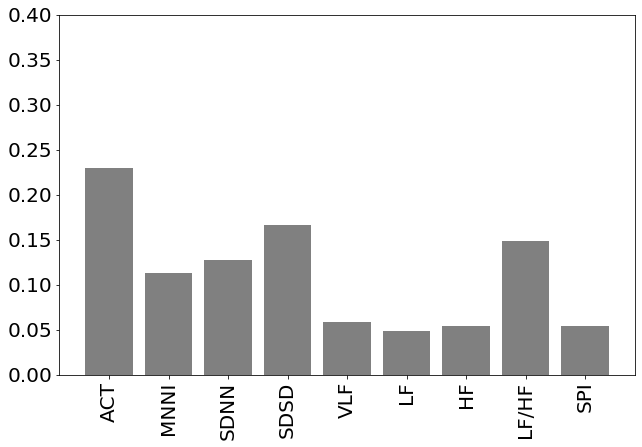}}%
\hfill 
\subcaptionbox{REMS}{\includegraphics[width=0.3\textwidth]{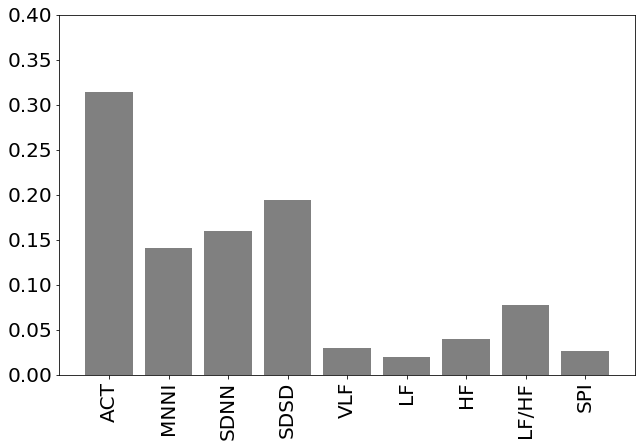}}%
\caption{The mean of total class activation value for each sleep stage from MESA dataset (ACT-HRV feature) using ResDeepCNN (Addition) in the late-stage fusion. ACT {:} Activity Counts,  MNNI {:} Mean NNI, SDNN{:} Standard Deviation of NNI, SDSD {:} Successive RR Interval Differences,VLF{:} Very-Low-Frequency Band, LF{:} Low-Frequency Band, HF{:} High-Frequency Band, LF/HF{:} The ratio of Low Frequency to High Frequency, SPI{:} The Signal Power Intensity}
\label{fig:featue_importance}
\end{figure}

To reduce the input channels, filtering them by feature importance scores calculated on the CAM value is a feasible method. We randomly selected 1k correctly predicted samples for each sleep stage from the test set and calculated their CAM values under the window length of 101. For each sleep epoch, we counted the number of time steps if their CAM value was greater than 0.8 and summed and normalized them as occurrences for each intermediate feature to obtain their relative importance score. Afterwards, we then calculated the mean value of all samples for each intermediate feature per sleep stage. Figure~\ref{fig:featue_importance} shows the feature importance for each sleep stage. 

\begin{figure}[t!]
\centering
\subcaptionbox{Wake}{\includegraphics[width=0.333\textwidth]{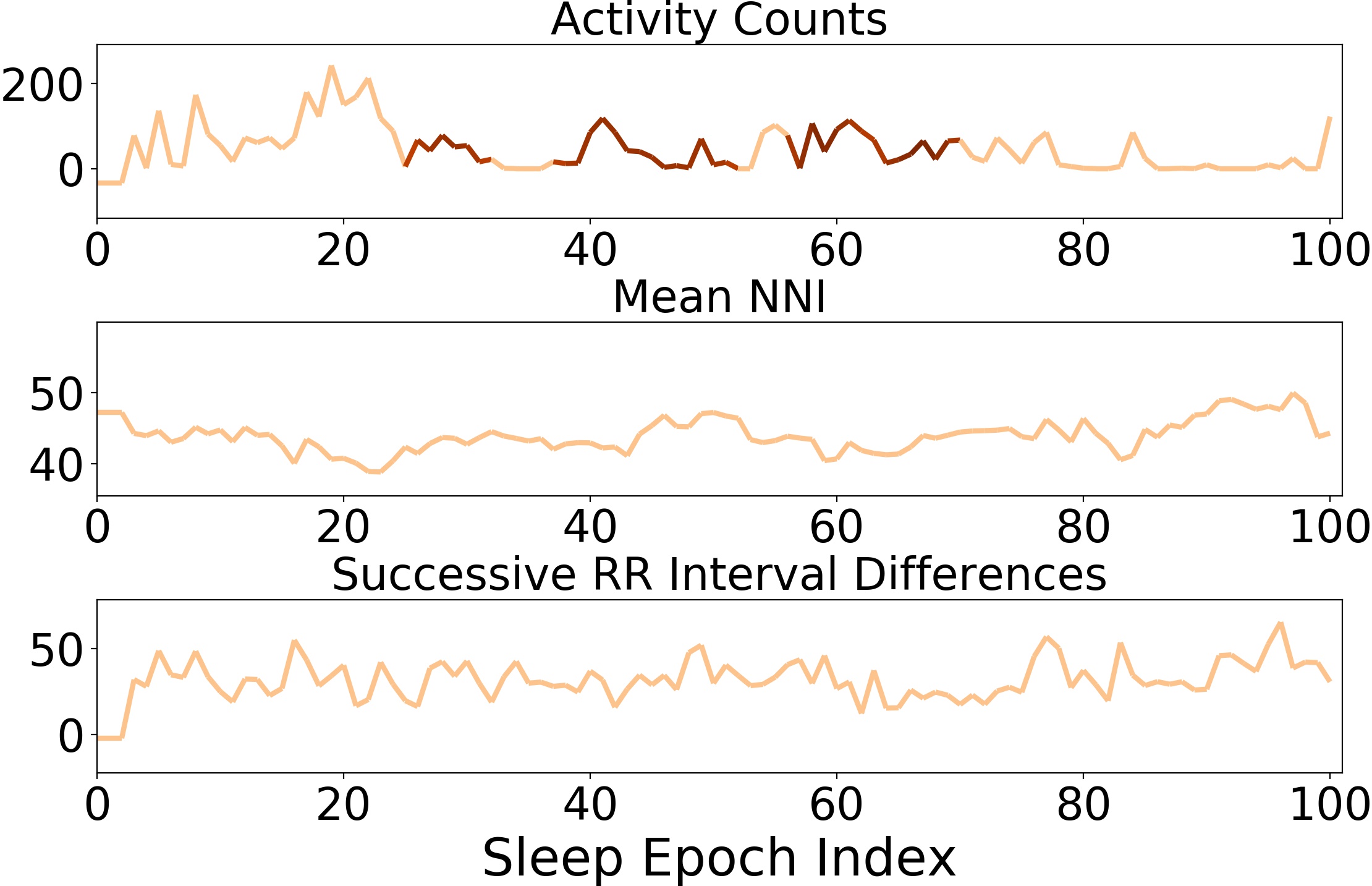}}%
\hfill 
\subcaptionbox{NREMS}{\includegraphics[width=0.333\textwidth]{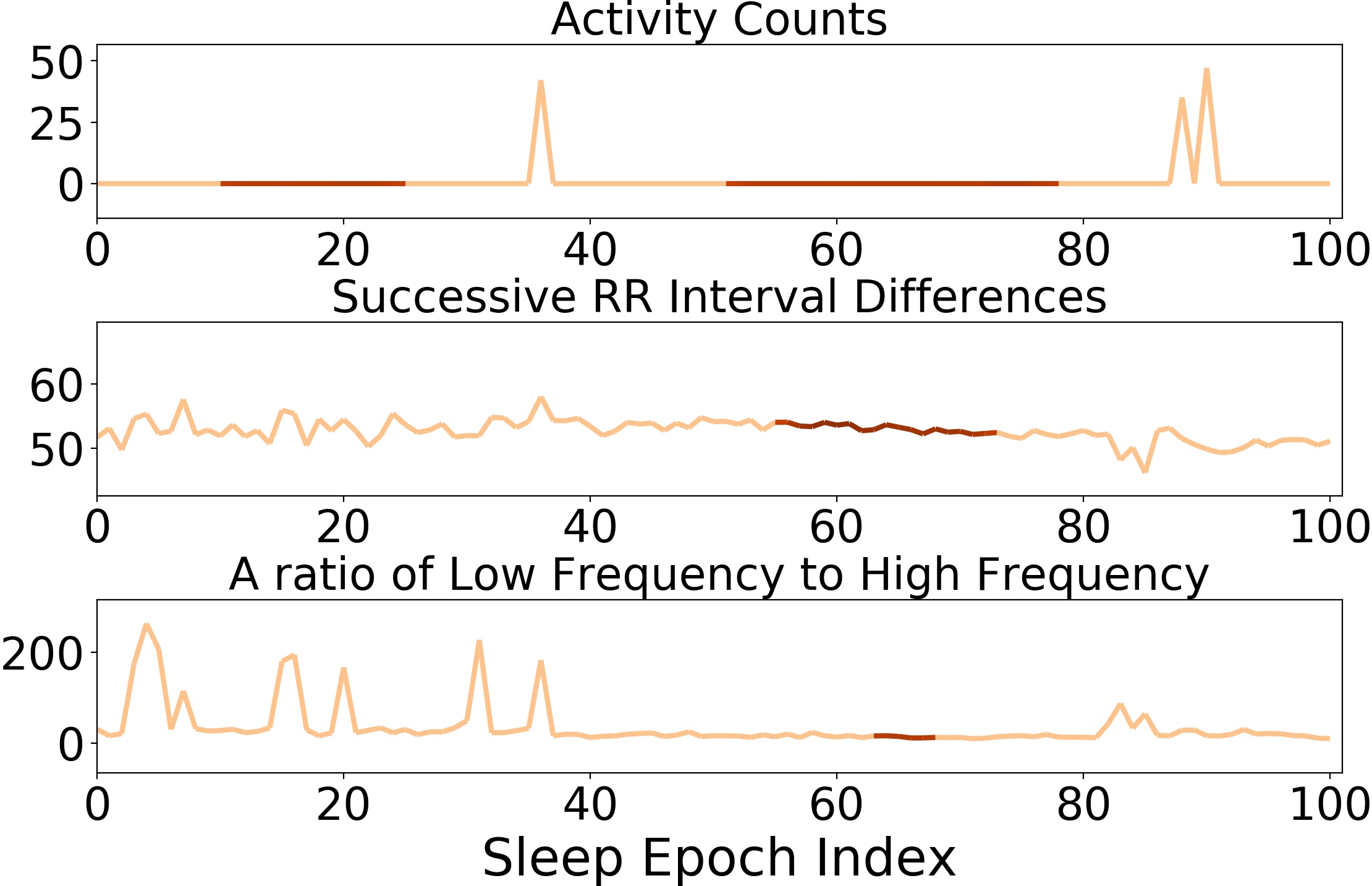}}%
\hfill 
\subcaptionbox{REMS}{\includegraphics[width=0.333\textwidth]{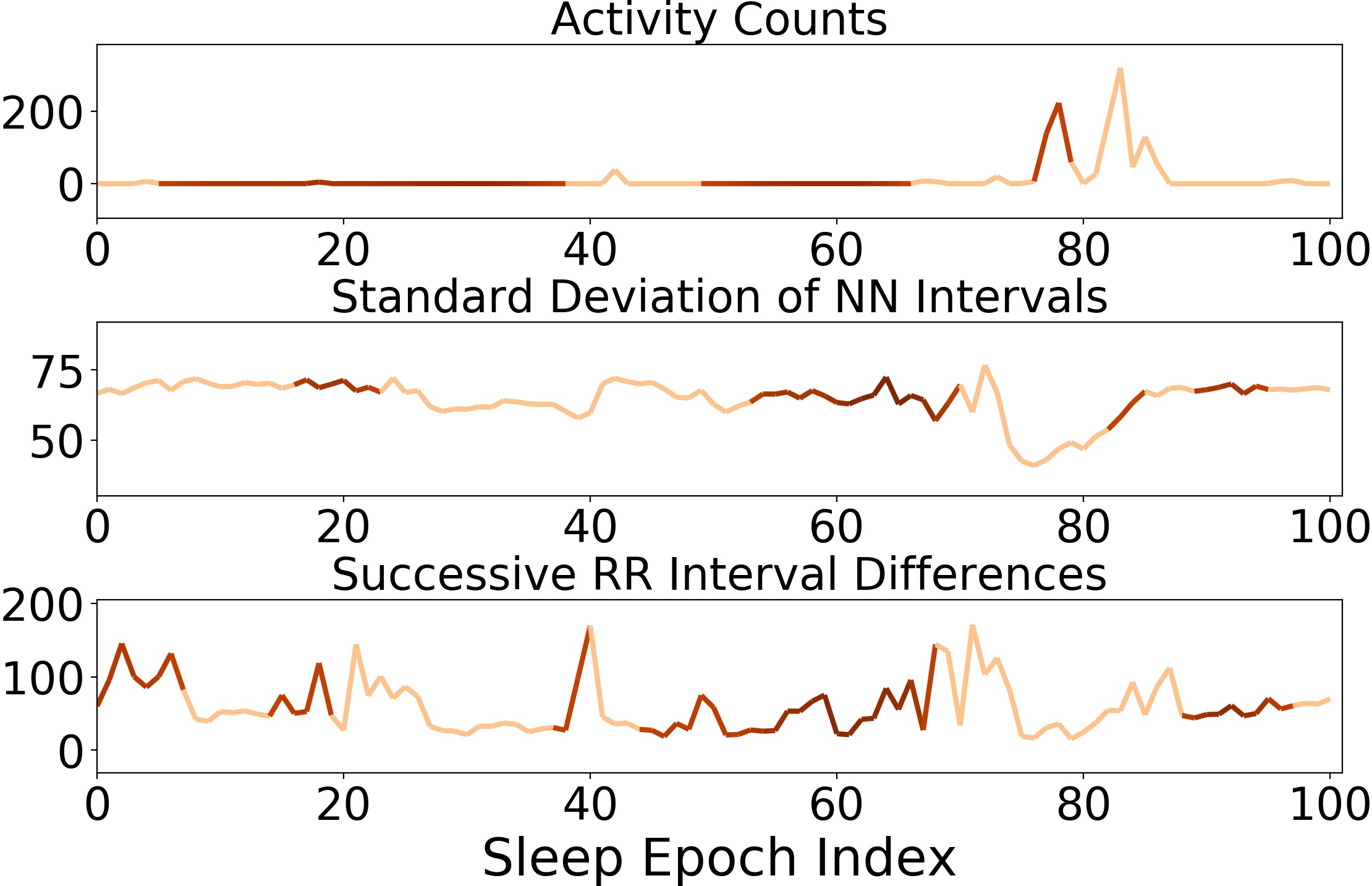}}%
\caption{The Grad-CAM plot of three selected examples from MESA dataset (ACT-HRV feature) using ResDeepCNN (Addition) in the late-stage fusion. Each row is the activation map for the input clinical features. To obtain a clear graph, the highlighted areas are the activation values over the threshold of 0.8 and the light color areas represents the activation values below the threshold of 0.8}
\label{fig:cam_plot}
\end{figure}

We retained the top three channels of each sleep stage to simplify the visualization as shown in Figure~\ref{fig:cam_plot}. To test whether this visualization is useful and can be understood by humans, we designed a game system\footnote{https://gradcamvisual1.azurewebsites.net/} that could conducted the exploratory study with users. The game system serves the purpose of engaging user to read and understand these visualization. The study consists of two phases, which investigate the accuracy of sleep stage classification by humans based solely on the input signals in the cases of Non-CAM and CAM visualization, respectively. In each phase, we encoded a continuous period of sleep data (intermediate feature data and hypnogram) for each sleep stage into videos to speed up the training process. 
In each phase, users would first watch the training videos; then they would be asked to recognize nine randomly selected sleep epochs that did not belong to the training videos. At the end of the test, users will be informed of their sleep stage classification accuracy. 
\begin{figure}[h!]
\centering
\subcaptionbox{}{\includegraphics[width=0.3\textwidth]{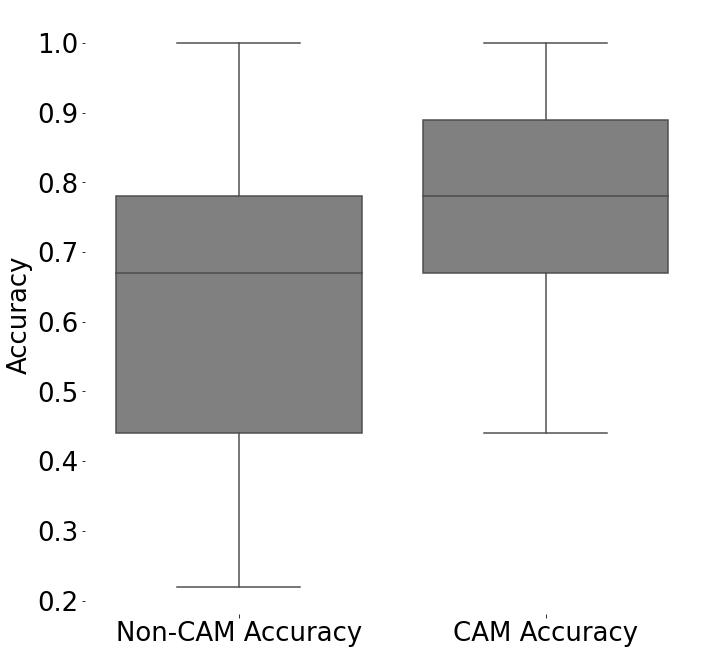}}%
\hfill 
\subcaptionbox{}{\includegraphics[width=0.3\textwidth]{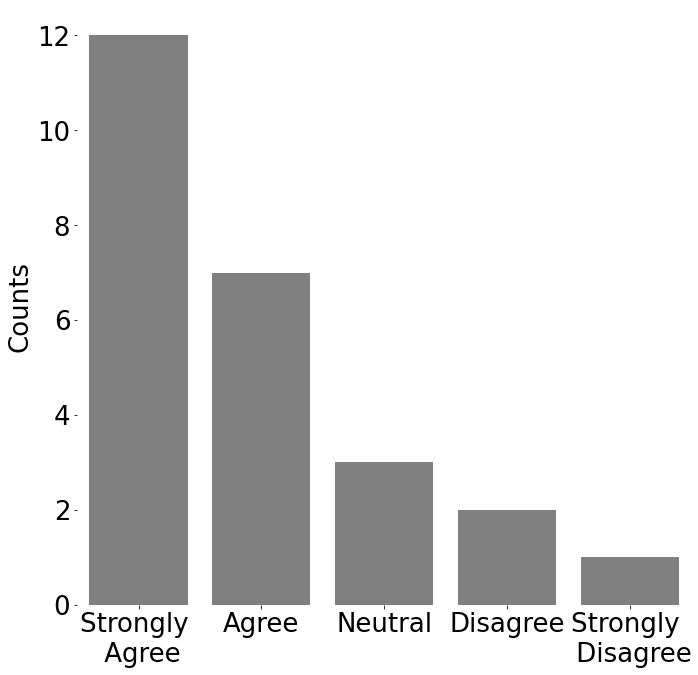}}%
\hfill 
\subcaptionbox{}{\includegraphics[width=0.3\textwidth]{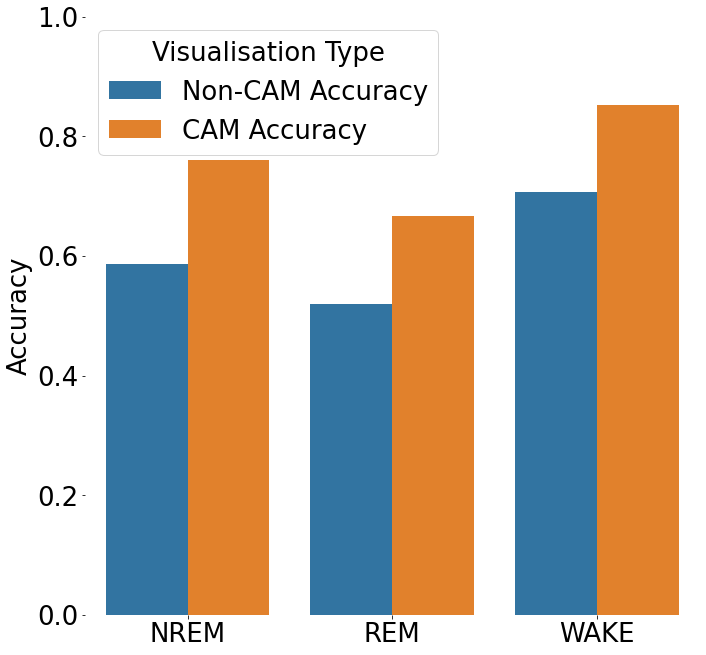}}%

\caption{ \textbf{(a)} A user study of three-stage classification accuracy calculated per participant-wise for Non-CAM and CAM assisted visualization.
\textbf{(b)} The answer of \textit{The machine-assisted visualization helped me to understand the difference between each sleep stage}.
\textbf{(c)} The breakdown of classification accuracy calculated for each sleep stage.
}
\label{fig:user_study}
\end{figure}

There are not established rules for sleep stage classification using movement and cardiac sensing data. As a pilot study, we recruited 25 individuals from Amazon’s Mechanical Turk. The task took, on average, 20-45 minutes to complete and participants were compensated USD 7.00. All procedures received ethical approval from the University’s ethical review board and the Research Ethics Committees (RECs). In total, we received 25 answers. Figure~\ref{fig:user_study} (a) shows the classification accuracy of CAM-assisted sleep stage classification is higher than the Non-CAM sleep stage classification. We further analyzed the results in detail by sleep stages in Figure~\ref{fig:user_study} (c). As can be seen, CAM-assisted visualizations improved human recognition accuracy on all sleep stages. To test whether the system can improve user's understanding of the visualization, we also designed a question a five point Likert scale. The results showed that the majority of participants either \textit{Strongly Agree} or \textit{Agree} that the machine-assisted visualization helped them to understand the difference between each sleep stage. 

\section{Discussion}

\subsection{Simple Fusion Method and Fusion Strategy} 
\subsubsection{Concatenation and Addition}

All three fusion strategies used the concatenation method.  In the early-stage fusion, the backbone network fused the latent features from each modality at every convolutional layer. All models studied in this paper surpassed previous studies, except for the models using the bilinear method. With the Apple Watch dataset, the skip connection numerically improved the performance for models in early-stage fusion, but the performance  with the MESA dataset decreased. A possible explanation for this might be that simply adding a skip connection may not benefit the model prediction performance in early-stage fusion when the training data is sufficient. 

In the late-stage fusion, the concatenation method numerically improved the prediction performance of all metrics for ResDeepCNN compared with the early-stage fusion. In terms of the concatenation method, parameter size and reasoning time increased by five times and six times, respectively, but the performance did not increase by that much. The addition method in late-stage fusion achieved the highest performance with the Apple Watch dataset. This may indicate the benefit of keeping latent features separate, and fusing them at a higher level might produce better results. Another possible explanation for this is the increase in network parameters.

For the hybrid fusion, we first fused cardiac intermediate features at the early stage and fused them with the movement sensing representation at the late stage. With the Apple Watch dataset, the simple operation method produced the same or better results compared with early-stage fusion. The methods in this category increased model parameters and inference time, but the classification performance hardly improved. The addition method aggregated the modal representation at the later stage, and similarly, it only obtained comparable results. A similar pattern was observed with the MESA dataset. These findings suggest that the effectiveness of simple operation methods may be affected by fusion strategies.

\subsection{Complex Fusion Method and Fusion Strategy}

Bilinear pooling turned out to be the weakest fusion method in the hybrid fusion strategy. This is a particularly interesting result, because the tensor-based methods showed improvements in the multimodal fusion literature such as with the task of visual question answering~\cite{Yu2017Multi-modalAnswering}.  It is possible that these results could be due to the failure to use the CNN network to learn about a post-bilinear latent feature. The model parameters were too large to be suitable for mobile computing scenarios. The cost of exploring the potential solutions exceeded the benefits.

With the attention mechanism, with the MESA dataset using the ACT-HRV feature setting, the mean F1 and Cohen's $\kappa$ of ResDeepCNN using the Attention-on-Mov method were statistically higher than the those in early-stage fusion. With the Apple Watch dataset, the same method can also produce comparable results to the highest performing method in the late-stage fusion, and the inference speed is three times faster. Moreover, the time deviation value was balanced in the prediction of each sleep stage.

Compared to the Attention-on-Car method, we observed a similar pattern with the MESA dataset, that is, applying attention weights to the latent features of movement produced higher results. It can therefore be assumed that the attention method improved the network’s ability to learn better representations that can benefit three-stage sleep classification by adjusting the weights of movement sensing representations, as it was difficult to discern REM sleep and NREM sleep using movement sensing alone.

\subsection{Model Selection}
The model parameters, model architecture, model inference time and model performance were the key considerations for the model selection in ubiquitous computing. In addition to these factors, the time deviation should also be considered.  It reflected the bias of the model prediction for each sleep stage. With imbalanced sleep data sets, biased models may constantly overestimate the duration of certain sleep stages and may lead to unreasonable health decisions. The mean and standard error of time deviation should be as close to zero as possible. In terms of inference time, predicting sleep data for a whole night, using the late fusion strategy was 6.2 times slower than when using the early-stage fusion strategy. As the model calculates the feature maps of each input channel individually using the convolution method, the time consumption of this method was related to the number of input intermediate features. In addition, for the design of the fusion strategy, we should also consider the ratio of module parameters to inference time. For instance, the parameters of the bilinear model were 77 times larger than the early-stage fusion models, but the speed was only three times slower. because most of the model parameters belonged to a fully connected layer in the bilinear module, which reduced the feature dimension of the feature matrix (the results of the matrix outer product).

From the results of Appendix~\ref{apx:raw_acc_feature_hrs_experiment}, the use of raw accelerometer data generated comparable results when compared to the use of handcraft features but with increased model parameters and inference time. It was considered a sub-optimal solution for long-term sleep stage monitoring.

In summary, if the movement and cardiac sensing data can be transmitted to the smartphone, the ResDeepCNN with the addition method in the late-stage fusion may be a feasible model for everyday use. Since it achieved the highest F1 score and with a balanced time deviation on each sleep stage. In the scenarios with limited computing resources such as smartwatches, using the early-stage fusion and the ResDeepCNN model may be a practical choice. The cost of using the late-stage fusion has increased by 9 time in inference time.

\subsection{Cross Dataset Comparison}
Based on activity counts and HRS features, the highest performing model for the MESA dataset achieved higher performance than the highest such model for the Apple Watch dataset.  Not only were the classification metrics of the MESA dataset higher than these of the Apple Watch dataset, but the standard error on the MESA dataset was smaller. It seems possible that these differences were due to two reasons. The first reason was that the Apple Watch dataset contained far fewer subjects compared with the MESA dataset. The second possible reason was the differences in data acquisition equipment and data pre-processing methods. The HR sensing module of Apple Watch dynamically calculated the HR data within two-five seconds, whereas the cardiac sensing used in the MESA dataset is IHR. The higher resolution IHR data might provide more discriminant information in order to discern three sleep stages.

\subsection{Exploratory Research of Visualization}

One of the objectives of this work is to better understand in what way the decision-making process of neural networks using multimodal fusion techniques on sleep stage classification can be understood by humans. Based on Grad-CAM scores, a simplified visualization method was adopted in this study and an exploratory study with users was performed. Our visualization tool can highlight both the key temporal signal segments and the most discriminant feature channels, and by providing important clues/patterns for users it can serve as an assistant tool in understanding different sleep stage signals.  

Compared with highlighting the temporal steps, the channel dimension reduction retained the minimum number of discriminant channels for sleep stage classification, which showed a combined reduction that could improve user understanding. Based on the repeated patterns of activity count and HRV features during a continuous sleep period, the results demonstrated that reducing information overload could improve human understanding on three-stage sleep recognition performance. The visualization increased users’ understanding in terms of the neural network decision making process to some extent.  

Wake recognition accuracy was higher than in the other two sleep stages, which indicating the wrist movement, is obvious to classify wake. This result is consistent with the known capability of actigraphy can be used to distinguish between wake and sleep. The highlighted patterns that appeared in the continuous sleep stage agreed with previous sleep physiology findings to some extent. 

The modeling process has window bias and Grad-CAM modeling bias. For example, this study adopted a window length of 101 (50.5 minutes). The highlighted areas will move backwards as the window moving forwards. The network is capable of locating signatures in a window that are meaningful for the current sleep stage recognition. This is very different to the annotation process using high-resolution (over 100Hz) PSG data. An interesting question for future work is to investigate whether these patterns have physiological meaning. 

Many existing interpretation and visualization techniques have been developed for visual data, yet it is unclear whether these methods are suitable for explaining sleep time-series data. This is a pilot study to investigate whether Grad-CAM applied to sleep time-series wearable data may be useful to humans. It is one of the mainstream methods used in visual and text data but is subject to the network structure. For instance, it is difficult to highlight the important areas in channels in early-stage fusion models without substantial changes to the method. On the other hand, it is difficult for humans to understand the highlighted patterns in the time dimension. 

We also observed the interesting results, e.g., 3 out 25 users experienced negative impacts on their understandings. In addition, the questionnaire data also highlighted that not everyone agreed that the system helped them to understand the difference in the patterns between sleep stages. A possible explanation for this might be that the visualization may not be understood by every person, or the training and testing process may be problematic. Future studies may consider conducting experiments with more detailed personalized questions.

The visualization was designed as a pilot study to understand the decision-making process of multimodal fusion for sleep stage classification. So, we did not investigate other mainstream interpretation methods, nor did we conduct large-scale user research. Future work may consider investigating the other interpretation methods such as SHAP, Anchor, etc. or even create a special algorithm for time-series data to facilitate intuitive understanding by humans.

\subsection{Comparison with Previous Work and Implications}
\begin{figure} [h!]
\centering
\begin{tabular}{cccc}
\includegraphics[width=0.3\textwidth]{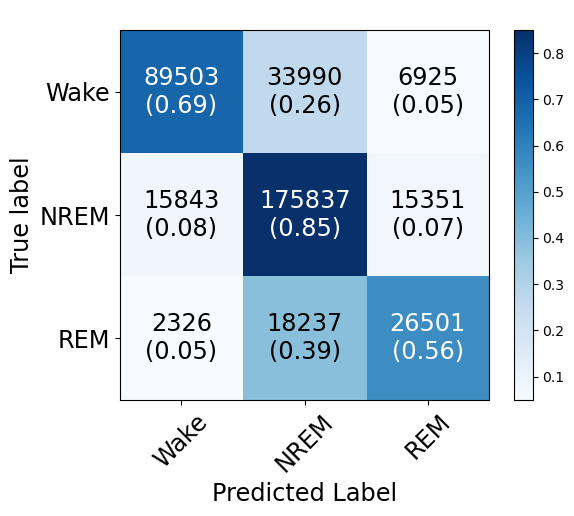} &
\includegraphics[width=0.3\textwidth]{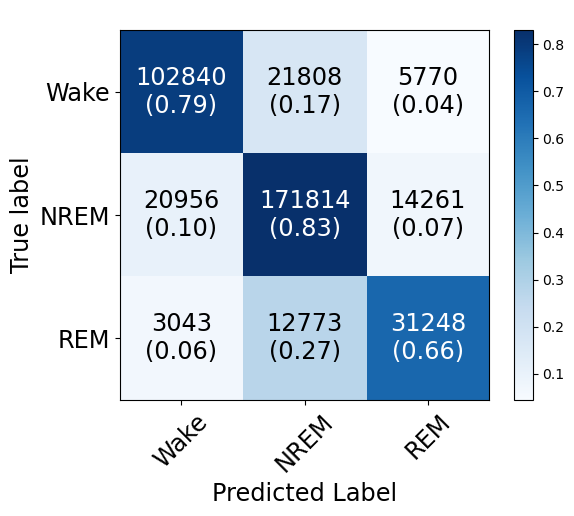}  \\
\textbf{(a)}  & \textbf{(b)}  \\[6pt]
\end{tabular}
\begin{tabular}{cccc}
\includegraphics[width=0.3\textwidth]{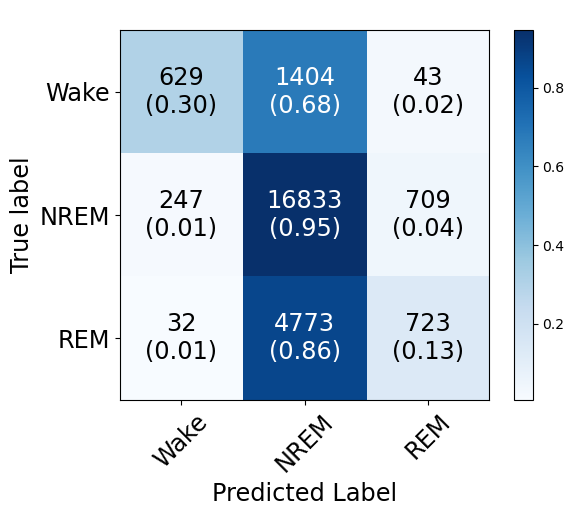} &
\includegraphics[width=0.3\textwidth]{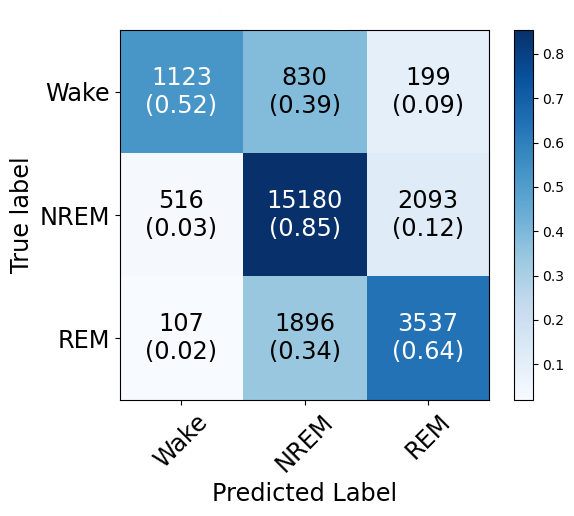} &
\includegraphics[width=0.3\textwidth]{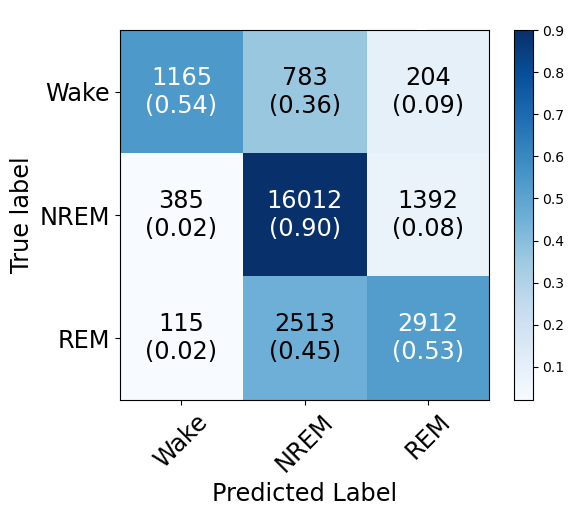} \\
\textbf{(c)}  & \textbf{(d)}   & \textbf{(e)} \\[6pt]
\end{tabular}
\caption{ \textbf{(a)} The CNN(101)  and early-stage fusion based on the MESA (ACT-HRV) dataset used in~\cite{Zhai2020MakingSensing}.
\textbf{(b)} Hybrid fusion using ResDeepCNN (Attention-on-Act) based on the MESA (ACT-HRV) dataset
\textbf{(c)} Walch et al. using multiple layer perception based on activity counts, HR and circadian time ~\cite{Walch2019SleepDevice} 
\textbf{(d)} Late-stage fusion using ResDeepCNN (Addition) based on the Apple Watch Dataset
\textbf{(e)} Late-stage fusion using ResDeepMixCNN (Concatenation) using raw accelerometer data and HRS based on the Apple Watch dataset.
}
\label{fig:confusion_matrix_2}
\end{figure}

\begin{table}[htbp]
  \centering
  \caption{
  Three-stage sleep classification prediction results compared with previous work evaluated at subject level  (mean $\pm$ standard error at 95\% confidence interval)  during the recording period.
}
  \resizebox{\textwidth}{!}{
\begin{tabular}{c|c|c|c|c|c|c|c|c}
\toprule
\multicolumn{3}{c|}{\textbf{Fusion Specifics}} & \multicolumn{3}{c|}{\textbf{Performance Metrics}} & \multicolumn{3}{c}{\textbf{Time Deviation (min.)}} \\
\midrule
\textbf{Dataset and Feature Set} & \textbf{Fusion Stage} & \textbf{Model} & \textbf{Accuracy (\%)} & \textbf{Cohen's $\kappa$} & \textbf{Mean F1 (\%)} & \textbf{Non-REM sleep} & \textbf{REM sleep} & {\textbf{Wake}} \\
\midrule
\multirow{2}[2]{*}{MESA (ACT-HRV)} & Early-stage Fusion & CNN (101) (Zhai, 2020) & 76.0 $\pm$ 1.0 & 58.6 $\pm$ 1.9 & 68.1 $\pm$ 1.3 & 14.9 $\pm$ 6.7 & -0.5 $\pm$ 4.3 & -14.4 $\pm$ 5.8 \\
  & Hybrid Fusion & \shortstack{ ResDeepCNN \\( Attention-on-Mov)} & \boldmath{}\textbf{79.8 $\pm$ 0.9}\unboldmath{} & \boldmath{}\textbf{65.5 $\pm$ 1.8}\unboldmath{} & \boldmath{}\textbf{73.3 $\pm$ 1.3}\unboldmath{} & -0.9 $\pm$ 6.4 & 6.1 $\pm$ 3.7 & -5.1 $\pm$ 6.3 \\
\midrule
\shortstack{Apple Watch \\(Activity Counts, HR, Time)} & Early-stage Fusion & MLP (Walch, 2019) & 72.1 $\pm$ 2.4 & 23.7 $\pm$ 4.4 & 47.8 $\pm$ 3.6 & 84.2 $\pm$ 17.2 & -65.4 $\pm$ 15.0 & -18.8 $\pm$ 6.1 \\
\midrule
\shortstack{Apple Watch\\ (Activity Counts, HRS)} & Late-stage Fusion & \shortstack{ResDeepCNN \\ Addition} & \boldmath{}\textbf{78.2 $\pm$ 2.3}\unboldmath{} & \boldmath{}\textbf{52.0 $\pm$ 5.8}\unboldmath{} & \boldmath{}\textbf{66.5 $\pm$ 3.8}\unboldmath{} & 1.9 $\pm$ 11.7 & 4.7 $\pm$ 12.6 & -6.5 $\pm$ 7.3 \\
\bottomrule
\end{tabular}%

}
  \label{tab:compare_rp}%
\end{table}%

Table \ref{tab:compare_rp} and  Figure~\ref{fig:confusion_matrix_2} show the results compared with previous works. We have observed that the use of multimodal fusion strategies and fusion methods can improve model prediction performance. With the three-stage sleep classification dataset, the class imbalance issue causes the classifier to be biased towards the majority class, which is NREM sleep. 

To compare with the previous work, we conducted ten runs of the model with the highest mean F1 and the baseline model for each dataset, respectively. Each run used a different random number seed. We performed a $t$-test on the accuracy, Cohen's $\kappa$ and mean F1 score. Compared with the previous work\cite{Zhai2020MakingSensing} with the MESA dataset, the accuracy ($p$<0.001), Cohen's $\kappa$ ($p$<0.001) and mean F1 score ($p$<0.001) improved statistically significantly with the MESA dataset using the ACT-HRV feature set. With the Apple Watch dataset, the accuracy ($p$<0.001), mean F1 score ($p$<0.001) and Cohen's $\kappa$ ($p$<0.001) were also statistically higher than previous work~\cite{Walch2019SleepDevice}. 
These improvements suggest that the proper multimodal fusion strategy and method can improve the robustness of the model, which is a step towards automated three-stage sleep classification. The findings reported here suggest that reasonable performance may be achieved using the movement and cardiac features derived from consumer/research-grade wearable devices. 
\section{Conclusion}
Using actigraphy to monitoring sleep-wake has existed for many decades. But, for sleep stage classification, we have relied on the PSG study, which is an expensive, burdensome, laboratory sleep monitoring method. This limits many research advances in sleep and health. In recent years,  more and more new products using ubiquitous computing technology have been passed FDA clearance, such as the Apple Watch irregular heart rate detection function~\cite{Perez2019Large-ScaleFibrillation}. The achievements of these wearables provide important instrumental tools for study of longitudinal sleep and health. The core contribution of this work lies in our systematic study on how to better integrate multi-modal data to monitor three-stage sleep that may use consumer/research grade wearables. Through our study, the performance of several new models exceeded the previous benchmark studies significantly. We have provided a new multimodal fusion benchmark for the ubiquitous computing community. This has provided the potential for the use of consumer wearables to study the sleep health of large-scale populations in the future. One of the motivations for this work was to respond to previous research that called for more accurate and transparent sleep stage sensing algorithms on consumer wearable devices~\cite{Walch2019SleepDevice}. We hope this work will encourage more researchers, consumers, and application developers to use consumer/research grade wearables to study and understand sleep and health.
\begin{acks}
This research is funded through the EPSRC Centre for Doctoral Training in Digital Civics (EP/L016176/1). The Multi-Ethnic Study of Atherosclerosis (MESA) Sleep Ancillary study was funded by NIH-NHLBI Association of Sleep Disorders with Cardiovascular Health Across Ethnic Groups (RO1 HL098433). MESA is supported by NHLBI funded contracts HHSN268201500003I, N01-HC-95159, N01-HC-95160, N01-HC-95161, N01-HC-95162, N01-HC-95163, N01-HC-95164, N01-HC-95165, N01-HC-95166, N01-HC-95167, N01-HC-95168 and N01-HC-95169 from the National Heart, Lung, and Blood Institute, and by cooperative agreements UL1-TR-000040, UL1-TR-001079, and UL1-TR-001420 funded by NCATS. The National Sleep Research Resource was supported by the National Heart, Lung, and Blood Institute (R24 HL114473, 75N92019R002).
Thanks Duan Haoran, Ouyang Zizhou, Shao Shuai, Dr. Kirstie Anderson, Becky Zhu for their suggestions on model selection, experimental setup, and visualization, and sharing sleep knowledge with me.
\end{acks}

\bibliographystyle{ACM-Reference-Format}
\bibliography{camera_ready}


\begin{thebibliography}{86}


\ifx \showCODEN    \undefined \def \showCODEN     #1{\unskip}     \fi
\ifx \showDOI      \undefined \def \showDOI       #1{#1}\fi
\ifx \showISBNx    \undefined \def \showISBNx     #1{\unskip}     \fi
\ifx \showISBNxiii \undefined \def \showISBNxiii  #1{\unskip}     \fi
\ifx \showISSN     \undefined \def \showISSN      #1{\unskip}     \fi
\ifx \showLCCN     \undefined \def \showLCCN      #1{\unskip}     \fi
\ifx \shownote     \undefined \def \shownote      #1{#1}          \fi
\ifx \showarticletitle \undefined \def \showarticletitle #1{#1}   \fi
\ifx \showURL      \undefined \def \showURL       {\relax}        \fi
\providecommand\bibfield[2]{#2}
\providecommand\bibinfo[2]{#2}
\providecommand\natexlab[1]{#1}
\providecommand\showeprint[2][]{arXiv:#2}

\bibitem[\protect\citeauthoryear{Adadi and Berrada}{Adadi and Berrada}{2018}]%
        {Adadi2018PeekingXAI}
\bibfield{author}{\bibinfo{person}{Amina Adadi} {and} \bibinfo{person}{Mohammed
  Berrada}.} \bibinfo{year}{2018}\natexlab{}.
\newblock \showarticletitle{{Peeking Inside the Black-Box: A Survey on
  Explainable Artificial Intelligence (XAI)}}.
\newblock \bibinfo{journal}{\emph{IEEE Access}}  \bibinfo{volume}{6}
  (\bibinfo{date}{9} \bibinfo{year}{2018}), \bibinfo{pages}{52138--52160}.
\newblock
\showISSN{21693536}
\urldef\tempurl%
\url{https://doi.org/10.1109/ACCESS.2018.2870052}
\showDOI{\tempurl}


\bibitem[\protect\citeauthoryear{Afouras, Chung, Senior, Vinyals, and
  Zisserman}{Afouras et~al\mbox{.}}{2018}]%
        {Afouras2018DeepRecognition}
\bibfield{author}{\bibinfo{person}{Triantafyllos Afouras},
  \bibinfo{person}{Joon~Son Chung}, \bibinfo{person}{Andrew Senior},
  \bibinfo{person}{Oriol Vinyals}, {and} \bibinfo{person}{Andrew Zisserman}.}
  \bibinfo{year}{2018}\natexlab{}.
\newblock \showarticletitle{{Deep Audio-visual Speech Recognition}}.
\newblock \bibinfo{journal}{\emph{IEEE Transactions on Pattern Analysis and
  Machine Intelligence}} (\bibinfo{year}{2018}).
\newblock
\showISSN{19393539}
\urldef\tempurl%
\url{https://doi.org/10.1109/TPAMI.2018.2889052}
\showDOI{\tempurl}


\bibitem[\protect\citeauthoryear{Agargun and Cartwright}{Agargun and
  Cartwright}{2003}]%
        {Agargun2003REMPatients}
\bibfield{author}{\bibinfo{person}{Mehmet~Y. Agargun} {and}
  \bibinfo{person}{Rosalind Cartwright}.} \bibinfo{year}{2003}\natexlab{}.
\newblock \showarticletitle{{REM sleep, dream variables and suicidality in
  depressed patients}}.
\newblock \bibinfo{journal}{\emph{Psychiatry Research}} \bibinfo{volume}{119},
  \bibinfo{number}{1-2} (\bibinfo{date}{7} \bibinfo{year}{2003}),
  \bibinfo{pages}{33--39}.
\newblock
\showISSN{01651781}
\urldef\tempurl%
\url{https://doi.org/10.1016/S0165-1781(03)00111-2}
\showDOI{\tempurl}


\bibitem[\protect\citeauthoryear{Ancoli-Israel, Cole, Alessi, Chambers,
  Moorcroft, and Pollak}{Ancoli-Israel et~al\mbox{.}}{2003}]%
        {Ancoli-Israel2003TheRhythms}
\bibfield{author}{\bibinfo{person}{Sonia Ancoli-Israel}, \bibinfo{person}{Roger
  Cole}, \bibinfo{person}{Cathy Alessi}, \bibinfo{person}{Mark Chambers},
  \bibinfo{person}{William Moorcroft}, {and} \bibinfo{person}{Charles~P.
  Pollak}.} \bibinfo{year}{2003}\natexlab{}.
\newblock \bibinfo{title}{{The role of actigraphy in the study of sleep and
  circadian rhythms}}.
\newblock , \bibinfo{numpages}{342--392}~pages.
\newblock
\showISSN{01618105}
\urldef\tempurl%
\url{https://doi.org/10.1093/sleep/26.3.342}
\showDOI{\tempurl}


\bibitem[\protect\citeauthoryear{Bahdanau, Cho, and Bengio}{Bahdanau
  et~al\mbox{.}}{2015}]%
        {Bahdanau2015NeuralTranslate}
\bibfield{author}{\bibinfo{person}{Dzmitry Bahdanau},
  \bibinfo{person}{Kyung~Hyun Cho}, {and} \bibinfo{person}{Yoshua Bengio}.}
  \bibinfo{year}{2015}\natexlab{}.
\newblock \showarticletitle{{Neural machine translation by jointly learning to
  align and translate}}. In \bibinfo{booktitle}{\emph{3rd International
  Conference on Learning Representations, ICLR 2015 - Conference Track
  Proceedings}}. \bibinfo{publisher}{International Conference on Learning
  Representations, ICLR}.
\newblock


\bibitem[\protect\citeauthoryear{Bai, Guan, and Ng}{Bai et~al\mbox{.}}{2020}]%
        {Bai2020FatigueSensors}
\bibfield{author}{\bibinfo{person}{Yang Bai}, \bibinfo{person}{Yu Guan}, {and}
  \bibinfo{person}{Wan~Fai Ng}.} \bibinfo{year}{2020}\natexlab{}.
\newblock \showarticletitle{{Fatigue assessment using ECG and actigraphy
  sensors}}. In \bibinfo{booktitle}{\emph{Proceedings - International Symposium
  on Wearable Computers, ISWC}}. \bibinfo{publisher}{Association for Computing
  Machinery}, \bibinfo{pages}{12--16}.
\newblock
\showISBNx{9781450380775}
\showISSN{15504816}
\urldef\tempurl%
\url{https://doi.org/10.1145/3410531.3414308}
\showDOI{\tempurl}


\bibitem[\protect\citeauthoryear{Ben~Simon, Rossi, Harvey, and
  Walker}{Ben~Simon et~al\mbox{.}}{2020}]%
        {BenSimon2020OveranxiousUnderslept}
\bibfield{author}{\bibinfo{person}{Eti Ben~Simon}, \bibinfo{person}{Aubrey
  Rossi}, \bibinfo{person}{Allison~G. Harvey}, {and}
  \bibinfo{person}{Matthew~P. Walker}.} \bibinfo{year}{2020}\natexlab{}.
\newblock \showarticletitle{{Overanxious and underslept}}.
\newblock \bibinfo{journal}{\emph{Nature Human Behaviour}} \bibinfo{volume}{4},
  \bibinfo{number}{1} (\bibinfo{date}{1} \bibinfo{year}{2020}),
  \bibinfo{pages}{100--110}.
\newblock
\showISSN{23973374}
\urldef\tempurl%
\url{https://doi.org/10.1038/s41562-019-0754-8}
\showDOI{\tempurl}


\bibitem[\protect\citeauthoryear{Berry}{Berry}{2012}]%
        {Berry2012FundamentalsMedicine}
\bibfield{author}{\bibinfo{person}{Richard Berry}.}
  \bibinfo{year}{2012}\natexlab{}.
\newblock \bibinfo{booktitle}{\emph{{Fundamentals of Sleep Medicine}}}.
\newblock \bibinfo{publisher}{Elsevier Inc.}
\newblock
\showISBNx{9781437703269}
\urldef\tempurl%
\url{https://doi.org/10.1016/C2009-0-38997-7}
\showDOI{\tempurl}


\bibitem[\protect\citeauthoryear{Berry, Brooks, Gamaldo, Harding, Lloyd, Quan,
  Troester, and Vaughn}{Berry et~al\mbox{.}}{2017}]%
        {Berry2017AASM2.4}
\bibfield{author}{\bibinfo{person}{Richard~B. Berry}, \bibinfo{person}{Rita
  Brooks}, \bibinfo{person}{Charlene Gamaldo}, \bibinfo{person}{Susan~M.
  Harding}, \bibinfo{person}{Robin~M. Lloyd}, \bibinfo{person}{Stuart~F. Quan},
  \bibinfo{person}{Matthew~T. Troester}, {and} \bibinfo{person}{Bradley~V.
  Vaughn}.} \bibinfo{year}{2017}\natexlab{}.
\newblock \bibinfo{title}{{AASM scoring manual updates for 2017 (version
  2.4)}}.
\newblock , \bibinfo{numpages}{665--666}~pages.
\newblock
\showISSN{15509397}
\urldef\tempurl%
\url{https://doi.org/10.5664/jcsm.6576}
\showDOI{\tempurl}


\bibitem[\protect\citeauthoryear{Besedovsky, Lange, and Born}{Besedovsky
  et~al\mbox{.}}{2012}]%
        {Besedovsky2012SleepFunction}
\bibfield{author}{\bibinfo{person}{Luciana Besedovsky}, \bibinfo{person}{Tanja
  Lange}, {and} \bibinfo{person}{Jan Born}.} \bibinfo{year}{2012}\natexlab{}.
\newblock \bibinfo{title}{{Sleep and immune function}}.
\newblock , \bibinfo{numpages}{121--137}~pages.
\newblock
\showISSN{00316768}
\urldef\tempurl%
\url{https://doi.org/10.1007/s00424-011-1044-0}
\showDOI{\tempurl}


\bibitem[\protect\citeauthoryear{Bleichner, Lundbeck, Selisky, Minow,
  J{\"{a}}ger, Emkes, Debener, and De~Vos}{Bleichner et~al\mbox{.}}{2015}]%
        {Bleichner2015ExploringSee}
\bibfield{author}{\bibinfo{person}{Martin~G. Bleichner}, \bibinfo{person}{Micha
  Lundbeck}, \bibinfo{person}{Matthias Selisky}, \bibinfo{person}{Falk Minow},
  \bibinfo{person}{Manuela J{\"{a}}ger}, \bibinfo{person}{Reiner Emkes},
  \bibinfo{person}{Stefan Debener}, {and} \bibinfo{person}{Maarten De~Vos}.}
  \bibinfo{year}{2015}\natexlab{}.
\newblock \showarticletitle{{Exploring miniaturized EEG electrodes for
  brain-computer interfaces. An EEG you do not see?}}
\newblock \bibinfo{journal}{\emph{Physiological Reports}} \bibinfo{volume}{3},
  \bibinfo{number}{4} (\bibinfo{year}{2015}).
\newblock
\showISSN{2051817X}
\urldef\tempurl%
\url{https://doi.org/10.14814/phy2.12362}
\showDOI{\tempurl}


\bibitem[\protect\citeauthoryear{Boudreau, Yeh, Dumont, and Boivin}{Boudreau
  et~al\mbox{.}}{2013}]%
        {Boudreau2013CircadianStages}
\bibfield{author}{\bibinfo{person}{Philippe Boudreau},
  \bibinfo{person}{Wei~Hsien Yeh}, \bibinfo{person}{Guy~A. Dumont}, {and}
  \bibinfo{person}{Diane~B. Boivin}.} \bibinfo{year}{2013}\natexlab{}.
\newblock \showarticletitle{{Circadian variation of heart rate variability
  across sleep stages}}.
\newblock \bibinfo{journal}{\emph{Sleep}} \bibinfo{volume}{36},
  \bibinfo{number}{12} (\bibinfo{date}{12} \bibinfo{year}{2013}),
  \bibinfo{pages}{1919--1928}.
\newblock
\showISSN{01618105}
\urldef\tempurl%
\url{https://doi.org/10.5665/sleep.3230}
\showDOI{\tempurl}


\bibitem[\protect\citeauthoryear{Cabiddu, Cerutti, Viardot, Werner, and
  Bianchi}{Cabiddu et~al\mbox{.}}{2012}]%
        {Cabiddu2012ModulationRespiration}
\bibfield{author}{\bibinfo{person}{Ramona Cabiddu}, \bibinfo{person}{Sergio
  Cerutti}, \bibinfo{person}{Geoffrey Viardot}, \bibinfo{person}{Sandra
  Werner}, {and} \bibinfo{person}{Anna~M. Bianchi}.}
  \bibinfo{year}{2012}\natexlab{}.
\newblock \showarticletitle{{Modulation of the sympatho-vagal balance during
  sleep: Frequency domain study of heart rate variability and respiration}}.
\newblock \bibinfo{journal}{\emph{Frontiers in Physiology}}  \bibinfo{volume}{3
  MAR} (\bibinfo{year}{2012}).
\newblock
\showISSN{1664042X}
\urldef\tempurl%
\url{https://doi.org/10.3389/fphys.2012.00045}
\showDOI{\tempurl}


\bibitem[\protect\citeauthoryear{Chen, Wang, Zee, Lutsey, Javaheri,
  Alc{\'{a}}ntara, Jackson, Williams, and Redline}{Chen et~al\mbox{.}}{2015}]%
        {Chen2015Racial/ethnicMESA}
\bibfield{author}{\bibinfo{person}{Xiaoli Chen}, \bibinfo{person}{Rui Wang},
  \bibinfo{person}{Phyllis Zee}, \bibinfo{person}{Pamela~L. Lutsey},
  \bibinfo{person}{Sogol Javaheri}, \bibinfo{person}{Carmela Alc{\'{a}}ntara},
  \bibinfo{person}{Chandra~L. Jackson}, \bibinfo{person}{Michelle~A. Williams},
  {and} \bibinfo{person}{Susan Redline}.} \bibinfo{year}{2015}\natexlab{}.
\newblock \showarticletitle{{Racial/ethnic differences in sleep disturbances:
  The Multi-Ethnic Study of Atherosclerosis (MESA)}}.
\newblock \bibinfo{journal}{\emph{Sleep}} \bibinfo{volume}{38},
  \bibinfo{number}{6} (\bibinfo{date}{6} \bibinfo{year}{2015}),
  \bibinfo{pages}{877--888}.
\newblock
\showISSN{15509109}
\urldef\tempurl%
\url{https://doi.org/10.5665/sleep.4732}
\showDOI{\tempurl}


\bibitem[\protect\citeauthoryear{Chen, Dong, Gao, Liu, and Gu}{Chen
  et~al\mbox{.}}{2017}]%
        {Chen2017Rapid}
\bibfield{author}{\bibinfo{person}{Yuanying Chen}, \bibinfo{person}{Wei Dong},
  \bibinfo{person}{Yi Gao}, \bibinfo{person}{Xue Liu}, {and}
  \bibinfo{person}{Tao Gu}.} \bibinfo{year}{2017}\natexlab{}.
\newblock \showarticletitle{{Rapid}}.
\newblock \bibinfo{journal}{\emph{Proceedings of the ACM on Interactive,
  Mobile, Wearable and Ubiquitous Technologies}} \bibinfo{volume}{1},
  \bibinfo{number}{3} (\bibinfo{date}{9} \bibinfo{year}{2017}),
  \bibinfo{pages}{1--27}.
\newblock
\showISSN{2474-9567}
\urldef\tempurl%
\url{https://doi.org/10.1145/3130906}
\showDOI{\tempurl}


\bibitem[\protect\citeauthoryear{Chouchou and Desseilles}{Chouchou and
  Desseilles}{2014}]%
        {Chouchou2014HeartBrain}
\bibfield{author}{\bibinfo{person}{Florian Chouchou} {and}
  \bibinfo{person}{Martin Desseilles}.} \bibinfo{year}{2014}\natexlab{}.
\newblock \bibinfo{title}{{Heart rate variability: A tool to explore the
  sleeping brain?}}
\newblock , \bibinfo{numpages}{402}~pages.
\newblock
\showISSN{1662453X}
\urldef\tempurl%
\url{https://doi.org/10.3389/fnins.2014.00402}
\showDOI{\tempurl}


\bibitem[\protect\citeauthoryear{De~Vos and Debener}{De~Vos and
  Debener}{2014}]%
        {DeVos2014MobileCognition}
\bibfield{author}{\bibinfo{person}{Maarten De~Vos} {and}
  \bibinfo{person}{Stefan Debener}.} \bibinfo{year}{2014}\natexlab{}.
\newblock \bibinfo{title}{{Mobile eeg: Towards brain activity monitoring during
  natural action and cognition}}.
\newblock , \bibinfo{numpages}{2}~pages.
\newblock
\showISSN{01678760}
\urldef\tempurl%
\url{https://doi.org/10.1016/j.ijpsycho.2013.10.008}
\showDOI{\tempurl}


\bibitem[\protect\citeauthoryear{Duan, Wang, and Guan}{Duan
  et~al\mbox{.}}{2020}]%
        {Duan2020SOFA-Net:Counting}
\bibfield{author}{\bibinfo{person}{Haoran Duan}, \bibinfo{person}{Shidong
  Wang}, {and} \bibinfo{person}{Yu Guan}.} \bibinfo{year}{2020}\natexlab{}.
\newblock \showarticletitle{{SOFA-Net: Second-Order and First-order Attention
  Network for Crowd Counting}}.
\newblock  (\bibinfo{date}{8} \bibinfo{year}{2020}).
\newblock
\urldef\tempurl%
\url{https://arxiv.org/abs/2008.03723v1}
\showURL{%
\tempurl}


\bibitem[\protect\citeauthoryear{E, R, DA, and E}{E et~al\mbox{.}}{2008}]%
        {E2008Slow-waveHumans}
\bibfield{author}{\bibinfo{person}{Tasali E}, \bibinfo{person}{Leproult R},
  \bibinfo{person}{Ehrmann DA}, {and} \bibinfo{person}{Van~Cauter E}.}
  \bibinfo{year}{2008}\natexlab{}.
\newblock \showarticletitle{{Slow-wave sleep and the risk of type 2 diabetes in
  humans}}.
\newblock \bibinfo{journal}{\emph{Proceedings of the National Academy of
  Sciences of the United States of America}} \bibinfo{volume}{105},
  \bibinfo{number}{3} (\bibinfo{date}{1} \bibinfo{year}{2008}),
  \bibinfo{pages}{1044--1049}.
\newblock
\showISSN{1091-6490}
\urldef\tempurl%
\url{https://doi.org/10.1073/PNAS.0706446105}
\showDOI{\tempurl}


\bibitem[\protect\citeauthoryear{Fonseca, Den~Teuling, Long, and Aarts}{Fonseca
  et~al\mbox{.}}{2017}]%
        {Fonseca2017CardiorespiratoryFields}
\bibfield{author}{\bibinfo{person}{Pedro Fonseca}, \bibinfo{person}{Niek
  Den~Teuling}, \bibinfo{person}{Xi Long}, {and} \bibinfo{person}{Ronald~M.
  Aarts}.} \bibinfo{year}{2017}\natexlab{}.
\newblock \showarticletitle{{Cardiorespiratory Sleep Stage Detection Using
  Conditional Random Fields}}.
\newblock \bibinfo{journal}{\emph{IEEE Journal of Biomedical and Health
  Informatics}} \bibinfo{volume}{21}, \bibinfo{number}{4}
  (\bibinfo{year}{2017}), \bibinfo{pages}{956--966}.
\newblock
\showISSN{21682194}
\urldef\tempurl%
\url{https://doi.org/10.1109/JBHI.2016.2550104}
\showDOI{\tempurl}


\bibitem[\protect\citeauthoryear{Fonseca, Den~Teuling, Long, and Aarts}{Fonseca
  et~al\mbox{.}}{2018}]%
        {Fonseca2018AClassification}
\bibfield{author}{\bibinfo{person}{Pedro Fonseca}, \bibinfo{person}{Niek
  Den~Teuling}, \bibinfo{person}{Xi Long}, {and} \bibinfo{person}{Ronald~M.
  Aarts}.} \bibinfo{year}{2018}\natexlab{}.
\newblock \showarticletitle{{A comparison of probabilistic classifiers for
  sleep stage classification}}.
\newblock \bibinfo{journal}{\emph{Physiological Measurement}}
  \bibinfo{volume}{39}, \bibinfo{number}{5} (\bibinfo{year}{2018}).
\newblock
\showISSN{13616579}
\urldef\tempurl%
\url{https://doi.org/10.1088/1361-6579/aabbc2}
\showDOI{\tempurl}


\bibitem[\protect\citeauthoryear{Fukui, Park, Yang, Rohrbach, Darrell, and
  Rohrbach}{Fukui et~al\mbox{.}}{2016}]%
        {Fukui2016MultimodalGrounding}
\bibfield{author}{\bibinfo{person}{Akira Fukui}, \bibinfo{person}{Dong~Huk
  Park}, \bibinfo{person}{Daylen Yang}, \bibinfo{person}{Anna Rohrbach},
  \bibinfo{person}{Trevor Darrell}, {and} \bibinfo{person}{Marcus Rohrbach}.}
  \bibinfo{year}{2016}\natexlab{}.
\newblock \showarticletitle{{Multimodal compact bilinear pooling for visual
  question answering and visual grounding}}. In \bibinfo{booktitle}{\emph{EMNLP
  2016 - Conference on Empirical Methods in Natural Language Processing,
  Proceedings}}. \bibinfo{publisher}{Association for Computational Linguistics
  (ACL)}, \bibinfo{pages}{457--468}.
\newblock
\showISBNx{9781945626258}
\urldef\tempurl%
\url{https://doi.org/10.18653/v1/d16-1044}
\showDOI{\tempurl}


\bibitem[\protect\citeauthoryear{Fultz, Bonmassar, Setsompop, Stickgold, Rosen,
  Polimeni, and Lewis}{Fultz et~al\mbox{.}}{2019}]%
        {Fultz2019CoupledSleep}
\bibfield{author}{\bibinfo{person}{Nina~E. Fultz}, \bibinfo{person}{Giorgio
  Bonmassar}, \bibinfo{person}{Kawin Setsompop}, \bibinfo{person}{Robert~A.
  Stickgold}, \bibinfo{person}{Bruce~R. Rosen}, \bibinfo{person}{Jonathan~R.
  Polimeni}, {and} \bibinfo{person}{Laura~D. Lewis}.}
  \bibinfo{year}{2019}\natexlab{}.
\newblock \showarticletitle{{Coupled electrophysiological, hemodynamic, and
  cerebrospinal fluid oscillations in human sleep}}.
\newblock \bibinfo{journal}{\emph{Science}} \bibinfo{volume}{366},
  \bibinfo{number}{6465} (\bibinfo{date}{11} \bibinfo{year}{2019}),
  \bibinfo{pages}{628--631}.
\newblock
\showISSN{10959203}
\urldef\tempurl%
\url{https://doi.org/10.1126/science.aax5440}
\showDOI{\tempurl}


\bibitem[\protect\citeauthoryear{Giovangrandi, Inan, Wiard, Etemadi, and
  Kovacs}{Giovangrandi et~al\mbox{.}}{2011}]%
        {Giovangrandi2011BallistocardiographyRevisiting}
\bibfield{author}{\bibinfo{person}{Laurent Giovangrandi},
  \bibinfo{person}{Omer~T. Inan}, \bibinfo{person}{Richard~M. Wiard},
  \bibinfo{person}{Mozziyar Etemadi}, {and} \bibinfo{person}{Gregory~T.A.
  Kovacs}.} \bibinfo{year}{2011}\natexlab{}.
\newblock \showarticletitle{{Ballistocardiography - A method worth
  revisiting}}. In \bibinfo{booktitle}{\emph{Proceedings of the Annual
  International Conference of the IEEE Engineering in Medicine and Biology
  Society, EMBS}}, Vol.~\bibinfo{volume}{2011}. \bibinfo{publisher}{NIH Public
  Access}, \bibinfo{pages}{4279--4282}.
\newblock
\showISBNx{9781424441211}
\showISSN{1557170X}
\urldef\tempurl%
\url{https://doi.org/10.1109/IEMBS.2011.6091062}
\showDOI{\tempurl}


\bibitem[\protect\citeauthoryear{Guan, Li, and Roli}{Guan
  et~al\mbox{.}}{2015}]%
        {Guan2015OnMethod}
\bibfield{author}{\bibinfo{person}{Yu Guan}, \bibinfo{person}{Chang~Tsun Li},
  {and} \bibinfo{person}{Fabio Roli}.} \bibinfo{year}{2015}\natexlab{}.
\newblock \showarticletitle{{On Reducing the Effect of Covariate Factors in
  Gait Recognition: A Classifier Ensemble Method}}.
\newblock \bibinfo{journal}{\emph{IEEE Transactions on Pattern Analysis and
  Machine Intelligence}} \bibinfo{volume}{37}, \bibinfo{number}{7}
  (\bibinfo{date}{7} \bibinfo{year}{2015}), \bibinfo{pages}{1521--1528}.
\newblock
\showISSN{01628828}
\urldef\tempurl%
\url{https://doi.org/10.1109/TPAMI.2014.2366766}
\showDOI{\tempurl}


\bibitem[\protect\citeauthoryear{Guan and Pl{\"{o}}tz}{Guan and
  Pl{\"{o}}tz}{2017}]%
        {Guan2017EnsemblesWearables}
\bibfield{author}{\bibinfo{person}{Yu Guan} {and} \bibinfo{person}{Thomas
  Pl{\"{o}}tz}.} \bibinfo{year}{2017}\natexlab{}.
\newblock \showarticletitle{{Ensembles of Deep LSTM Learners for Activity
  Recognition using Wearables}}.
\newblock \bibinfo{journal}{\emph{Proceedings of the ACM on Interactive,
  Mobile, Wearable and Ubiquitous Technologies}} \bibinfo{volume}{1},
  \bibinfo{number}{2} (\bibinfo{date}{3} \bibinfo{year}{2017}),
  \bibinfo{pages}{1--28}.
\newblock
\showISSN{2474-9567}
\urldef\tempurl%
\url{https://doi.org/10.1145/3090076}
\showDOI{\tempurl}


\bibitem[\protect\citeauthoryear{Guidotti, Monreale, Ruggieri, Turini,
  Giannotti, and Pedreschi}{Guidotti et~al\mbox{.}}{2018}]%
        {Guidotti2018AModels}
\bibfield{author}{\bibinfo{person}{Riccardo Guidotti}, \bibinfo{person}{Anna
  Monreale}, \bibinfo{person}{Salvatore Ruggieri}, \bibinfo{person}{Franco
  Turini}, \bibinfo{person}{Fosca Giannotti}, {and} \bibinfo{person}{Dino
  Pedreschi}.} \bibinfo{year}{2018}\natexlab{}.
\newblock \showarticletitle{{A survey of methods for explaining black box
  models}}.
\newblock \bibinfo{journal}{\emph{Comput. Surveys}} \bibinfo{volume}{51},
  \bibinfo{number}{5} (\bibinfo{date}{8} \bibinfo{year}{2018}).
\newblock
\showISSN{15577341}
\urldef\tempurl%
\url{https://doi.org/10.1145/3236009}
\showDOI{\tempurl}


\bibitem[\protect\citeauthoryear{Hayano, Yuda, and Yoshida}{Hayano
  et~al\mbox{.}}{2017}]%
        {Hayano2017SleepSignals}
\bibfield{author}{\bibinfo{person}{Junichiro Hayano}, \bibinfo{person}{Emi
  Yuda}, {and} \bibinfo{person}{Yutaka Yoshida}.}
  \bibinfo{year}{2017}\natexlab{}.
\newblock \showarticletitle{{Sleep stage classification by combination of
  actigraphic and heart rate signals}}.
\newblock \bibinfo{journal}{\emph{2017 IEEE International Conference on
  Consumer Electronics - Taiwan, ICCE-TW 2017}} (\bibinfo{year}{2017}),
  \bibinfo{pages}{387--388}.
\newblock
\showISBNx{9781509040179}
\urldef\tempurl%
\url{https://doi.org/10.1109/ICCE-China.2017.7991158}
\showDOI{\tempurl}


\bibitem[\protect\citeauthoryear{Hazirbas, Ma, Domokos, and Cremers}{Hazirbas
  et~al\mbox{.}}{2017}]%
        {Hazirbas2017FuseNet:Architecture}
\bibfield{author}{\bibinfo{person}{Caner Hazirbas}, \bibinfo{person}{Lingni
  Ma}, \bibinfo{person}{Csaba Domokos}, {and} \bibinfo{person}{Daniel
  Cremers}.} \bibinfo{year}{2017}\natexlab{}.
\newblock \showarticletitle{{FuseNet: Incorporating depth into semantic
  segmentation via fusion-based CNN architecture}}. In
  \bibinfo{booktitle}{\emph{Lecture Notes in Computer Science (including
  subseries Lecture Notes in Artificial Intelligence and Lecture Notes in
  Bioinformatics)}}, Vol.~\bibinfo{volume}{10111 LNCS}.
  \bibinfo{publisher}{Springer Verlag}, \bibinfo{pages}{213--228}.
\newblock
\showISBNx{9783319541808}
\showISSN{16113349}
\urldef\tempurl%
\url{https://doi.org/10.1007/978-3-319-54181-5{\_}14}
\showDOI{\tempurl}


\bibitem[\protect\citeauthoryear{He, Zhang, Ren, and Sun}{He
  et~al\mbox{.}}{2016}]%
        {He2016IdentityNetworks}
\bibfield{author}{\bibinfo{person}{Kaiming He}, \bibinfo{person}{Xiangyu
  Zhang}, \bibinfo{person}{Shaoqing Ren}, {and} \bibinfo{person}{Jian Sun}.}
  \bibinfo{year}{2016}\natexlab{}.
\newblock \showarticletitle{{Identity mappings in deep residual networks}}. In
  \bibinfo{booktitle}{\emph{Lecture Notes in Computer Science (including
  subseries Lecture Notes in Artificial Intelligence and Lecture Notes in
  Bioinformatics)}}, Vol.~\bibinfo{volume}{9908 LNCS}.
  \bibinfo{publisher}{Springer Verlag}, \bibinfo{pages}{630--645}.
\newblock
\showISBNx{9783319464923}
\showISSN{16113349}
\urldef\tempurl%
\url{https://doi.org/10.1007/978-3-319-46493-0{\_}38}
\showDOI{\tempurl}


\bibitem[\protect\citeauthoryear{Heaton}{Heaton}{2018}]%
        {Heaton2018IanLearning}
\bibfield{author}{\bibinfo{person}{Jeff Heaton}.}
  \bibinfo{year}{2018}\natexlab{}.
\newblock \showarticletitle{{Ian Goodfellow, Yoshua Bengio, and Aaron
  Courville: Deep learning}}.
\newblock \bibinfo{journal}{\emph{Genetic Programming and Evolvable Machines}}
  \bibinfo{volume}{19}, \bibinfo{number}{1-2} (\bibinfo{date}{6}
  \bibinfo{year}{2018}), \bibinfo{pages}{305--307}.
\newblock
\showISSN{1389-2576}
\urldef\tempurl%
\url{https://doi.org/10.1007/s10710-017-9314-z}
\showDOI{\tempurl}


\bibitem[\protect\citeauthoryear{Hsu, Ahuja, Yue, Hristov, Kabelac, and
  Katabi}{Hsu et~al\mbox{.}}{2017}]%
        {Hsu2017Zero-EffortSignals}
\bibfield{author}{\bibinfo{person}{Chen-Yu Hsu}, \bibinfo{person}{Aayush
  Ahuja}, \bibinfo{person}{Shichao Yue}, \bibinfo{person}{Rumen Hristov},
  \bibinfo{person}{Zachary Kabelac}, {and} \bibinfo{person}{Dina Katabi}.}
  \bibinfo{year}{2017}\natexlab{}.
\newblock \showarticletitle{{Zero-Effort In-Home Sleep and Insomnia Monitoring
  using Radio Signals}}.
\newblock \bibinfo{journal}{\emph{Proceedings of the ACM on Interactive,
  Mobile, Wearable and Ubiquitous Technologies}} \bibinfo{volume}{1},
  \bibinfo{number}{3} (\bibinfo{date}{9} \bibinfo{year}{2017}),
  \bibinfo{pages}{1--18}.
\newblock
\showISSN{2474-9567}
\urldef\tempurl%
\url{https://doi.org/10.1145/3130924}
\showDOI{\tempurl}


\bibitem[\protect\citeauthoryear{{IBER} and {C.}}{{IBER} and {C.}}{2007}]%
        {IBER2007TheRules}
\bibfield{author}{\bibinfo{person}{{IBER}} {and} \bibinfo{person}{{C.}}}
  \bibinfo{year}{2007}\natexlab{}.
\newblock \showarticletitle{{The AASM Manual for the Scoring of Sleep and
  Associated Events : Rules}}.
\newblock \bibinfo{journal}{\emph{Terminology and Technical Specification}}
  (\bibinfo{year}{2007}).
\newblock
\urldef\tempurl%
\url{https://ci.nii.ac.jp/naid/10024500923}
\showURL{%
\tempurl}


\bibitem[\protect\citeauthoryear{Kapoor and Picard}{Kapoor and Picard}{2005}]%
        {Kapoor2005MultimodalEnvironments}
\bibfield{author}{\bibinfo{person}{Ashish Kapoor} {and}
  \bibinfo{person}{Rosalind~W. Picard}.} \bibinfo{year}{2005}\natexlab{}.
\newblock \showarticletitle{{Multimodal affect recognition in learning
  environments}}. In \bibinfo{booktitle}{\emph{Proceedings of the 13th ACM
  International Conference on Multimedia, MM 2005}}. \bibinfo{publisher}{ACM
  Press}, \bibinfo{address}{New York, New York, USA},
  \bibinfo{pages}{677--682}.
\newblock
\showISBNx{1595930442}
\urldef\tempurl%
\url{https://doi.org/10.1145/1101149.1101300}
\showDOI{\tempurl}


\bibitem[\protect\citeauthoryear{Kapur, Auckley, Chowdhuri, Kuhlmann, Mehra,
  Ramar, and Harrod}{Kapur et~al\mbox{.}}{2017}]%
        {Kapur2017ClinicalGuideline}
\bibfield{author}{\bibinfo{person}{Vishesh~K. Kapur},
  \bibinfo{person}{Dennis~H. Auckley}, \bibinfo{person}{Susmita Chowdhuri},
  \bibinfo{person}{David~C. Kuhlmann}, \bibinfo{person}{Reena Mehra},
  \bibinfo{person}{Kannan Ramar}, {and} \bibinfo{person}{Christopher~G.
  Harrod}.} \bibinfo{year}{2017}\natexlab{}.
\newblock \showarticletitle{{Clinical Practice Guideline for Diagnostic Testing
  for Adult Obstructive Sleep Apnea: An American Academy of Sleep Medicine
  Clinical Practice Guideline}}.
\newblock \bibinfo{journal}{\emph{Journal of Clinical Sleep Medicine : JCSM :
  Official Publication of the American Academy of Sleep Medicine}}
  \bibinfo{volume}{13}, \bibinfo{number}{3} (\bibinfo{year}{2017}),
  \bibinfo{pages}{479}.
\newblock
\urldef\tempurl%
\url{https://doi.org/10.5664/JCSM.6506}
\showDOI{\tempurl}


\bibitem[\protect\citeauthoryear{Khosla, Deak, Gault, Goldstein, Hwang, Kwon,
  O'Hearn, Schutte-Rodin, Yurcheshen, Rosen, Kirsch, Chervin, Carden, Ramar,
  Nisha~Aurora, Kristo, Malhotra, Martin, Olson, Rosen, and Rowley}{Khosla
  et~al\mbox{.}}{2018}]%
        {Khosla2018ConsumerStatement}
\bibfield{author}{\bibinfo{person}{Seema Khosla}, \bibinfo{person}{Maryann~C.
  Deak}, \bibinfo{person}{Dominic Gault}, \bibinfo{person}{Cathy~A. Goldstein},
  \bibinfo{person}{Dennis Hwang}, \bibinfo{person}{Younghoon Kwon},
  \bibinfo{person}{Daniel O'Hearn}, \bibinfo{person}{Sharon Schutte-Rodin},
  \bibinfo{person}{Michael Yurcheshen}, \bibinfo{person}{Ilene~M. Rosen},
  \bibinfo{person}{Douglas~B. Kirsch}, \bibinfo{person}{Ronald~D. Chervin},
  \bibinfo{person}{Kelly~A. Carden}, \bibinfo{person}{Kannan Ramar},
  \bibinfo{person}{R. Nisha~Aurora}, \bibinfo{person}{David~A. Kristo},
  \bibinfo{person}{Raman~K. Malhotra}, \bibinfo{person}{Jennifer~L. Martin},
  \bibinfo{person}{Eric~J. Olson}, \bibinfo{person}{Carol~L. Rosen}, {and}
  \bibinfo{person}{James~A. Rowley}.} \bibinfo{year}{2018}\natexlab{}.
\newblock \showarticletitle{{Consumer sleep technology: An American academy of
  sleep medicine position statement}}.
\newblock \bibinfo{journal}{\emph{Journal of Clinical Sleep Medicine}}
  \bibinfo{volume}{14}, \bibinfo{number}{5} (\bibinfo{date}{5}
  \bibinfo{year}{2018}), \bibinfo{pages}{877--880}.
\newblock
\showISSN{15509397}
\urldef\tempurl%
\url{https://doi.org/10.5664/jcsm.7128}
\showDOI{\tempurl}


\bibitem[\protect\citeauthoryear{Kingma and Ba}{Kingma and Ba}{2015}]%
        {Kingma2015Adam:Optimization}
\bibfield{author}{\bibinfo{person}{Diederik~P. Kingma} {and}
  \bibinfo{person}{Jimmy~Lei Ba}.} \bibinfo{year}{2015}\natexlab{}.
\newblock \showarticletitle{{Adam: A method for stochastic optimization}}. In
  \bibinfo{booktitle}{\emph{3rd International Conference on Learning
  Representations, ICLR 2015 - Conference Track Proceedings}}.
  \bibinfo{publisher}{International Conference on Learning Representations,
  ICLR}.
\newblock
\urldef\tempurl%
\url{https://arxiv.org/abs/1412.6980v9}
\showURL{%
\tempurl}


\bibitem[\protect\citeauthoryear{Kohansieh and Makaryus}{Kohansieh and
  Makaryus}{2015}]%
        {Kohansieh2015SleepDisease}
\bibfield{author}{\bibinfo{person}{Michelle Kohansieh} {and}
  \bibinfo{person}{Amgad~N. Makaryus}.} \bibinfo{year}{2015}\natexlab{}.
\newblock \bibinfo{title}{{Sleep Deficiency and Deprivation Leading to
  Cardiovascular Disease}}.
\newblock
\newblock
\showISSN{20900392}
\urldef\tempurl%
\url{https://doi.org/10.1155/2015/615681}
\showDOI{\tempurl}


\bibitem[\protect\citeauthoryear{Kupfer and Foster}{Kupfer and Foster}{1972}]%
        {Kupfer1972INTERVALDEPRESSION}
\bibfield{author}{\bibinfo{person}{David~J. Kupfer} {and}
  \bibinfo{person}{F.~Gordon Foster}.} \bibinfo{year}{1972}\natexlab{}.
\newblock \showarticletitle{{INTERVAL BETWEEN ONSET OF SLEEP AND
  RAPID-EYE-MOVEMENT SLEEP AS AN INDICATOR OF DEPRESSION}}.
\newblock \bibinfo{journal}{\emph{The Lancet}} \bibinfo{volume}{300},
  \bibinfo{number}{7779} (\bibinfo{date}{9} \bibinfo{year}{1972}),
  \bibinfo{pages}{684--686}.
\newblock
\showISSN{01406736}
\urldef\tempurl%
\url{https://doi.org/10.1016/S0140-6736(72)92090-9}
\showDOI{\tempurl}


\bibitem[\protect\citeauthoryear{L, C, A, A, and D}{L et~al\mbox{.}}{2013}]%
        {L2013REMArt}
\bibfield{author}{\bibinfo{person}{Palagini L}, \bibinfo{person}{Baglioni C},
  \bibinfo{person}{Ciapparelli A}, \bibinfo{person}{Gemignani A}, {and}
  \bibinfo{person}{Riemann D}.} \bibinfo{year}{2013}\natexlab{}.
\newblock \showarticletitle{{REM sleep dysregulation in depression: state of
  the art}}.
\newblock \bibinfo{journal}{\emph{Sleep medicine reviews}}
  \bibinfo{volume}{17}, \bibinfo{number}{5} (\bibinfo{date}{10}
  \bibinfo{year}{2013}), \bibinfo{pages}{377--390}.
\newblock
\showISSN{1532-2955}
\urldef\tempurl%
\url{https://doi.org/10.1016/J.SMRV.2012.11.001}
\showDOI{\tempurl}


\bibitem[\protect\citeauthoryear{Lim, Dey, and Avrahami}{Lim
  et~al\mbox{.}}{2009}]%
        {Lim2009WhySystems}
\bibfield{author}{\bibinfo{person}{Brian~Y. Lim}, \bibinfo{person}{Anind~K.
  Dey}, {and} \bibinfo{person}{Daniel Avrahami}.}
  \bibinfo{year}{2009}\natexlab{}.
\newblock \showarticletitle{{Why and why not explanations improve the
  intelligibility of context-aware intelligent systems}}.
\newblock \bibinfo{journal}{\emph{Conference on Human Factors in Computing
  Systems - Proceedings}} (\bibinfo{year}{2009}), \bibinfo{pages}{2119--2128}.
\newblock
\urldef\tempurl%
\url{https://doi.org/10.1145/1518701.1519023}
\showDOI{\tempurl}


\bibitem[\protect\citeauthoryear{Lin, Roychowdhury, and Maji}{Lin
  et~al\mbox{.}}{2015}]%
        {Lin2015BilinearRecognition}
\bibfield{author}{\bibinfo{person}{Tsung~Yu Lin}, \bibinfo{person}{Aruni
  Roychowdhury}, {and} \bibinfo{person}{Subhransu Maji}.}
  \bibinfo{year}{2015}\natexlab{}.
\newblock \showarticletitle{{Bilinear CNN models for fine-grained visual
  recognition}}. In \bibinfo{booktitle}{\emph{Proceedings of the IEEE
  International Conference on Computer Vision}}, Vol.~\bibinfo{volume}{2015
  Inter}. \bibinfo{pages}{1449--1457}.
\newblock
\showISBNx{9781467383912}
\showISSN{15505499}
\urldef\tempurl%
\url{https://doi.org/10.1109/ICCV.2015.170}
\showDOI{\tempurl}


\bibitem[\protect\citeauthoryear{Liu, Yao, Li, Liu, Wang, Shao, and
  Abdelzaher}{Liu et~al\mbox{.}}{2020}]%
        {Liu2020GlobalFusion:Fusion}
\bibfield{author}{\bibinfo{person}{Shengzhong Liu}, \bibinfo{person}{Shuochao
  Yao}, \bibinfo{person}{Jinyang Li}, \bibinfo{person}{Dongxin Liu},
  \bibinfo{person}{Tianshi Wang}, \bibinfo{person}{Huajie Shao}, {and}
  \bibinfo{person}{Tarek Abdelzaher}.} \bibinfo{year}{2020}\natexlab{}.
\newblock \showarticletitle{{GlobalFusion: A global attentional deep learning
  framework for multisensor information fusion}}.
\newblock \bibinfo{journal}{\emph{Proceedings of the ACM on Interactive,
  Mobile, Wearable and Ubiquitous Technologies}} \bibinfo{volume}{4},
  \bibinfo{number}{1} (\bibinfo{date}{3} \bibinfo{year}{2020}),
  \bibinfo{pages}{1--27}.
\newblock
\showISSN{24749567}
\urldef\tempurl%
\url{https://doi.org/10.1145/3380999}
\showDOI{\tempurl}


\bibitem[\protect\citeauthoryear{Looney, Kidmose, Park, Ungstrup, Rank,
  Rosenkranz, and Mandic}{Looney et~al\mbox{.}}{2012}]%
        {Looney2012TheMonitoring}
\bibfield{author}{\bibinfo{person}{David Looney}, \bibinfo{person}{Preben
  Kidmose}, \bibinfo{person}{Cheolsoo Park}, \bibinfo{person}{Michael
  Ungstrup}, \bibinfo{person}{Mike Rank}, \bibinfo{person}{Karin Rosenkranz},
  {and} \bibinfo{person}{Danilo Mandic}.} \bibinfo{year}{2012}\natexlab{}.
\newblock \showarticletitle{{The in-the-ear recording concept: User-centered
  and wearable brain monitoring}}.
\newblock \bibinfo{journal}{\emph{IEEE Pulse}} \bibinfo{volume}{3},
  \bibinfo{number}{6} (\bibinfo{year}{2012}), \bibinfo{pages}{32--42}.
\newblock
\showISSN{21542287}
\urldef\tempurl%
\url{https://doi.org/10.1109/MPUL.2012.2216717}
\showDOI{\tempurl}


\bibitem[\protect\citeauthoryear{Lu, Yang, Batra, and Parikh}{Lu
  et~al\mbox{.}}{2016}]%
        {Lu2016HierarchicalAnswering}
\bibfield{author}{\bibinfo{person}{Jiasen Lu}, \bibinfo{person}{Jianwei Yang},
  \bibinfo{person}{Dhruv Batra}, {and} \bibinfo{person}{Devi Parikh}.}
  \bibinfo{year}{2016}\natexlab{}.
\newblock \showarticletitle{{Hierarchical question-image co-attention for
  visual question answering}}. In \bibinfo{booktitle}{\emph{Advances in Neural
  Information Processing Systems}}. \bibinfo{publisher}{Neural information
  processing systems foundation}, \bibinfo{pages}{289--297}.
\newblock
\showISSN{10495258}


\bibitem[\protect\citeauthoryear{Malliani, Pagani, Lombardi, and
  Cerutti}{Malliani et~al\mbox{.}}{1991}]%
        {Malliani1991CardiovascularDomain}
\bibfield{author}{\bibinfo{person}{A. Malliani}, \bibinfo{person}{M. Pagani},
  \bibinfo{person}{F. Lombardi}, {and} \bibinfo{person}{S. Cerutti}.}
  \bibinfo{year}{1991}\natexlab{}.
\newblock \showarticletitle{{Cardiovascular neural regulation explored in the
  frequency domain}}.
\newblock \bibinfo{journal}{\emph{Circulation}} \bibinfo{volume}{84},
  \bibinfo{number}{2} (\bibinfo{year}{1991}), \bibinfo{pages}{482--492}.
\newblock
\showISSN{00097322}
\urldef\tempurl%
\url{https://doi.org/10.1161/01.CIR.84.2.482}
\showDOI{\tempurl}


\bibitem[\protect\citeauthoryear{M{\'{e}}ndez, Blanchi, Villantieri, and
  Cerutti}{M{\'{e}}ndez et~al\mbox{.}}{2006}]%
        {Mendez2006Time-varyingStages}
\bibfield{author}{\bibinfo{person}{M. M{\'{e}}ndez}, \bibinfo{person}{A.~M.
  Blanchi}, \bibinfo{person}{O. Villantieri}, {and} \bibinfo{person}{S.
  Cerutti}.} \bibinfo{year}{2006}\natexlab{}.
\newblock \showarticletitle{{Time-varying analysis of the heart rate
  variability during REM and non REM sleep stages}}. In
  \bibinfo{booktitle}{\emph{Annual International Conference of the IEEE
  Engineering in Medicine and Biology - Proceedings}}.
  \bibinfo{pages}{3576--3579}.
\newblock
\showISBNx{1424400325}
\showISSN{05891019}
\urldef\tempurl%
\url{https://doi.org/10.1109/IEMBS.2006.260067}
\showDOI{\tempurl}


\bibitem[\protect\citeauthoryear{Mikkelsen, Tabar, Kappel, Christensen, Toft,
  Hemmsen, Rank, Otto, and Kidmose}{Mikkelsen et~al\mbox{.}}{2019}]%
        {Mikkelsen2019AccurateEar-EEG}
\bibfield{author}{\bibinfo{person}{Kaare~B. Mikkelsen},
  \bibinfo{person}{Yousef~R. Tabar}, \bibinfo{person}{Simon~L. Kappel},
  \bibinfo{person}{Christian~B. Christensen}, \bibinfo{person}{Hans~O. Toft},
  \bibinfo{person}{Martin~C. Hemmsen}, \bibinfo{person}{Mike~L. Rank},
  \bibinfo{person}{Marit Otto}, {and} \bibinfo{person}{Preben Kidmose}.}
  \bibinfo{year}{2019}\natexlab{}.
\newblock \showarticletitle{{Accurate whole-night sleep monitoring with
  dry-contact ear-EEG}}.
\newblock \bibinfo{journal}{\emph{Scientific Reports}} \bibinfo{volume}{9},
  \bibinfo{number}{1} (\bibinfo{date}{12} \bibinfo{year}{2019}),
  \bibinfo{pages}{1--12}.
\newblock
\showISSN{20452322}
\urldef\tempurl%
\url{https://doi.org/10.1038/s41598-019-53115-3}
\showDOI{\tempurl}


\bibitem[\protect\citeauthoryear{Mikkelsen, Villadsen, Otto, and
  Kidmose}{Mikkelsen et~al\mbox{.}}{2017}]%
        {Mikkelsen2017AutomaticEar-EEG}
\bibfield{author}{\bibinfo{person}{Kaare~B. Mikkelsen},
  \bibinfo{person}{David~Bové Villadsen}, \bibinfo{person}{Marit Otto}, {and}
  \bibinfo{person}{Preben Kidmose}.} \bibinfo{year}{2017}\natexlab{}.
\newblock \showarticletitle{{Automatic sleep staging using ear-EEG}}.
\newblock \bibinfo{journal}{\emph{BioMedical Engineering Online}}
  \bibinfo{volume}{16}, \bibinfo{number}{1} (\bibinfo{date}{9}
  \bibinfo{year}{2017}), \bibinfo{pages}{111}.
\newblock
\showISSN{1475925X}
\urldef\tempurl%
\url{https://doi.org/10.1186/s12938-017-0400-5}
\showDOI{\tempurl}


\bibitem[\protect\citeauthoryear{Montano, Porta, Cogliati, Costantino,
  Tobaldini, Casali, and Iellamo}{Montano et~al\mbox{.}}{2009}]%
        {Montano2009HeartBehavior}
\bibfield{author}{\bibinfo{person}{Nicola Montano}, \bibinfo{person}{Alberto
  Porta}, \bibinfo{person}{Chiara Cogliati}, \bibinfo{person}{Giorgio
  Costantino}, \bibinfo{person}{Eleonora Tobaldini},
  \bibinfo{person}{Karina~Rabello Casali}, {and} \bibinfo{person}{Ferdinando
  Iellamo}.} \bibinfo{year}{2009}\natexlab{}.
\newblock \bibinfo{title}{{Heart rate variability explored in the frequency
  domain: A tool to investigate the link between heart and behavior}}.
\newblock , \bibinfo{numpages}{71--80}~pages.
\newblock
\showISSN{01497634}
\urldef\tempurl%
\url{https://doi.org/10.1016/j.neubiorev.2008.07.006}
\showDOI{\tempurl}


\bibitem[\protect\citeauthoryear{Monti, Medigue, Nedelcoux, and
  Escourrou}{Monti et~al\mbox{.}}{2002}]%
        {Monti2002AutonomicSubjects}
\bibfield{author}{\bibinfo{person}{A Monti}, \bibinfo{person}{C. Medigue},
  \bibinfo{person}{H. Nedelcoux}, {and} \bibinfo{person}{P. Escourrou}.}
  \bibinfo{year}{2002}\natexlab{}.
\newblock \showarticletitle{{Autonomic control of the cardiovascular system
  during sleep in normal subjects}}.
\newblock \bibinfo{journal}{\emph{European Journal of Applied Physiology}}
  \bibinfo{volume}{87}, \bibinfo{number}{2} (\bibinfo{date}{6}
  \bibinfo{year}{2002}), \bibinfo{pages}{174--181}.
\newblock
\showISSN{14396319}
\urldef\tempurl%
\url{https://doi.org/10.1007/s00421-002-0597-1}
\showDOI{\tempurl}


\bibitem[\protect\citeauthoryear{Neverova, Wolf, Taylor, and Nebout}{Neverova
  et~al\mbox{.}}{2016}]%
        {Neverova2016ModDrop:Recognition}
\bibfield{author}{\bibinfo{person}{Natalia Neverova},
  \bibinfo{person}{Christian Wolf}, \bibinfo{person}{Graham Taylor}, {and}
  \bibinfo{person}{Florian Nebout}.} \bibinfo{year}{2016}\natexlab{}.
\newblock \showarticletitle{{ModDrop: Adaptive multi-modal gesture
  recognition}}.
\newblock \bibinfo{journal}{\emph{IEEE Transactions on Pattern Analysis and
  Machine Intelligence}} \bibinfo{volume}{38}, \bibinfo{number}{8}
  (\bibinfo{date}{8} \bibinfo{year}{2016}), \bibinfo{pages}{1692--1706}.
\newblock
\showISSN{01628828}
\urldef\tempurl%
\url{https://doi.org/10.1109/TPAMI.2015.2461544}
\showDOI{\tempurl}


\bibitem[\protect\citeauthoryear{Orhan and Pitkow}{Orhan and Pitkow}{2017}]%
        {Orhan2017SkipSingularities}
\bibfield{author}{\bibinfo{person}{A.~Emin Orhan} {and} \bibinfo{person}{Xaq
  Pitkow}.} \bibinfo{year}{2017}\natexlab{}.
\newblock \showarticletitle{{Skip Connections Eliminate Singularities}}.
\newblock \bibinfo{journal}{\emph{arXiv}} (\bibinfo{date}{1}
  \bibinfo{year}{2017}).
\newblock
\urldef\tempurl%
\url{http://arxiv.org/abs/1701.09175}
\showURL{%
\tempurl}


\bibitem[\protect\citeauthoryear{Palotti, Mall, Aupetit, Rueschman, Singh,
  Sathyanarayana, Taheri, and Fernandez-Luque}{Palotti et~al\mbox{.}}{2019}]%
        {Palotti2019BenchmarkTechniques}
\bibfield{author}{\bibinfo{person}{Joao Palotti}, \bibinfo{person}{Raghvendra
  Mall}, \bibinfo{person}{Michael Aupetit}, \bibinfo{person}{Michael
  Rueschman}, \bibinfo{person}{Meghna Singh}, \bibinfo{person}{Aarti
  Sathyanarayana}, \bibinfo{person}{Shahrad Taheri}, {and}
  \bibinfo{person}{Luis Fernandez-Luque}.} \bibinfo{year}{2019}\natexlab{}.
\newblock \showarticletitle{{Benchmark on a large cohort for sleep-wake
  classification with machine learning techniques}}.
\newblock \bibinfo{journal}{\emph{npj Digital Medicine}} \bibinfo{volume}{2},
  \bibinfo{number}{1} (\bibinfo{date}{12} \bibinfo{year}{2019}),
  \bibinfo{pages}{1--9}.
\newblock
\showISSN{2398-6352}
\urldef\tempurl%
\url{https://doi.org/10.1038/s41746-019-0126-9}
\showDOI{\tempurl}


\bibitem[\protect\citeauthoryear{Park, Hwang, Jung, Yoon, and Lee}{Park
  et~al\mbox{.}}{2014}]%
        {Park2014BallistocardiographyEstimation}
\bibfield{author}{\bibinfo{person}{Kwang~Suk Park}, \bibinfo{person}{Su~Hwan
  Hwang}, \bibinfo{person}{Da~Woon Jung}, \bibinfo{person}{Hee~Nam Yoon}, {and}
  \bibinfo{person}{Won~Kyu Lee}.} \bibinfo{year}{2014}\natexlab{}.
\newblock \showarticletitle{{Ballistocardiography for nonintrusive sleep
  structure estimation}}. In \bibinfo{booktitle}{\emph{2014 36th Annual
  International Conference of the IEEE Engineering in Medicine and Biology
  Society, EMBC 2014}}. \bibinfo{publisher}{Institute of Electrical and
  Electronics Engineers Inc.}, \bibinfo{pages}{5184--5187}.
\newblock
\showISBNx{9781424479290}
\urldef\tempurl%
\url{https://doi.org/10.1109/EMBC.2014.6944793}
\showDOI{\tempurl}


\bibitem[\protect\citeauthoryear{Perez, Mahaffey, Hedlin, Rumsfeld, Garcia,
  Ferris, Balasubramanian, Russo, Rajmane, Cheung, Hung, Lee, Kowey, Talati,
  Nag, Gummidipundi, Beatty, Hills, Desai, Granger, Desai, and Turakhia}{Perez
  et~al\mbox{.}}{2019}]%
        {Perez2019Large-ScaleFibrillation}
\bibfield{author}{\bibinfo{person}{Marco~V. Perez}, \bibinfo{person}{Kenneth~W.
  Mahaffey}, \bibinfo{person}{Haley Hedlin}, \bibinfo{person}{John~S.
  Rumsfeld}, \bibinfo{person}{Ariadna Garcia}, \bibinfo{person}{Todd Ferris},
  \bibinfo{person}{Vidhya Balasubramanian}, \bibinfo{person}{Andrea~M. Russo},
  \bibinfo{person}{Amol Rajmane}, \bibinfo{person}{Lauren Cheung},
  \bibinfo{person}{Grace Hung}, \bibinfo{person}{Justin Lee},
  \bibinfo{person}{Peter Kowey}, \bibinfo{person}{Nisha Talati},
  \bibinfo{person}{Divya Nag}, \bibinfo{person}{Santosh~E. Gummidipundi},
  \bibinfo{person}{Alexis Beatty}, \bibinfo{person}{Mellanie~True Hills},
  \bibinfo{person}{Sumbul Desai}, \bibinfo{person}{Christopher~B. Granger},
  \bibinfo{person}{Manisha Desai}, {and} \bibinfo{person}{Mintu~P. Turakhia}.}
  \bibinfo{year}{2019}\natexlab{}.
\newblock \showarticletitle{{Large-Scale Assessment of a Smartwatch to Identify
  Atrial Fibrillation}}.
\newblock \bibinfo{journal}{\emph{New England Journal of Medicine}}
  \bibinfo{volume}{381}, \bibinfo{number}{20} (\bibinfo{date}{11}
  \bibinfo{year}{2019}), \bibinfo{pages}{1909--1917}.
\newblock
\showISSN{0028-4793}
\urldef\tempurl%
\url{https://doi.org/10.1056/NEJMoa1901183}
\showDOI{\tempurl}


\bibitem[\protect\citeauthoryear{Perez-Pozuelo, Zhai, Palotti, Mall, Aupetit,
  Garcia-Gomez, Taheri, Guan, and Fernandez-Luque}{Perez-Pozuelo
  et~al\mbox{.}}{2020}]%
        {Perez-Pozuelo2020TheMedicine}
\bibfield{author}{\bibinfo{person}{Ignacio Perez-Pozuelo},
  \bibinfo{person}{Bing Zhai}, \bibinfo{person}{Joao Palotti},
  \bibinfo{person}{Raghvendra Mall}, \bibinfo{person}{Michaël Aupetit},
  \bibinfo{person}{Juan~M. Garcia-Gomez}, \bibinfo{person}{Shahrad Taheri},
  \bibinfo{person}{Yu Guan}, {and} \bibinfo{person}{Luis Fernandez-Luque}.}
  \bibinfo{year}{2020}\natexlab{}.
\newblock \bibinfo{title}{{The future of sleep health: a data-driven revolution
  in sleep science and medicine}}.
\newblock , \bibinfo{numpages}{15}~pages.
\newblock
\showISSN{23986352}
\urldef\tempurl%
\url{https://doi.org/10.1038/s41746-020-0244-4}
\showDOI{\tempurl}


\bibitem[\protect\citeauthoryear{Peters}{Peters}{2021}]%
        {Peters2021OvernightResults}
\bibfield{author}{\bibinfo{person}{Brandon Peters}.}
  \bibinfo{year}{2021}\natexlab{}.
\newblock \bibinfo{title}{{Overnight Sleep Study: Uses, Procedure, Results}}.
\newblock
\newblock
\urldef\tempurl%
\url{https://www.verywellhealth.com/what-to-expect-in-a-sleep-study-3015121}
\showURL{%
\tempurl}


\bibitem[\protect\citeauthoryear{Phan, Andreotti, Cooray, Chen, and
  De~Vos}{Phan et~al\mbox{.}}{2019}]%
        {Phan2019SeqSleepNet:Staging}
\bibfield{author}{\bibinfo{person}{Huy Phan}, \bibinfo{person}{Fernando
  Andreotti}, \bibinfo{person}{Navin Cooray}, \bibinfo{person}{Oliver~Y. Chen},
  {and} \bibinfo{person}{Maarten De~Vos}.} \bibinfo{year}{2019}\natexlab{}.
\newblock \showarticletitle{{SeqSleepNet: End-to-End Hierarchical Recurrent
  Neural Network for Sequence-to-Sequence Automatic Sleep Staging}}.
\newblock \bibinfo{journal}{\emph{IEEE Transactions on Neural Systems and
  Rehabilitation Engineering}} \bibinfo{volume}{27}, \bibinfo{number}{3}
  (\bibinfo{year}{2019}), \bibinfo{pages}{400--410}.
\newblock
\showISSN{15344320}
\urldef\tempurl%
\url{https://doi.org/10.1109/TNSRE.2019.2896659}
\showDOI{\tempurl}


\bibitem[\protect\citeauthoryear{Phan, Chen, Tran, Koch, Mertins, and
  De~Vos}{Phan et~al\mbox{.}}{2021}]%
        {Phan2021XSleepNet:Staging}
\bibfield{author}{\bibinfo{person}{Huy Phan}, \bibinfo{person}{Oliver~Y. Chen},
  \bibinfo{person}{Minh~C. Tran}, \bibinfo{person}{Philipp Koch},
  \bibinfo{person}{Alfred Mertins}, {and} \bibinfo{person}{Maarten De~Vos}.}
  \bibinfo{year}{2021}\natexlab{}.
\newblock \showarticletitle{{XSleepNet: Multi-View Sequential Model for
  Automatic Sleep Staging}}.
\newblock \bibinfo{journal}{\emph{IEEE Transactions on Pattern Analysis and
  Machine Intelligence}} (\bibinfo{year}{2021}).
\newblock
\showISSN{19393539}
\urldef\tempurl%
\url{https://doi.org/10.1109/TPAMI.2021.3070057}
\showDOI{\tempurl}


\bibitem[\protect\citeauthoryear{Radu, Tong, Bhattacharya, Lane, Mascolo,
  Marina, and Kawsar}{Radu et~al\mbox{.}}{2018}]%
        {radu2018multimodal}
\bibfield{author}{\bibinfo{person}{Valentin Radu}, \bibinfo{person}{Catherine
  Tong}, \bibinfo{person}{Sourav Bhattacharya}, \bibinfo{person}{Nicholas~D
  Lane}, \bibinfo{person}{Cecilia Mascolo}, \bibinfo{person}{Mahesh~K. Marina},
  {and} \bibinfo{person}{Fahim Kawsar}.} \bibinfo{year}{2018}\natexlab{}.
\newblock \showarticletitle{{Multimodal Deep Learning for Activity and Context
  Recognition}}.
\newblock \bibinfo{journal}{\emph{Proceedings of the ACM on Interactive,
  Mobile, Wearable and Ubiquitous Technologies}} \bibinfo{volume}{1},
  \bibinfo{number}{4} (\bibinfo{year}{2018}), \bibinfo{pages}{1--27}.
\newblock
\showISSN{2474-9567}
\urldef\tempurl%
\url{https://doi.org/10.1145/3161174}
\showDOI{\tempurl}


\bibitem[\protect\citeauthoryear{Rattani, Kisku, Bicego, and
  Tistarelli}{Rattani et~al\mbox{.}}{2007}]%
        {Rattani2007FeatureBiometrics}
\bibfield{author}{\bibinfo{person}{A. Rattani}, \bibinfo{person}{D.~R. Kisku},
  \bibinfo{person}{M. Bicego}, {and} \bibinfo{person}{M. Tistarelli}.}
  \bibinfo{year}{2007}\natexlab{}.
\newblock \showarticletitle{{Feature level fusion of face and fingerprint
  biometrics}}. In \bibinfo{booktitle}{\emph{IEEE Conference on Biometrics:
  Theory, Applications and Systems, BTAS'07}}.
\newblock
\showISBNx{9781424415977}
\urldef\tempurl%
\url{https://doi.org/10.1109/BTAS.2007.4401919}
\showDOI{\tempurl}


\bibitem[\protect\citeauthoryear{Ravichandran, Sien, Patel, Kientz, and
  Pina}{Ravichandran et~al\mbox{.}}{2017}]%
        {Ravichandran2017MakingHealth}
\bibfield{author}{\bibinfo{person}{Ruth Ravichandran},
  \bibinfo{person}{Sang~Wha Sien}, \bibinfo{person}{Shwetak~N. Patel},
  \bibinfo{person}{Julie~A. Kientz}, {and} \bibinfo{person}{Laura~R. Pina}.}
  \bibinfo{year}{2017}\natexlab{}.
\newblock \showarticletitle{{Making sense of sleep sensors: How sleep sensing
  technologies support and undermine sleep health}}.
\newblock \bibinfo{journal}{\emph{Conference on Human Factors in Computing
  Systems - Proceedings}}  \bibinfo{volume}{2017-May} (\bibinfo{date}{5}
  \bibinfo{year}{2017}), \bibinfo{pages}{6864--6875}.
\newblock
\urldef\tempurl%
\url{https://doi.org/10.1145/3025453.3025557}
\showDOI{\tempurl}


\bibitem[\protect\citeauthoryear{S and E}{S and E}{2018}]%
        {S2018SleepDiabetes}
\bibfield{author}{\bibinfo{person}{Reutrakul S} {and}
  \bibinfo{person}{Van~Cauter E}.} \bibinfo{year}{2018}\natexlab{}.
\newblock \showarticletitle{{Sleep influences on obesity, insulin resistance,
  and risk of type 2 diabetes}}.
\newblock \bibinfo{journal}{\emph{Metabolism: clinical and experimental}}
  \bibinfo{volume}{84} (\bibinfo{date}{7} \bibinfo{year}{2018}),
  \bibinfo{pages}{56--66}.
\newblock
\showISSN{1532-8600}
\urldef\tempurl%
\url{https://doi.org/10.1016/J.METABOL.2018.02.010}
\showDOI{\tempurl}


\bibitem[\protect\citeauthoryear{Schwartz and Roth}{Schwartz and Roth}{2009}]%
        {Schwartz2009NeurophysiologyImplications}
\bibfield{author}{\bibinfo{person}{Jonathan Schwartz} {and}
  \bibinfo{person}{Thomas Roth}.} \bibinfo{year}{2009}\natexlab{}.
\newblock \showarticletitle{{Neurophysiology of Sleep and Wakefulness: Basic
  Science and Clinical Implications}}.
\newblock \bibinfo{journal}{\emph{Current Neuropharmacology}}
  \bibinfo{volume}{6}, \bibinfo{number}{4} (\bibinfo{date}{2}
  \bibinfo{year}{2009}), \bibinfo{pages}{367--378}.
\newblock
\showISSN{1570159X}
\urldef\tempurl%
\url{https://doi.org/10.2174/157015908787386050}
\showDOI{\tempurl}


\bibitem[\protect\citeauthoryear{Selvaraju, Cogswell, Das, Vedantam, Parikh,
  and Batra}{Selvaraju et~al\mbox{.}}{2020}]%
        {Selvaraju2020Grad-CAM:Localization}
\bibfield{author}{\bibinfo{person}{Ramprasaath~R Selvaraju},
  \bibinfo{person}{Michael Cogswell}, \bibinfo{person}{Abhishek Das},
  \bibinfo{person}{Ramakrishna Vedantam}, \bibinfo{person}{Devi Parikh}, {and}
  \bibinfo{person}{Dhruv Batra}.} \bibinfo{year}{2020}\natexlab{}.
\newblock \showarticletitle{{Grad-CAM: Visual Explanations from Deep Networks
  via Gradient-Based Localization}}.
\newblock \bibinfo{journal}{\emph{International Journal of Computer Vision}}
  \bibinfo{volume}{128}, \bibinfo{number}{2} (\bibinfo{year}{2020}),
  \bibinfo{pages}{336--359}.
\newblock
\showISSN{15731405}
\urldef\tempurl%
\url{https://doi.org/10.1007/s11263-019-01228-7}
\showDOI{\tempurl}


\bibitem[\protect\citeauthoryear{Simonyan and Zisserman}{Simonyan and
  Zisserman}{2015}]%
        {Simonyan2015VeryRecognition}
\bibfield{author}{\bibinfo{person}{Karen Simonyan} {and}
  \bibinfo{person}{Andrew Zisserman}.} \bibinfo{year}{2015}\natexlab{}.
\newblock \showarticletitle{{Very deep convolutional networks for large-scale
  image recognition}}. In \bibinfo{booktitle}{\emph{3rd International
  Conference on Learning Representations, ICLR 2015 - Conference Track
  Proceedings}}. \bibinfo{publisher}{International Conference on Learning
  Representations, ICLR}.
\newblock
\urldef\tempurl%
\url{http://www.robots.ox.ac.uk/}
\showURL{%
\tempurl}


\bibitem[\protect\citeauthoryear{Stefani and H{\"{o}}gl}{Stefani and
  H{\"{o}}gl}{2019}]%
        {Stefani2019SleepDisease}
\bibfield{author}{\bibinfo{person}{Ambra Stefani} {and} \bibinfo{person}{Birgit
  H{\"{o}}gl}.} \bibinfo{year}{2019}\natexlab{}.
\newblock \showarticletitle{{Sleep in Parkinson’s disease}}.
\newblock \bibinfo{journal}{\emph{Neuropsychopharmacology 2019 45:1}}
  \bibinfo{volume}{45}, \bibinfo{number}{1} (\bibinfo{date}{6}
  \bibinfo{year}{2019}), \bibinfo{pages}{121--128}.
\newblock
\showISSN{1740-634X}
\urldef\tempurl%
\url{https://doi.org/10.1038/s41386-019-0448-y}
\showDOI{\tempurl}


\bibitem[\protect\citeauthoryear{Te~Lindert and Van~Someren}{Te~Lindert and
  Van~Someren}{2013}]%
        {TeLindert2013SleepMEMS}
\bibfield{author}{\bibinfo{person}{Bart~H.W. Te~Lindert} {and}
  \bibinfo{person}{Eus~J.W. Van~Someren}.} \bibinfo{year}{2013}\natexlab{}.
\newblock \showarticletitle{{Sleep estimates using microelectromechanical
  systems (MEMS)}}.
\newblock \bibinfo{journal}{\emph{Sleep}} \bibinfo{volume}{36},
  \bibinfo{number}{5} (\bibinfo{year}{2013}), \bibinfo{pages}{781--789}.
\newblock
\showISSN{01618105}
\urldef\tempurl%
\url{https://doi.org/10.5665/sleep.2648}
\showDOI{\tempurl}


\bibitem[\protect\citeauthoryear{Teo, Davila, Yang, Hii, Pua, Yap, Tan,
  Sahl{\'{e}}n, Chin, Teh, Rozen, Cook, Yeo, Tan, and Lim}{Teo
  et~al\mbox{.}}{2019}]%
        {Teo2019DigitalAging}
\bibfield{author}{\bibinfo{person}{Jing~Xian Teo}, \bibinfo{person}{Sonia
  Davila}, \bibinfo{person}{Chengxi Yang}, \bibinfo{person}{An~An Hii},
  \bibinfo{person}{Chee~Jian Pua}, \bibinfo{person}{Jonathan Yap},
  \bibinfo{person}{Swee~Yaw Tan}, \bibinfo{person}{Anders Sahl{\'{e}}n},
  \bibinfo{person}{Calvin Woon-Loong Chin}, \bibinfo{person}{Bin~Tean Teh},
  \bibinfo{person}{Steven~G. Rozen}, \bibinfo{person}{Stuart~Alexander Cook},
  \bibinfo{person}{Khung~Keong Yeo}, \bibinfo{person}{Patrick Tan}, {and}
  \bibinfo{person}{Weng~Khong Lim}.} \bibinfo{year}{2019}\natexlab{}.
\newblock \showarticletitle{{Digital phenotyping by consumer wearables
  identifies sleep-associated markers of cardiovascular disease risk and
  biological aging}}.
\newblock \bibinfo{journal}{\emph{Communications Biology 2019 2:1}}
  \bibinfo{volume}{2}, \bibinfo{number}{1} (\bibinfo{date}{10}
  \bibinfo{year}{2019}), \bibinfo{pages}{1--10}.
\newblock
\showISSN{2399-3642}
\urldef\tempurl%
\url{https://doi.org/10.1038/s42003-019-0605-1}
\showDOI{\tempurl}


\bibitem[\protect\citeauthoryear{Tobaldini, Nobili, Strada, Casali, Braghiroli,
  and Montano}{Tobaldini et~al\mbox{.}}{2013}]%
        {Tobaldini2013HeartSleep}
\bibfield{author}{\bibinfo{person}{Eleonora Tobaldini}, \bibinfo{person}{Lino
  Nobili}, \bibinfo{person}{Silvia Strada}, \bibinfo{person}{Karina~R. Casali},
  \bibinfo{person}{Alberto Braghiroli}, {and} \bibinfo{person}{Nicola
  Montano}.} \bibinfo{year}{2013}\natexlab{}.
\newblock \showarticletitle{{Heart rate variability in normal and pathological
  sleep}}.
\newblock \bibinfo{journal}{\emph{Frontiers in Physiology}}
  \bibinfo{volume}{4} (\bibinfo{date}{10} \bibinfo{year}{2013}),
  \bibinfo{pages}{1--11}.
\newblock
\showISSN{1664042X}
\urldef\tempurl%
\url{https://doi.org/10.3389/fphys.2013.00294}
\showDOI{\tempurl}


\bibitem[\protect\citeauthoryear{Vanoli, Adamson, {Ba-Lin}, Pinna, Lazzara, and
  Orr}{Vanoli et~al\mbox{.}}{1995}]%
        {Vanoli1995a}
\bibfield{author}{\bibinfo{person}{Emilio Vanoli}, \bibinfo{person}{Philip~B.
  Adamson}, \bibinfo{person}{{Ba-Lin}}, \bibinfo{person}{Gian~D. Pinna},
  \bibinfo{person}{Ralph Lazzara}, {and} \bibinfo{person}{William~C. Orr}.}
  \bibinfo{year}{1995}\natexlab{}.
\newblock \showarticletitle{{Heart Rate Variability During Specific Sleep
  Stages}}.
\newblock \bibinfo{journal}{\emph{Circulation}} \bibinfo{volume}{91},
  \bibinfo{number}{7} (\bibinfo{date}{4} \bibinfo{year}{1995}),
  \bibinfo{pages}{1918--1922}.
\newblock
\urldef\tempurl%
\url{https://doi.org/10.1161/01.CIR.91.7.1918}
\showDOI{\tempurl}


\bibitem[\protect\citeauthoryear{Walch}{Walch}{2019}]%
        {Walch2019MotionV1.0.0}
\bibfield{author}{\bibinfo{person}{Olivia Walch}.}
  \bibinfo{year}{2019}\natexlab{}.
\newblock \bibinfo{title}{{Motion and heart rate from a wrist-worn wearable and
  labeled sleep from polysomnography v1.0.0}}.
\newblock
\newblock
\urldef\tempurl%
\url{https://physionet.org/content/sleep-accel/1.0.0/}
\showURL{%
\tempurl}


\bibitem[\protect\citeauthoryear{Walch, Huang, Forger, and Goldstein}{Walch
  et~al\mbox{.}}{2019}]%
        {Walch2019SleepDevice}
\bibfield{author}{\bibinfo{person}{Olivia Walch}, \bibinfo{person}{Yitong
  Huang}, \bibinfo{person}{Daniel Forger}, {and} \bibinfo{person}{Cathy
  Goldstein}.} \bibinfo{year}{2019}\natexlab{}.
\newblock \showarticletitle{{Sleep stage prediction with raw acceleration and
  photoplethysmography heart rate data derived from a consumer wearable
  device}}.
\newblock \bibinfo{journal}{\emph{Sleep}} \bibinfo{volume}{42},
  \bibinfo{number}{12} (\bibinfo{date}{12} \bibinfo{year}{2019}).
\newblock
\showISSN{15509109}
\urldef\tempurl%
\url{https://doi.org/10.1093/sleep/zsz180}
\showDOI{\tempurl}


\bibitem[\protect\citeauthoryear{Wang, Li, Zhang, Zhang, Qu, and Huang}{Wang
  et~al\mbox{.}}{2015}]%
        {Wang2015TheDepression}
\bibfield{author}{\bibinfo{person}{Yi-Qun Wang}, \bibinfo{person}{Rui Li},
  \bibinfo{person}{Meng-Qi Zhang}, \bibinfo{person}{Ze Zhang},
  \bibinfo{person}{Wei-Min Qu}, {and} \bibinfo{person}{Zhi-Li Huang}.}
  \bibinfo{year}{2015}\natexlab{}.
\newblock \showarticletitle{{The Neurobiological Mechanisms and Treatments of
  REM Sleep Disturbances in Depression}}.
\newblock \bibinfo{journal}{\emph{Current Neuropharmacology}}
  \bibinfo{volume}{13}, \bibinfo{number}{4} (\bibinfo{date}{9}
  \bibinfo{year}{2015}), \bibinfo{pages}{543}.
\newblock
\urldef\tempurl%
\url{https://doi.org/10.2174/1570159X13666150310002540}
\showDOI{\tempurl}


\bibitem[\protect\citeauthoryear{Wang, Li, Duan, Su, Zhang, and Guan}{Wang
  et~al\mbox{.}}{2021}]%
        {Wang2021MedicalTransform}
\bibfield{author}{\bibinfo{person}{Zeyu Wang}, \bibinfo{person}{Xiongfei Li},
  \bibinfo{person}{Haoran Duan}, \bibinfo{person}{Yanchi Su},
  \bibinfo{person}{Xiaoli Zhang}, {and} \bibinfo{person}{Xinjiang Guan}.}
  \bibinfo{year}{2021}\natexlab{}.
\newblock \showarticletitle{{Medical image fusion based on convolutional neural
  networks and non-subsampled contourlet transform}}.
\newblock \bibinfo{journal}{\emph{Expert Systems with Applications}}
  \bibinfo{volume}{171} (\bibinfo{date}{6} \bibinfo{year}{2021}),
  \bibinfo{pages}{114574}.
\newblock
\showISSN{0957-4174}
\urldef\tempurl%
\url{https://doi.org/10.1016/J.ESWA.2021.114574}
\showDOI{\tempurl}


\bibitem[\protect\citeauthoryear{Wang, Li, Duan, Zhang, and Wang}{Wang
  et~al\mbox{.}}{2019}]%
        {Wang2019MultifocusDomain}
\bibfield{author}{\bibinfo{person}{Zeyu Wang}, \bibinfo{person}{Xiongfei Li},
  \bibinfo{person}{Haoran Duan}, \bibinfo{person}{Xiaoli Zhang}, {and}
  \bibinfo{person}{Hancheng Wang}.} \bibinfo{year}{2019}\natexlab{}.
\newblock \showarticletitle{{Multifocus image fusion using convolutional neural
  networks in the discrete wavelet transform domain}}.
\newblock \bibinfo{journal}{\emph{Multimedia Tools and Applications 2019
  78:24}} \bibinfo{volume}{78}, \bibinfo{number}{24} (\bibinfo{date}{8}
  \bibinfo{year}{2019}), \bibinfo{pages}{34483--34512}.
\newblock
\showISSN{1573-7721}
\urldef\tempurl%
\url{https://doi.org/10.1007/S11042-019-08070-6}
\showDOI{\tempurl}


\bibitem[\protect\citeauthoryear{Yang, Nguyen, San, Li, and Krishnaswamy}{Yang
  et~al\mbox{.}}{2015}]%
        {Yang2015DeepRecognition}
\bibfield{author}{\bibinfo{person}{Jian~Bo Yang}, \bibinfo{person}{Minh~Nhut
  Nguyen}, \bibinfo{person}{Phyo~Phyo San}, \bibinfo{person}{Xiao~Li Li}, {and}
  \bibinfo{person}{Shonali Krishnaswamy}.} \bibinfo{year}{2015}\natexlab{}.
\newblock \showarticletitle{{Deep convolutional neural networks on multichannel
  time series for human activity recognition}}. In
  \bibinfo{booktitle}{\emph{IJCAI International Joint Conference on Artificial
  Intelligence}}, Vol.~\bibinfo{volume}{2015-Janua}.
  \bibinfo{pages}{3995--4001}.
\newblock
\showISBNx{9781577357384}
\showISSN{10450823}


\bibitem[\protect\citeauthoryear{Yang, He, Gao, Deng, and Smola}{Yang
  et~al\mbox{.}}{2016}]%
        {Yang2016StackedAnswering}
\bibfield{author}{\bibinfo{person}{Zichao Yang}, \bibinfo{person}{Xiaodong He},
  \bibinfo{person}{Jianfeng Gao}, \bibinfo{person}{Li Deng}, {and}
  \bibinfo{person}{Alex Smola}.} \bibinfo{year}{2016}\natexlab{}.
\newblock \showarticletitle{{Stacked attention networks for image question
  answering}}. In \bibinfo{booktitle}{\emph{Proceedings of the IEEE Computer
  Society Conference on Computer Vision and Pattern Recognition}},
  Vol.~\bibinfo{volume}{2016-Decem}. \bibinfo{publisher}{IEEE Computer
  Society}, \bibinfo{pages}{21--29}.
\newblock
\showISBNx{9781467388504}
\showISSN{10636919}
\urldef\tempurl%
\url{https://doi.org/10.1109/CVPR.2016.10}
\showDOI{\tempurl}


\bibitem[\protect\citeauthoryear{You, Jin, Wang, Fang, and Luo}{You
  et~al\mbox{.}}{2016}]%
        {You2016ImageAttention}
\bibfield{author}{\bibinfo{person}{Quanzeng You}, \bibinfo{person}{Hailin Jin},
  \bibinfo{person}{Zhaowen Wang}, \bibinfo{person}{Chen Fang}, {and}
  \bibinfo{person}{Jiebo Luo}.} \bibinfo{year}{2016}\natexlab{}.
\newblock \showarticletitle{{Image captioning with semantic attention}}. In
  \bibinfo{booktitle}{\emph{Proceedings of the IEEE Computer Society Conference
  on Computer Vision and Pattern Recognition}},
  Vol.~\bibinfo{volume}{2016-Decem}. \bibinfo{publisher}{IEEE Computer
  Society}, \bibinfo{pages}{4651--4659}.
\newblock
\showISBNx{9781467388504}
\showISSN{10636919}
\urldef\tempurl%
\url{https://doi.org/10.1109/CVPR.2016.503}
\showDOI{\tempurl}


\bibitem[\protect\citeauthoryear{Yu, Yu, Fan, and Tao}{Yu
  et~al\mbox{.}}{2017}]%
        {Yu2017Multi-modalAnswering}
\bibfield{author}{\bibinfo{person}{Zhou Yu}, \bibinfo{person}{Jun Yu},
  \bibinfo{person}{Jianping Fan}, {and} \bibinfo{person}{Dacheng Tao}.}
  \bibinfo{year}{2017}\natexlab{}.
\newblock \showarticletitle{{Multi-modal Factorized Bilinear Pooling with
  Co-attention Learning for Visual Question Answering}}. In
  \bibinfo{booktitle}{\emph{Proceedings of the IEEE International Conference on
  Computer Vision}}, Vol.~\bibinfo{volume}{2017-Octob}.
  \bibinfo{publisher}{Institute of Electrical and Electronics Engineers Inc.},
  \bibinfo{pages}{1839--1848}.
\newblock
\showISBNx{9781538610329}
\showISSN{15505499}
\urldef\tempurl%
\url{https://doi.org/10.1109/ICCV.2017.202}
\showDOI{\tempurl}


\bibitem[\protect\citeauthoryear{Yue, Yang, Wang, Rahul, and Katabi}{Yue
  et~al\mbox{.}}{2020}]%
        {Yue2020BodyCompass:Signals}
\bibfield{author}{\bibinfo{person}{Shichao Yue}, \bibinfo{person}{Yuzhe Yang},
  \bibinfo{person}{Hao Wang}, \bibinfo{person}{Hariharan Rahul}, {and}
  \bibinfo{person}{DIna Katabi}.} \bibinfo{year}{2020}\natexlab{}.
\newblock \showarticletitle{{BodyCompass: Monitoring Sleep Posture with
  Wireless Signals}}.
\newblock \bibinfo{journal}{\emph{Proceedings of the ACM on Interactive,
  Mobile, Wearable and Ubiquitous Technologies}} \bibinfo{volume}{4},
  \bibinfo{number}{2} (\bibinfo{date}{6} \bibinfo{year}{2020}),
  \bibinfo{pages}{1--25}.
\newblock
\showISSN{24749567}
\urldef\tempurl%
\url{https://doi.org/10.1145/3397311}
\showDOI{\tempurl}


\bibitem[\protect\citeauthoryear{Zeng, Guo, Duan, and Wu}{Zeng
  et~al\mbox{.}}{2021}]%
        {Zeng2021Multi-levelCounting}
\bibfield{author}{\bibinfo{person}{Xin Zeng}, \bibinfo{person}{Qiang Guo},
  \bibinfo{person}{Haoran Duan}, {and} \bibinfo{person}{Yunpeng Wu}.}
  \bibinfo{year}{2021}\natexlab{}.
\newblock \showarticletitle{{Multi-level features extraction network with
  gating mechanism for crowd counting}}.
\newblock \bibinfo{journal}{\emph{IET Image Processing}}
  (\bibinfo{year}{2021}).
\newblock
\urldef\tempurl%
\url{https://doi.org/10.1049/IPR2.12304}
\showDOI{\tempurl}


\bibitem[\protect\citeauthoryear{Zhai, Perez-Pozuelo, Clifton, Palotti, and
  Guan}{Zhai et~al\mbox{.}}{2020}]%
        {Zhai2020MakingSensing}
\bibfield{author}{\bibinfo{person}{Bing Zhai}, \bibinfo{person}{Ignacio
  Perez-Pozuelo}, \bibinfo{person}{Emma~A.D. Clifton}, \bibinfo{person}{Joao
  Palotti}, {and} \bibinfo{person}{Yu Guan}.} \bibinfo{year}{2020}\natexlab{}.
\newblock \showarticletitle{{Making Sense of Sleep: Multimodal Sleep Stage
  Classification in a Large, Diverse Population Using Movement and Cardiac
  Sensing}}.
\newblock \bibinfo{journal}{\emph{Proceedings of the ACM on Interactive,
  Mobile, Wearable and Ubiquitous Technologies}} \bibinfo{volume}{4},
  \bibinfo{number}{2} (\bibinfo{date}{6} \bibinfo{year}{2020}),
  \bibinfo{pages}{1--33}.
\newblock
\showISSN{24749567}
\urldef\tempurl%
\url{https://doi.org/10.1145/3397325}
\showDOI{\tempurl}


\bibitem[\protect\citeauthoryear{Zhang, Yang, He, and Deng}{Zhang
  et~al\mbox{.}}{2020}]%
        {Zhang2020MultimodalApplications}
\bibfield{author}{\bibinfo{person}{Chao Zhang}, \bibinfo{person}{Zichao Yang},
  \bibinfo{person}{Xiaodong He}, {and} \bibinfo{person}{Li Deng}.}
  \bibinfo{year}{2020}\natexlab{}.
\newblock \showarticletitle{{Multimodal Intelligence: Representation Learning,
  Information Fusion, and Applications}}.
\newblock \bibinfo{journal}{\emph{IEEE Journal on Selected Topics in Signal
  Processing}} \bibinfo{volume}{14}, \bibinfo{number}{3} (\bibinfo{date}{3}
  \bibinfo{year}{2020}), \bibinfo{pages}{478--493}.
\newblock
\showISSN{19410484}
\urldef\tempurl%
\url{https://doi.org/10.1109/JSTSP.2020.2987728}
\showDOI{\tempurl}


\bibitem[\protect\citeauthoryear{Zhang, Cui, Mueller, Tao, Kim, Rueschman,
  Mariani, Mobley, and Redline}{Zhang et~al\mbox{.}}{2018}]%
        {Zhang2018TheCommons}
\bibfield{author}{\bibinfo{person}{Guo~Qiang Zhang}, \bibinfo{person}{Licong
  Cui}, \bibinfo{person}{Remo Mueller}, \bibinfo{person}{Shiqiang Tao},
  \bibinfo{person}{Matthew Kim}, \bibinfo{person}{Michael Rueschman},
  \bibinfo{person}{Sara Mariani}, \bibinfo{person}{Daniel Mobley}, {and}
  \bibinfo{person}{Susan Redline}.} \bibinfo{year}{2018}\natexlab{}.
\newblock \showarticletitle{{The National Sleep Research Resource: Towards a
  sleep data commons}}.
\newblock \bibinfo{journal}{\emph{Journal of the American Medical Informatics
  Association}} \bibinfo{volume}{25}, \bibinfo{number}{10} (\bibinfo{date}{10}
  \bibinfo{year}{2018}), \bibinfo{pages}{1351--1358}.
\newblock
\showISSN{1527974X}
\urldef\tempurl%
\url{https://doi.org/10.1093/jamia/ocy064}
\showDOI{\tempurl}


\end{thebibliography}
\appendix

\section{HYPERPARAMETERS TUNING AND RESULTS}
The hyperparameter tuning was performed based on the designed backbone network from three to five convolutional blocks (7-13 convolutional layers). The first two blocks consisted of two convolutional layers. The third, fourth and fifth convolutional blocks consisted of three convolutional layers. The hyperparameter search aimed to reduce the search space and maintain suitable temporal lengths of the latent features. The hyperparameter tuning only focused on the number of kernels for each convolutional block. The convolutional layer kernel length has been investigated in the previous study~\cite{Zhai2020MakingSensing}. We set the kernel length of all convolutional layers to 3. 

The number of hidden units in the fully connected layers was all set to the same value during the hyperparameter tuning process to reduce the search space. Furthermore, we performed the hyperparameter tuning based on the MESA dataset - the largest dataset containing the cardiac and activity data to date. Therefore, we expected the hyperparameter tuning could discover robust backbone networks for this study. 
\label{apx:hyper-parameter tuning}
\begin{figure*}[!ht]
\begin{tabular}{cc}
\rotatebox[origin=c]{90}{} &
\includegraphics[width=15cm,valign=c]{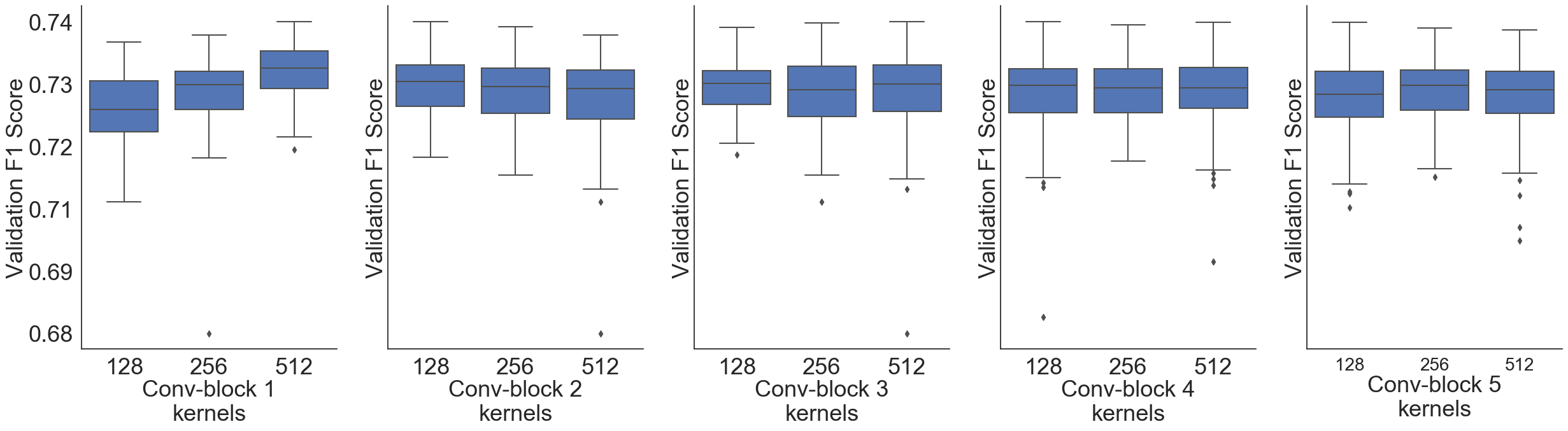}\\
    \addlinespace
\rotatebox[origin=c]{90}{} &
\includegraphics[width=15cm,valign=c]{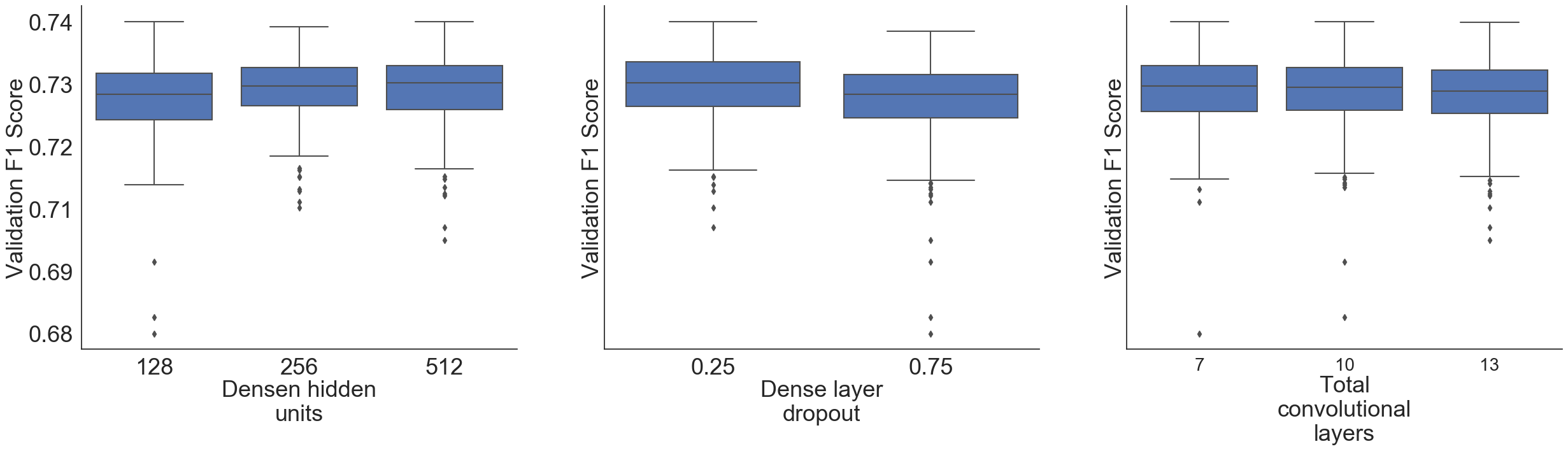}\\
    \addlinespace

\end{tabular}
\caption{DeepCNN backbone network hyper-parameters tuning results}
\label{fig:cnn_hp_plot}%
\end{figure*}

\begin{table}[h!]
  \centering
  \caption{Hyper-parameters tuning for backbone networks}
    \label{tab:hyperparameters}
    \centering
    \resizebox{\textwidth}{!}{ 
\begin{tabular}{c|c|c|c|c|c|c|c}
\toprule
\textbf{Block/Group} & \textbf{Layer } & \textbf{Kernel size} & \textbf{Number of Kernels} & \textbf{Padding} & \textbf{ Stride} & \textbf{Hidden Units} & \textbf{Drop out Rate} \\
\midrule
\multirow{2}[4]{*}{Convolutional Block 1} & Convolutional Layer 1 \& 2 & 3 & 128, 256, 512 & 1 & 1 &   &  \\
\cmidrule{2-8}  & Maxpooling & 2 &   &   & 2 &   &  \\
\midrule
\multirow{2}[4]{*}{Convolutional Block 2} & Convolutional Layer 3 \& 4 & 3 & 128, 256, 512 & 1 & 1 &   &  \\
\cmidrule{2-8}  & Maxpooling & 2 &   &   & 2 &   &  \\
\midrule
Convolutional Block 3 & Convolutional Layer 5, 6 \& 7 & 3 & 128, 256, 512 & 1 & 1 &   &  \\
\midrule
  & Maxpooling & 2 &   &   & 2 &   &  \\
\midrule
Convolutional Block 4 & Convolutional Layer 8, 9 \& 10 & 3 & 128, 256, 512 & 1 & 1 &   &  \\
\midrule
  & Maxpooling & 2 &   &   & 2 &   &  \\
\midrule
Convolutional Block 4 & Convolutional Layer 11, 12 \& 13 & 3 & 128, 256, 512 & 1 & 1 &   &  \\
\midrule
  & Maxpooling & 2 &   &   & 2 &   &  \\
\midrule
\multirow{2}[4]{*}{FC Block} & Fully Connected Layer 1, 2 &   &   &   &   & 128, 256, 512 &  \\
\cmidrule{2-8}  & Drop Out &   &   &   &   &   & 0.25, 0.75 \\
\bottomrule
\end{tabular}%

    }
\end{table}%

\section{HEART RATE STATISTIC FEATURES COMBINED WITH DEEP MOVEMENT FEATURES}

\begin{figure}[h]
    \centering
    \includegraphics[width=\linewidth]{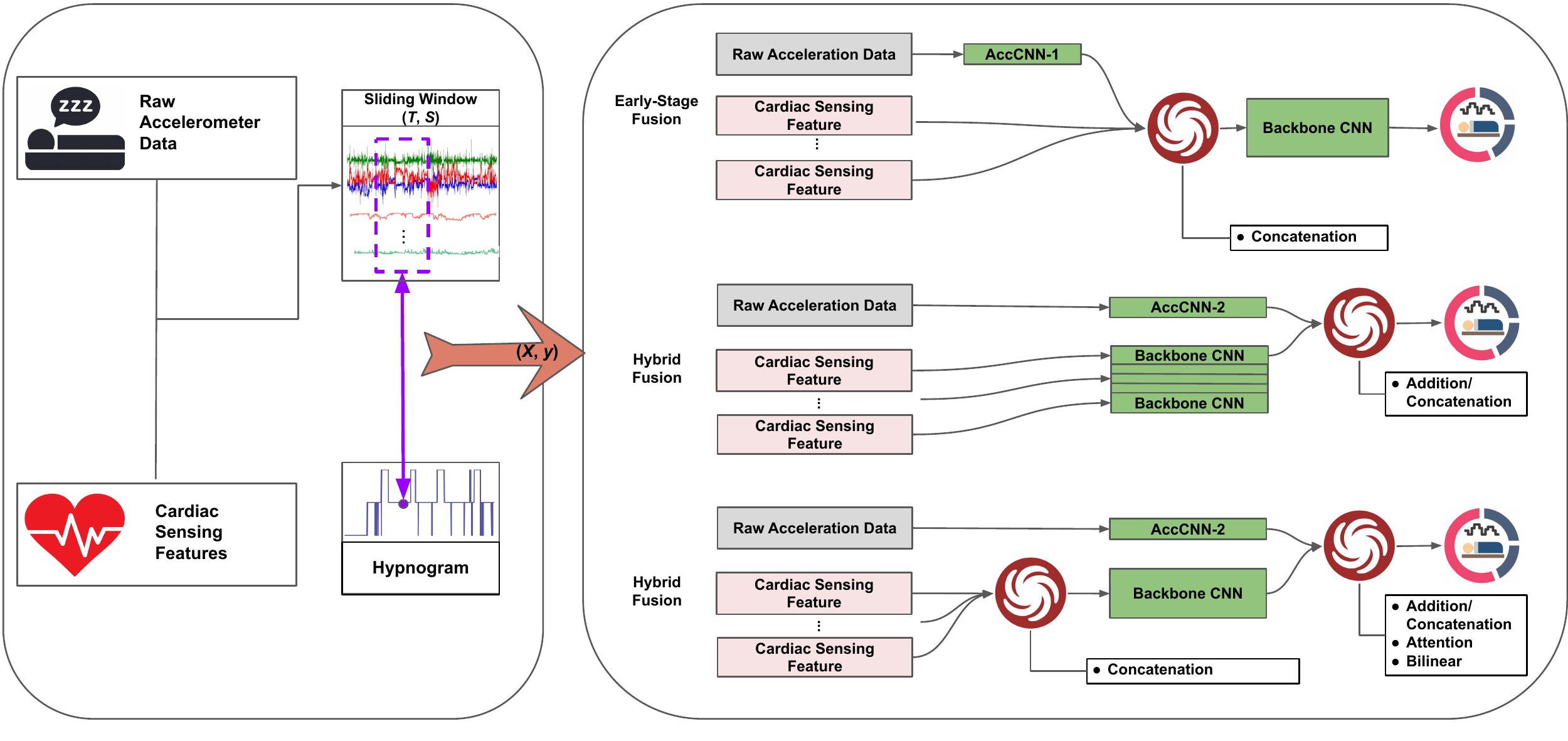}
	\caption{An overview of the three-stage sleep classification system using the raw accelerometer data with HR statistics features. The raw accelerometer data and HR statistic features were extracted for each sleep epoch (30s). The sliding window method divides the sleep data into multiple segments with window length $T$ and stride $S$. In this experiment, we have $T=101$, and $S=1$. We firstly use the AccCNN to learn deep features then fuse them with HR statistic features.  The hypnogram represents the stages of sleep over time. Two fusion strategies and four fusion methods were studied.}
    \label{fig:raw_system_overview}
\end{figure}

\begin{figure}[h]
    \centering
    \includegraphics[width=\linewidth]{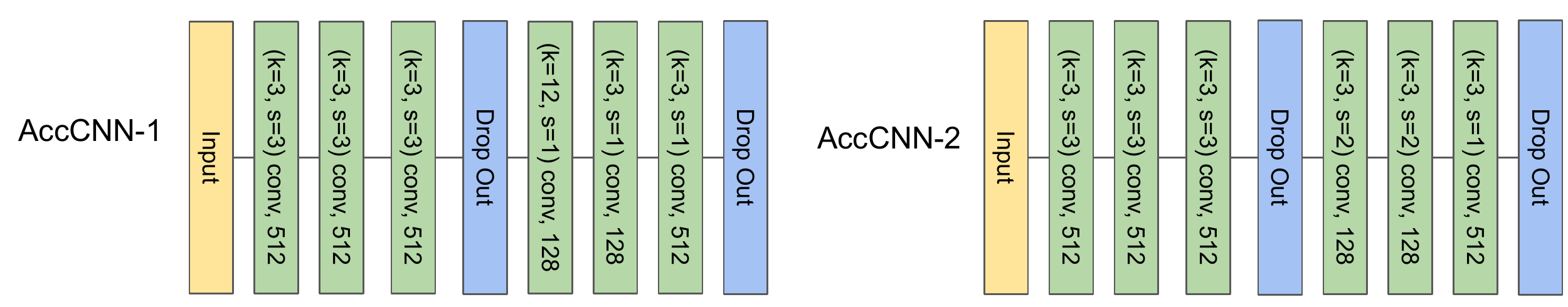}
	\caption{An overview of the two subnets used to extract the deep features from the raw accelerometer data. }
    \label{fig:acc_cnn_network_structure}
\end{figure}

In our study, we also tested whether using the raw accelerometer data could achieve better results. Therefore, we designed two feasible CNNs that could extract the compatible deep features fused with the HR statistic features.The rational behind the network design was to produce a compatible representation of the HR intermediate feature. To match up the dimension of latent feature, we firstly reduced the accelerometer data sampling rate from 50Hz/20Hz to 1Hz. We then designed two CNNs to bridge the sampling gap between movement and cardiac sensing features, one for the early stage fusion and another for the late-stage and hybrid fusion. Each consisted of six convolutional layers (two convolutional blocks) to extract the deep movement feature used for fusion study.  For the early-stage fusion, the network was called AccCNN-1. The hybrid and late-stage fusion used the same network to extract the latent representations, and we called it AccCNN-2. We referred to the entire network as DeepMixCNN and ResDeepMixCNN, respectively. The network structure and the experiment setting details can be seen in Figure~\ref{fig:raw_system_overview} and Figure~\ref{fig:acc_cnn_network_structure}.
We adopted the leave-two-subjects-out cross validation experimental setting on Apple Watch dataset. The training, validation and testing process used the same settings as the main content. However, we did not conduct the hyperparameter search together with the backbone network. Therefore, the network designed in the study merely served as feasible networks for the study, yet it might not be the best performing CNN.  We focused on the fusion techniques rather than the contribution of network structure.

The MESA dataset contained the activity counts sampled at 1/30 Hz, which technically was no the raw data. In addition, cardiac sensing was acquired via the PSG equipment, which may be difficult to wear everyday. Therefore, the HRV features derived from the RR intervals were most likely to be available from the commercial wearable devices (e.g., photoplethysmogram data), so we did not conduct the experiments on the raw PSG data.The details of this experiment are listed in Table~\ref{tab:raw_data_feature_extraction}

\subsection{Raw Accelerometer Data and HRS Features}

\begin{table}[h]
  \centering
  \caption{
  Three-stage sleep classification results (mean $\pm$ standard error at 95\% confidence interval) using raw accelerometer data and HRS features based on DeepMixCNN and ResDeepMixCNN with the Apple Watch Dataset for each combination of fusion strategy and method. The experiments were performed using the same experimental setting as in the main content and evaluated at the subject level during recording period based on window length of 101.
  }
  \resizebox{0.95\textwidth}{!}{%
\begin{tabular}{c|c|c|c|c|c|c|c}
\toprule
\multicolumn{3}{c|}{\textbf{Fusion Specifics}} & \multicolumn{3}{c|}{\textbf{Performance Metrics}} & \multicolumn{2}{c}{\textbf{Deployment Metrics}} \\
\midrule
\textbf{Fusion Strategy} & \textbf{Network} & \textbf{Fusion Method} & \textbf{Accuracy (\%)} & \textbf{Cohen's $\kappa$} & \textbf{Mean F1 (\%)} & \textbf{Model Size (M)} & \textbf{Inference Time (ms)} \\
\midrule
\multicolumn{1}{c|}{\multirow{2}[4]{*}{Early-Stage}} & DeepMixCNN & Concatenation & 74.8 $\pm$ 2.7 & 34.0 $\pm$ 5.9 & 57.2 $\pm$ 3.5 & 11.9 & 18.86$\pm$1.19 \\
\cmidrule{2-8}  & ResDeepMixCNN & Concatenation & 79.8 $\pm$ 3.1 & 48.9 $\pm$ 7.1 & 64.5 $\pm$ 4.5 & 11.9 & 16.45$\pm$0.43 \\
\midrule
\multicolumn{1}{c|}{\multirow{4}[4]{*}{Late-Stage Fusion}} & \multirow{2}[2]{*}{DeepMixCNN} & Concatenation & 79.2 $\pm$ 2.8 & 48.9 $\pm$ 7.6 & 66.0 $\pm$ 4.6 & 50.8 & 37.03$\pm$0.34 \\
  &   & Addition & 76.7 $\pm$ 2.1 & 38.7 $\pm$ 5.9 & 58.0 $\pm$ 3.7 & 11.5 & 32.75$\pm$0.19 \\
\cmidrule{2-8}  & \multirow{2}[2]{*}{ResDeepMixCNN} & Concatenation & \boldmath{}\textbf{79.1 $\pm$ 3.3}\unboldmath{} & \boldmath{}\textbf{51.4 $\pm$ 8.0}\unboldmath{} & \boldmath{}\textbf{66.7 $\pm$ 4.7}\unboldmath{} & 50.8 & 31.19$\pm$0.72 \\
  &   & Addition & 78.9 $\pm$ 2.8 & 48.4 $\pm$ 6.4 & 63.9 $\pm$ 4.0 & 11.5 & 33.11$\pm$0.33 \\
\midrule
\multicolumn{1}{c|}{\multirow{10}[4]{*}{Hybrid}} & \multirow{5}[2]{*}{DeepMixCNN} & Concatenation & 77.3 $\pm$ 3.3 & 43.9 $\pm$ 7.3 & 63.6 $\pm$ 4.3 & 18.0 & 15.94$\pm$0.22 \\
  &   & Addition & 75.8 $\pm$ 2.6 & 36.7 $\pm$ 7.3 & 59.0 $\pm$ 4.0 & 11.5 & 16.28$\pm$0.57 \\
  &   & Attention-on-Mov & 75.8 $\pm$ 3.1 & 39.4 $\pm$ 7.9 & 60.8 $\pm$ 4.8 & 18.3 & 15.7$\pm$0.18 \\
  &   & Attention-on-Car & 71.9 $\pm$ 3.1 & 30.4 $\pm$ 8.1 & 55.7 $\pm$ 4.8 & 18.3 & 15.7$\pm$0.18 \\
  &   & Bilinear & 73.1 $\pm$ 3.4 & 31.4 $\pm$ 8.2 & 52.3 $\pm$ 5.2 & 273.9 & 18.89$\pm$0.43 \\
\cmidrule{2-8}  & \multirow{5}[2]{*}{ResDeepMixCNN} & Concatenation & 77.7 $\pm$ 2.6 & 42.8 $\pm$ 6.0 & 62.6 $\pm$ 3.5 & 18.0 & 15.6$\pm$0.48 \\
  &   & Addition & 80.3 $\pm$ 2.9 & 48.0 $\pm$ 7.3 & 64.4 $\pm$ 4.3 & 11.5 & 15.65$\pm$0.31 \\
  &   & Attention-on-Mov & 77.1 $\pm$ 2.4 & 44.8 $\pm$ 5.9 & 62.7 $\pm$ 3.6 & 18.3 & 16.24$\pm$0.32 \\
  &   & Attention-on-Car & 75.9 $\pm$ 3.0 & 37.0 $\pm$ 7.7 & 57.7 $\pm$ 4.4 & 18.3 & 16.24$\pm$0.32 \\
  &   & Bilinear & 72.8 $\pm$ 3.2 & 29.8 $\pm$ 6.5 & 52.1 $\pm$ 4.4 & 273.9 & 18.88$\pm$0.4 \\
\bottomrule
\end{tabular}%

}
  \label{tab:raw_data_feature_extraction}%
\end{table}%
. The highest performed model was ResDeepMixCNN in late-stage fusion, using the concatenation method. Its accuracy, the Cohen's $\kappa$ score and the mean F1 reached 79.1 \%, 51.4  and 66.7 \% respectively. Thus, the results were comparable to the handcraft features.

\subsection{Comparison of Raw Data and Intermediate Features}
We compared the performance difference between using the raw accelerometer data and using the clinical/handcraft features based on the window length of 101. The ResDeepMixCNN has achieved the comparable performance on the Apple Watch dataset in terms of accuracy, Cohen's $\kappa$ and mean F1, using the concatenation method in the late-stage fusion. The confusion matrices shown in Figure~\ref{fig:confusion_matrix_2} demonstrated the model prediction using raw accelerometer data is biased to NREM sleep. Three reasons might cause the increased bias. The first reason may be the Apple Watch dataset has class imbalance issue. The second reason may be the modality bias of the raw accelerometer data because the wrist movement may not reflect the sleep stage (mainly NREM and REM) that much.  The third reason may be caused by the lacking of hyperparameter search on the network. 

Our observations corroborate a study using raw PSG signals for sleep stage classification~\cite{Phan2021XSleepNet:Staging}. That is using intermediate features instead of raw accelerometer data may alleviate the modality bias in the three-sleep stage classification task, while reducing the model parameters.
\label{apx:raw_acc_feature_hrs_experiment}
\section{THREE SLEEP STAGE CLASSIFICATION PERFORMANCE ON 21, 51 WINDOW LENGTH}
\subsection{The Effects of Sliding Windows Length}
In addition to the window length of 101, we also conducted experiments based on the window lengths 51 and 21 followed the previous work~\cite{Zhai2020MakingSensing}. For the Apple Watch dataset, the models with the highest mean F1, accuracy and Cohen's $\kappa$ score in each fusion strategy were all based on the window length of 101. For the MESA dataset, we observed similar patterns on all feature settings. One possible explanation was when the time step of the input data became shorter, the intermediate features around the time point of the prediction might not contain enough information for three-stage sleep classification. This phenomenon corroborated the previous findings~\cite{Zhai2020MakingSensing}.

\begin{table}[htbp]
  \centering
  \caption{Three-stage sleep classification results (mean $\pm$ standard error at 95\% confidence interval) for each combination of the fusion strategy and method with the Apple Watch dataset using the ACT-HRS feature and evaluated at subject level during the recording period based on the window length of 51.}
    \resizebox{0.9\textwidth}{!}{%

\begin{tabular}{c|c|c|c|c|c|c|c|c}
\toprule
\multicolumn{3}{c|}{\textbf{Fusion Specifics}} & \multicolumn{3}{c|}{\textbf{Performance Metrics}} & \multicolumn{3}{c}{\textbf{Time Deviation (min.)}} \\
\midrule
\textbf{Fusion Strategy} & \textbf{Network} & \textbf{Fusion Method} & \textbf{Accuracy (\%)} & \textbf{Cohen's $\kappa$} & \textbf{Mean F1 (\%)} & \textbf{Non-REM sleep} & \textbf{REM sleep} & \textbf{Wake} \\
\midrule
\multicolumn{1}{c|}{\multirow{2}[4]{*}{Early-Stage Fusion}} & DeepCNN & Concatenation & 73.1 $\pm$ 3.1 & 40.5 $\pm$ 6.7 & 58.7 $\pm$ 4.4 & 9.1 $\pm$ 22.5 & -0.1 $\pm$ 23.5 & -9.0 $\pm$ 8.8 \\
\cmidrule{2-9}  & ResDeepCNN & Concatenation & 75.4 $\pm$ 2.9 & 43.9 $\pm$ 6.4 & 61.1 $\pm$ 3.9 & 23.0 $\pm$ 24.1 & -12.5 $\pm$ 23.7 & -10.5 $\pm$ 7.2 \\
\midrule
\multicolumn{1}{c|}{\multirow{4}[4]{*}{Late-Stage Fusion}} & \multirow{2}[2]{*}{DeepCNN} & Concatenation & 74.1 $\pm$ 2.7 & 41.9 $\pm$ 6.1 & 59.8 $\pm$ 3.7 & 12.0 $\pm$ 21.6 & -1.1 $\pm$ 22.1 & -10.9 $\pm$ 7.9 \\
  &   & Addition & 78.0 $\pm$ 2.3 & 50.1 $\pm$ 7.0 & 65.5 $\pm$ 3.6 & 1.3 $\pm$ 16.7 & 11.6 $\pm$ 16.8 & -12.9 $\pm$ 6.6 \\
\cmidrule{2-9}  & \multirow{2}[2]{*}{ResDeepCNN} & Concatenation & 76.6 $\pm$ 2.5 & 45.9 $\pm$ 6.9 & 62.6 $\pm$ 4.2 & 20.1 $\pm$ 18.8 & -13.5 $\pm$ 19.6 & -6.7 $\pm$ 7.8 \\
  &   & Addition & \boldmath{}\textbf{77.7 $\pm$ 2.2}\unboldmath{} & \boldmath{}\textbf{48.1 $\pm$ 6.8}\unboldmath{} & \boldmath{}\textbf{64.6 $\pm$ 4.0}\unboldmath{} & 7.6 $\pm$ 19.5 & 3.4 $\pm$ 19.6 & -11.0 $\pm$ 5.8 \\
\midrule
\multicolumn{1}{c|}{\multirow{10}[4]{*}{Hybrid Fusion}} & \multirow{5}[2]{*}{DeepCNN} & Concatenation & 72.5 $\pm$ 3.2 & 39.0 $\pm$ 6.2 & 58.8 $\pm$ 3.9 & 7.5 $\pm$ 24.6 & 5.3 $\pm$ 24.6 & -12.8 $\pm$ 6.8 \\
  &   & Addition & 73.3 $\pm$ 3.0 & 39.2 $\pm$ 6.3 & 58.4 $\pm$ 3.6 & 17.1 $\pm$ 24.1 & -0.8 $\pm$ 24.1 & -16.3 $\pm$ 5.5 \\
  &   & Attention-on-Mov & 72.6 $\pm$ 3.0 & 39.0 $\pm$ 6.0 & 59.4 $\pm$ 3.4 & 15.1 $\pm$ 23.6 & -1.7 $\pm$ 23.6 & -13.4 $\pm$ 6.6 \\
  &   & Attention-on-Car & 72.7 $\pm$ 2.9 & 37.2 $\pm$ 6.4 & 58.3 $\pm$ 3.7 & 23.7 $\pm$ 21.0 & -8.6 $\pm$ 20.3 & -15.2 $\pm$ 6.1 \\
  &   & Bilinear & 72.3 $\pm$ 2.6 & 38.2 $\pm$ 5.5 & 58.5 $\pm$ 3.2 & 1.4 $\pm$ 20.7 & 12.3 $\pm$ 20.7 & -13.7 $\pm$ 6.3 \\
\cmidrule{2-9}  & \multirow{5}[2]{*}{ResDeepCNN} & Concatenation & 73.6 $\pm$ 2.9 & 41.5 $\pm$ 6.5 & 60.7 $\pm$ 4.1 & 8.8 $\pm$ 23.1 & 1.4 $\pm$ 24.9 & -10.2 $\pm$ 7.4 \\
  &   & Addition & 73.1 $\pm$ 3.1 & 40.5 $\pm$ 6.5 & 59.7 $\pm$ 3.9 & 11.5 $\pm$ 23.8 & -0.8 $\pm$ 24.4 & -10.7 $\pm$ 6.8 \\
  &   & Attention-on-Mov & 74.4 $\pm$ 3.3 & 43.8 $\pm$ 6.1 & 61.7 $\pm$ 3.8 & 11.6 $\pm$ 20.9 & -1.5 $\pm$ 21.5 & -10.1 $\pm$ 6.9 \\
  &   & Attention-on-Car & 73.1 $\pm$ 3.0 & 40.6 $\pm$ 7.1 & 60.4 $\pm$ 4.1 & 2.1 $\pm$ 24.0 & 6.9 $\pm$ 24.3 & -9.0 $\pm$ 7.9 \\
  &   & Bilinear & 74.9 $\pm$ 2.6 & 42.2 $\pm$ 6.5 & 60.2 $\pm$ 3.9 & 22.2 $\pm$ 21.9 & -12.2 $\pm$ 22.0 & -9.9 $\pm$ 7.1 \\
\bottomrule
\end{tabular}%

\label{tab:apple_win_50}%
  }
\end{table}%


\begin{table}[htbp]
  \centering
  \caption{Three-stage sleep classification results (mean $\pm$ standard error at 95\% confidence interval) for each combination of fusion strategy and method with the Apple Watch dataset using the ACT-HRS feature and evaluated at subject level during the recording period based on the window length of 21.}
    \resizebox{0.9\textwidth}{!}{%

\begin{tabular}{c|c|c|c|c|c|c|c|c}
\toprule
\multicolumn{3}{c|}{\textbf{Fusion Specifics}} & \multicolumn{3}{c|}{\textbf{Performance Metrics}} & \multicolumn{3}{c}{\textbf{Time Deviation (min.)}} \\
\midrule
\textbf{Fusion Strategy} & \textbf{Network} & \textbf{Fusion Method} & \textbf{Accuracy (\%)} & \textbf{Cohen's $\kappa$} & \textbf{Mean F1 (\%)} & \textbf{Non-REM sleep} & \textbf{REM sleep} & \textbf{Wake} \\
\midrule
\multicolumn{1}{c|}{\multirow{2}[4]{*}{Early-Stage Fusion}} & DeepCNN & Concatenation & 71.9 $\pm$ 2.1 & 35.2 $\pm$ 5.5 & 54.9 $\pm$ 3.5 & 22.9 $\pm$ 16.6 & -13.9 $\pm$ 16.8 & -9.0 $\pm$ 8.3 \\
\cmidrule{2-9}  & ResDeepCNN & Concatenation & 73.4 $\pm$ 2.4 & 38.7 $\pm$ 5.8 & 58.0 $\pm$ 3.6 & 17.3 $\pm$ 15.2 & -9.6 $\pm$ 16.0 & -7.7 $\pm$ 9.2 \\
\midrule
\multicolumn{1}{c|}{\multirow{4}[4]{*}{Late-Stage Fusion}} & \multirow{2}[2]{*}{DeepCNN} & Concatenation & 72.4 $\pm$ 2.5 & 38.1 $\pm$ 6.1 & 55.9 $\pm$ 3.9 & 21.9 $\pm$ 18.2 & -7.4 $\pm$ 18.0 & -14.6 $\pm$ 8.6 \\
  &   & Addition & \boldmath{}\textbf{75.3 $\pm$ 2.4}\unboldmath{} & \boldmath{}\textbf{39.7 $\pm$ 7.7}\unboldmath{} & \boldmath{}\textbf{59.2 $\pm$ 4.2}\unboldmath{} & 14.6 $\pm$ 13.9 & -1.4 $\pm$ 15.4 & -13.1 $\pm$ 7.2 \\
\cmidrule{2-9}  & \multirow{2}[2]{*}{ResDeepCNN} & Concatenation & 72.7 $\pm$ 2.6 & 38.3 $\pm$ 6.8 & 57.3 $\pm$ 4.2 & 11.9 $\pm$ 20.3 & -4.5 $\pm$ 21.0 & -7.4 $\pm$ 8.8 \\
  &   & Addition & 74.3 $\pm$ 2.7 & 38.9 $\pm$ 7.5 & 58.9 $\pm$ 4.2 & 6.2 $\pm$ 16.6 & 4.5 $\pm$ 17.6 & -10.7 $\pm$ 7.9 \\
\midrule
\multicolumn{1}{c|}{\multirow{10}[4]{*}{Hybrid Fusion}} & \multirow{5}[2]{*}{DeepCNN} & Concatenation & 71.6 $\pm$ 2.5 & 35.8 $\pm$ 6.2 & 55.6 $\pm$ 4.0 & 17.7 $\pm$ 23.5 & -0.0 $\pm$ 23.4 & -17.7 $\pm$ 6.4 \\
  &   & Addition & 71.6 $\pm$ 2.6 & 35.0 $\pm$ 5.7 & 55.8 $\pm$ 3.6 & 22.9 $\pm$ 19.7 & -5.9 $\pm$ 19.7 & -17.0 $\pm$ 6.7 \\
  &   & Attention-on-Mov & 72.2 $\pm$ 2.6 & 37.3 $\pm$ 5.4 & 56.6 $\pm$ 3.4 & 20.5 $\pm$ 19.6 & -3.6 $\pm$ 20.8 & -16.9 $\pm$ 6.3 \\
  &   & Attention-on-Car & 73.2 $\pm$ 2.3 & 35.4 $\pm$ 5.9 & 56.6 $\pm$ 3.6 & 31.1 $\pm$ 20.6 & -13.3 $\pm$ 20.1 & -17.8 $\pm$ 6.4 \\
  &   & Bilinear & 71.5 $\pm$ 2.9 & 37.3 $\pm$ 6.0 & 57.2 $\pm$ 3.7 & 5.0 $\pm$ 19.1 & 6.9 $\pm$ 17.7 & -11.9 $\pm$ 7.7 \\
\cmidrule{2-9}  & \multirow{5}[2]{*}{ResDeepCNN} & Concatenation & 72.3 $\pm$ 2.7 & 36.6 $\pm$ 6.5 & 57.3 $\pm$ 4.0 & 21.3 $\pm$ 23.8 & -5.3 $\pm$ 23.3 & -15.9 $\pm$ 6.5 \\
  &   & Addition & 71.6 $\pm$ 2.2 & 35.2 $\pm$ 5.2 & 55.5 $\pm$ 3.4 & 24.7 $\pm$ 18.2 & -11.6 $\pm$ 19.1 & -13.2 $\pm$ 6.9 \\
  &   & Attention-on-Mov & 73.3 $\pm$ 2.3 & 38.0 $\pm$ 5.4 & 57.9 $\pm$ 3.3 & 33.5 $\pm$ 17.3 & -19.3 $\pm$ 17.7 & -14.2 $\pm$ 6.4 \\
  &   & Attention-on-Car & 72.1 $\pm$ 2.7 & 37.8 $\pm$ 5.5 & 57.8 $\pm$ 3.6 & 17.5 $\pm$ 19.2 & -5.8 $\pm$ 18.8 & -11.7 $\pm$ 7.9 \\
  &   & Bilinear & 70.5 $\pm$ 2.7 & 35.7 $\pm$ 5.8 & 55.9 $\pm$ 3.6 & 3.1 $\pm$ 19.2 & 1.2 $\pm$ 19.5 & -4.3 $\pm$ 9.8 \\
\bottomrule
\end{tabular}%

  \label{tab:apple_win_20}%
  }
\end{table}%


\begin{table}[htbp]
  \centering
  \caption{Three-stage sleep classification results (mean $\pm$ standard error at 95\% confidence interval) for each combination of fusion strategy and method with the MESA test dataset using the ACT-HRS evaluated at subject level during the recording period based on the window length of 51.}
   \resizebox{0.9\textwidth}{!}{%

\begin{tabular}{c|c|c|c|c|c|c|c|c}
\toprule
\multicolumn{3}{c|}{\textbf{Fusion Specifics}} & \multicolumn{3}{c|}{\textbf{Performance Metrics}} & \multicolumn{3}{c}{\textbf{Time Deviation (min.)}} \\
\midrule
\textbf{Fusion Strategy} & \textbf{Network} & \textbf{Fusion Method} & \textbf{Accuracy (\%)} & \textbf{Cohen's $\kappa$} & \textbf{Mean F1 (\%)} & \textbf{Non-REM sleep} & \textbf{REM sleep} & \textbf{Wake} \\
\midrule
\multicolumn{1}{c|}{\multirow{2}[4]{*}{Early-Stage Fusion}} & DeepCNN & Concatenation & 78.1 $\pm$ 0.9 & 60.1 $\pm$ 1.8 & 69.3 $\pm$ 1.3 & 14.6 $\pm$ 7.3 & -18.7 $\pm$ 3.6 & 4.1 $\pm$ 6.9 \\
\cmidrule{2-9}  & ResDeepCNN & Concatenation & 76.7 $\pm$ 1.0 & 60.2 $\pm$ 1.9 & 70.3 $\pm$ 1.3 & -15.6 $\pm$ 7.4 & 12.6 $\pm$ 5.0 & 3.0 $\pm$ 6.8 \\
\midrule
\multicolumn{1}{c|}{\multirow{4}[4]{*}{Late-Stage Fusion}} & \multirow{2}[2]{*}{DeepCNN} & Concatenation & \boldmath{}\textbf{78.4 $\pm$ 1.0}\unboldmath{} & \boldmath{}\textbf{62.5 $\pm$ 1.8}\unboldmath{} & \boldmath{}\textbf{71.4 $\pm$ 1.3}\unboldmath{} & 9.9 $\pm$ 7.3 & 0.2 $\pm$ 4.1 & -10.0 $\pm$ 6.4 \\
  &   & Addition & 76.5 $\pm$ 0.9 & 58.8 $\pm$ 1.7 & 66.7 $\pm$ 1.2 & 62.3 $\pm$ 7.4 & -21.7 $\pm$ 3.7 & -40.6 $\pm$ 6.9 \\
\cmidrule{2-9}  & \multirow{2}[2]{*}{ResDeepCNN} & Concatenation & 77.7 $\pm$ 0.9 & 61.0 $\pm$ 1.8 & 69.9 $\pm$ 1.2 & 17.3 $\pm$ 7.6 & -4.6 $\pm$ 4.2 & -12.6 $\pm$ 6.6 \\
  &   & Addition & 74.8 $\pm$ 1.0 & 55.7 $\pm$ 1.8 & 66.2 $\pm$ 1.3 & -21.0 $\pm$ 7.7 & -16.0 $\pm$ 4.2 & 37.0 $\pm$ 7.5 \\
\midrule
\multicolumn{1}{c|}{\multirow{10}[4]{*}{Hybrid Fusion}} & \multirow{5}[2]{*}{DeepCNN} & Concatenation & 76.4 $\pm$ 1.1 & 61.1 $\pm$ 1.8 & 70.0 $\pm$ 1.3 & -9.4 $\pm$ 8.0 & 17.9 $\pm$ 5.0 & -8.5 $\pm$ 6.9 \\
  &   & Addition & 77.2 $\pm$ 1.0 & 61.3 $\pm$ 1.7 & 70.6 $\pm$ 1.2 & -17.9 $\pm$ 7.5 & 2.5 $\pm$ 4.3 & 15.4 $\pm$ 7.0 \\
  &   & Attention-on-Mov & 74.6 $\pm$ 1.1 & 59.1 $\pm$ 1.8 & 69.0 $\pm$ 1.2 & -24.0 $\pm$ 8.1 & 39.2 $\pm$ 5.7 & -15.3 $\pm$ 6.6 \\
  &   & Attention-on-Car & 77.8 $\pm$ 0.9 & 60.7 $\pm$ 1.8 & 69.8 $\pm$ 1.2 & 5.4 $\pm$ 7.4 & -9.2 $\pm$ 4.1 & 3.8 $\pm$ 6.7 \\
  &   & Bilinear & 77.1 $\pm$ 0.9 & 60.1 $\pm$ 1.8 & 70.2 $\pm$ 1.2 & 6.8 $\pm$ 8.0 & 13.9 $\pm$ 5.0 & -20.8 $\pm$ 6.5 \\
\cmidrule{2-9}  & \multirow{5}[2]{*}{ResDeepCNN} & Concatenation & 77.7 $\pm$ 1.0 & 62.4 $\pm$ 1.7 & 71.0 $\pm$ 1.2 & 8.3 $\pm$ 7.7 & 15.5 $\pm$ 5.0 & -23.8 $\pm$ 6.3 \\
  &   & Addition & 78.5 $\pm$ 0.9 & 62.0 $\pm$ 1.7 & 71.3 $\pm$ 1.2 & 30.4 $\pm$ 7.3 & 4.1 $\pm$ 4.3 & -34.5 $\pm$ 6.5 \\
  &   & Attention-on-Mov & 76.1 $\pm$ 1.1 & 60.7 $\pm$ 1.8 & 70.4 $\pm$ 1.3 & -9.9 $\pm$ 8.2 & 31.3 $\pm$ 5.7 & -21.4 $\pm$ 6.8 \\
  &   & Attention-on-Car & 76.6 $\pm$ 1.1 & 61.2 $\pm$ 1.8 & 70.7 $\pm$ 1.2 & -0.3 $\pm$ 7.6 & 25.8 $\pm$ 5.1 & -25.6 $\pm$ 6.2 \\
  &   & Bilinear & 76.7 $\pm$ 0.9 & 60.2 $\pm$ 1.7 & 69.7 $\pm$ 1.2 & -9.5 $\pm$ 7.6 & 9.4 $\pm$ 4.6 & 0.1 $\pm$ 6.6 \\
\bottomrule
\end{tabular}%

    }
  \label{tab:mesa_hrs_win_50}%
\end{table}%


\begin{table}[htbp]
  \centering
  \caption{Three-stage sleep classification results (mean $\pm$ standard error at 95\% confidence interval) for each combination of fusion strategy and method with the MESA test dataset using the ACT-HRS evaluated at subject level during the recording period based on the window length of 21.}
     \resizebox{0.9\textwidth}{!}{%
\begin{tabular}{c|c|c|c|c|c|c|c|c}
\toprule
\multicolumn{3}{c|}{\textbf{Fusion Specifics}} & \multicolumn{3}{c|}{\textbf{Performance Metrics}} & \multicolumn{3}{c}{\textbf{Time Deviation (min.)}} \\
\midrule
\textbf{Fusion Strategy} & \textbf{Network} & \textbf{Fusion Method} & \textbf{Accuracy (\%)} & \textbf{Cohen's $\kappa$} & \textbf{Mean F1 (\%)} & \textbf{Non-REM sleep} & \textbf{REM sleep} & \textbf{Wake} \\
\midrule
\multirow{2}[4]{*}{Early-Stage Fusion} & DeepCNN & Concatenation & 75.3 $\pm$ 1.0 & 55.2 $\pm$ 1.8 & 66.7 $\pm$ 1.2 & 61.4 $\pm$ 7.7 & -4.3 $\pm$ 4.2 & -57.0 $\pm$ 6.8 \\
\cmidrule{2-9}  & ResDeepCNN & Concatenation & 75.0 $\pm$ 0.9 & 56.6 $\pm$ 1.7 & 68.3 $\pm$ 1.1 & 13.3 $\pm$ 7.4 & 16.7 $\pm$ 4.6 & -30.0 $\pm$ 6.5 \\
\midrule
\multicolumn{1}{c|}{\multirow{4}[4]{*}{Late-Stage Fusion}} & \multirow{2}[2]{*}{DeepCNN} & Concatenation & 76.6 $\pm$ 0.9 & 57.8 $\pm$ 1.6 & 68.0 $\pm$ 1.2 & 29.8 $\pm$ 7.0 & -13.9 $\pm$ 3.7 & -15.9 $\pm$ 6.4 \\
  &   & Addition & 74.4 $\pm$ 0.9 & 56.9 $\pm$ 1.6 & 66.8 $\pm$ 1.1 & 13.7 $\pm$ 7.5 & 6.9 $\pm$ 4.8 & -20.7 $\pm$ 6.4 \\
\cmidrule{2-9}  & \multirow{2}[2]{*}{ResDeepCNN} & Concatenation & 75.9 $\pm$ 0.9 & 58.1 $\pm$ 1.6 & 68.8 $\pm$ 1.1 & 7.8 $\pm$ 7.1 & 6.8 $\pm$ 4.1 & -14.7 $\pm$ 6.4 \\
  &   & Addition & 74.7 $\pm$ 0.9 & 56.6 $\pm$ 1.6 & 66.6 $\pm$ 1.1 & 33.4 $\pm$ 7.4 & 2.7 $\pm$ 4.7 & -36.1 $\pm$ 6.4 \\
\midrule
\multicolumn{1}{c|}{\multirow{10}[4]{*}{Hybrid Fusion}} & \multirow{5}[2]{*}{DeepCNN} & Concatenation & 76.0 $\pm$ 0.9 & 57.1 $\pm$ 1.7 & 68.6 $\pm$ 1.2 & 33.8 $\pm$ 7.6 & 8.2 $\pm$ 4.4 & -42.0 $\pm$ 6.7 \\
  &   & Addition & 74.8 $\pm$ 1.0 & 55.7 $\pm$ 1.8 & 67.9 $\pm$ 1.2 & 22.6 $\pm$ 7.8 & 21.7 $\pm$ 4.9 & -44.3 $\pm$ 6.6 \\
  &   & Attention-on-Mov & 74.2 $\pm$ 1.0 & 56.8 $\pm$ 1.7 & 67.9 $\pm$ 1.2 & -25.4 $\pm$ 7.9 & 18.4 $\pm$ 4.7 & 7.0 $\pm$ 7.0 \\
  &   & Attention-on-Car & 76.1 $\pm$ 0.9 & 57.4 $\pm$ 1.6 & 68.3 $\pm$ 1.1 & 13.1 $\pm$ 7.5 & -1.0 $\pm$ 4.3 & -12.1 $\pm$ 6.7 \\
  &   & Bilinear & 76.9 $\pm$ 0.9 & 57.6 $\pm$ 1.7 & 68.0 $\pm$ 1.2 & 26.4 $\pm$ 7.2 & -20.1 $\pm$ 3.6 & -6.3 $\pm$ 6.8 \\
\cmidrule{2-9}  & \multirow{5}[2]{*}{ResDeepCNN} & Concatenation & 76.5 $\pm$ 0.9 & 57.6 $\pm$ 1.7 & 68.6 $\pm$ 1.2 & 47.2 $\pm$ 7.4 & -0.2 $\pm$ 4.1 & -47.0 $\pm$ 6.6 \\
  &   & Addition & 76.2 $\pm$ 0.9 & 57.7 $\pm$ 1.7 & 68.8 $\pm$ 1.2 & 25.0 $\pm$ 7.4 & 7.6 $\pm$ 4.3 & -32.6 $\pm$ 6.6 \\
  &   & Attention-on-Mov & 75.9 $\pm$ 0.9 & 58.1 $\pm$ 1.7 & \boldmath{}\textbf{69.0 $\pm$ 1.1}\unboldmath{} & -0.2 $\pm$ 7.5 & 10.0 $\pm$ 4.3 & -9.8 $\pm$ 6.7 \\
  &   & Attention-on-Car & 75.5 $\pm$ 0.9 & 58.2 $\pm$ 1.7 & 68.8 $\pm$ 1.2 & -5.1 $\pm$ 7.5 & 13.2 $\pm$ 4.6 & -8.2 $\pm$ 6.6 \\
  &   & Bilinear & \boldmath{}\textbf{77.0 $\pm$ 0.9}\unboldmath{} & \boldmath{}\textbf{58.7 $\pm$ 1.6}\unboldmath{} & 68.6 $\pm$ 1.1 & 39.9 $\pm$ 7.1 & -13.1 $\pm$ 3.9 & -26.8 $\pm$ 6.5 \\
\bottomrule
\end{tabular}%

}
  \label{tab:mesa_hrs_win_20}%
\end{table}%

\begin{table}[htbp]
  \centering
  \caption{Three-stage sleep classification results (mean $\pm$ standard error at 95\% confidence interval) for each combination of fusion strategy and method in the MESA test dataset using the ACT-HRV feature set evaluated at subject level during the recording period based on the window length of 51.}
  \resizebox{0.9\textwidth}{!}{%
\begin{tabular}{c|c|c|c|c|c|c|c|c}
\toprule
\multicolumn{3}{c|}{\textbf{Fusion Specifics}} & \multicolumn{3}{c|}{\textbf{Performance Metrics}} & \multicolumn{3}{c}{\textbf{Time Deviation (min.)}} \\
\midrule
\textbf{Fusion Strategy} & \textbf{Network} & \textbf{Fusion Method} & \textbf{Accuracy (\%)} & \textbf{Cohen's $\kappa$} & \textbf{Mean F1 (\%)} & \textbf{Non-REM sleep} & \textbf{REM sleep} & \textbf{Wake} \\
\midrule
\multirow{2}[4]{*}{Early-Stage Fusion} & DeepCNN & Concatenation & 75.7 $\pm$ 1.0 & 56.7 $\pm$ 1.8 & 67.2 $\pm$ 1.3 & 40.0 $\pm$ 6.9 & -11.1 $\pm$ 3.9 & -28.9 $\pm$ 6.5 \\
\cmidrule{2-9}  & ResDeepCNN & Concatenation & 76.0 $\pm$ 0.9 & 56.9 $\pm$ 1.8 & 66.8 $\pm$ 1.3 & 63.1 $\pm$ 7.3 & -17.3 $\pm$ 3.6 & -45.9 $\pm$ 6.8 \\
\midrule
\multicolumn{1}{c|}{\multirow{4}[4]{*}{Late-Stage Fusion}} & \multirow{2}[2]{*}{DeepCNN} & Concatenation & 78.4 $\pm$ 0.9 & 61.7 $\pm$ 1.8 & 70.2 $\pm$ 1.3 & 20.8 $\pm$ 6.9 & -10.9 $\pm$ 3.7 & -9.8 $\pm$ 6.4 \\
  &   & Addition & 77.6 $\pm$ 0.9 & 60.5 $\pm$ 1.8 & 69.2 $\pm$ 1.2 & 34.3 $\pm$ 6.9 & -8.6 $\pm$ 3.9 & -25.7 $\pm$ 6.5 \\
\cmidrule{2-9}  & \multirow{2}[2]{*}{ResDeepCNN} & Concatenation & 78.0 $\pm$ 1.0 & \boldmath{}\textbf{62.4 $\pm$ 1.7}\unboldmath{} & \boldmath{}\textbf{71.1 $\pm$ 1.2}\unboldmath{} & 2.8 $\pm$ 7.2 & 3.3 $\pm$ 4.1 & -6.1 $\pm$ 6.4 \\
  &   & Addition & 77.5 $\pm$ 0.9 & 61.1 $\pm$ 1.8 & 70.3 $\pm$ 1.2 & 16.8 $\pm$ 6.9 & 2.4 $\pm$ 4.0 & -19.2 $\pm$ 6.5 \\
\midrule
\multicolumn{1}{c|}{\multirow{10}[4]{*}{Hybrid Fusion}} & \multirow{5}[2]{*}{DeepCNN} & Concatenation & 77.2 $\pm$ 1.0 & 60.3 $\pm$ 1.7 & 69.7 $\pm$ 1.3 & 12.5 $\pm$ 7.5 & -0.5 $\pm$ 4.4 & -12.0 $\pm$ 6.5 \\
  &   & Addition & 76.6 $\pm$ 1.0 & 59.7 $\pm$ 1.8 & 69.7 $\pm$ 1.3 & 14.2 $\pm$ 7.4 & 13.4 $\pm$ 4.6 & -27.6 $\pm$ 6.5 \\
  &   & Attention-on-Mov & 77.5 $\pm$ 1.0 & 61.2 $\pm$ 1.7 & 70.5 $\pm$ 1.3 & 35.3 $\pm$ 7.3 & 5.7 $\pm$ 4.8 & -40.9 $\pm$ 6.4 \\
  &   & Attention-on-Car & 77.8 $\pm$ 0.9 & 60.4 $\pm$ 1.7 & 68.5 $\pm$ 1.3 & 45.2 $\pm$ 7.1 & -20.6 $\pm$ 3.8 & -24.7 $\pm$ 6.4 \\
  &   & Bilinear & 77.1 $\pm$ 1.0 & 60.8 $\pm$ 1.7 & 70.6 $\pm$ 1.2 & 5.4 $\pm$ 6.9 & 12.2 $\pm$ 4.4 & -17.7 $\pm$ 6.2 \\
\cmidrule{2-9}  & \multirow{5}[2]{*}{ResDeepCNN} & Concatenation & 76.7 $\pm$ 1.1 & 60.7 $\pm$ 1.9 & 70.6 $\pm$ 1.3 & 6.2 $\pm$ 7.9 & 23.4 $\pm$ 5.4 & -29.6 $\pm$ 6.3 \\
  &   & Addition & 76.7 $\pm$ 1.1 & 61.3 $\pm$ 1.8 & 70.9 $\pm$ 1.3 & -27.4 $\pm$ 7.2 & 23.5 $\pm$ 4.8 & 3.8 $\pm$ 6.6 \\
  &   & Attention-on-Mov & \boldmath{}\textbf{78.7 $\pm$ 0.9}\unboldmath{} & 62.2 $\pm$ 1.7 & 70.0 $\pm$ 1.3 & 43.9 $\pm$ 6.9 & -17.6 $\pm$ 3.8 & -26.3 $\pm$ 6.4 \\
  &   & Attention-on-Car & 78.3 $\pm$ 0.9 & 62.2 $\pm$ 1.7 & 70.6 $\pm$ 1.3 & 37.7 $\pm$ 7.1 & -4.7 $\pm$ 4.4 & -33.0 $\pm$ 6.2 \\
  &   & Bilinear & 76.8 $\pm$ 1.0 & 59.1 $\pm$ 1.8 & 69.6 $\pm$ 1.2 & 24.0 $\pm$ 7.0 & 6.7 $\pm$ 4.2 & -30.6 $\pm$ 6.2 \\
\bottomrule
\end{tabular}%

    }
  \label{tab:mesa_hrv_win_50}%

\end{table}%

\begin{table}[htbp]
  \centering
  \caption{Three-stage sleep classification results (mean $\pm$ standard error at 95\% confidence interval) for each combination of fusion strategy and method with the MESA test dataset using the ACT-HRV evaluated at subject level during the recording period based on the window length of 21.}
  \resizebox{0.9\textwidth}{!}{%
\begin{tabular}{c|c|c|c|c|c|c|c|c}
\toprule
\multicolumn{3}{c|}{\textbf{Fusion Specifics}} & \multicolumn{3}{c|}{\textbf{Performance Metrics}} & \multicolumn{3}{c}{\textbf{Time Deviation (min.)}} \\
\midrule
\textbf{Fusion Strategy} & \textbf{Network} & \textbf{Fusion Method} & \textbf{Accuracy (\%)} & \textbf{Cohen's $\kappa$} & \textbf{Mean F1 (\%)} & \textbf{Non-REM sleep} & \textbf{REM sleep} & \textbf{Wake} \\
\midrule
\multirow{2}[4]{*}{Early-Stage Fusion} & DeepCNN & Concatenation & 75.9 $\pm$ 0.9 & 57.0 $\pm$ 1.7 & 67.5 $\pm$ 1.2 & 34.7 $\pm$ 7.0 & -10.2 $\pm$ 3.7 & -24.6 $\pm$ 6.6 \\
\cmidrule{2-9}  & ResDeepCNN & Concatenation & 75.4 $\pm$ 0.9 & 56.0 $\pm$ 1.7 & 67.2 $\pm$ 1.2 & 8.5 $\pm$ 7.1 & -10.9 $\pm$ 3.7 & 2.4 $\pm$ 6.7 \\
\midrule
\multicolumn{1}{c|}{\multirow{4}[4]{*}{Late-Stage Fusion}} & \multirow{2}[2]{*}{DeepCNN} & Concatenation & 76.0 $\pm$ 0.9 & 57.5 $\pm$ 1.7 & 68.0 $\pm$ 1.1 & 12.1 $\pm$ 7.1 & -4.5 $\pm$ 3.9 & -7.6 $\pm$ 6.5 \\
  &   & Addition & 74.4 $\pm$ 0.9 & 54.4 $\pm$ 1.7 & 64.0 $\pm$ 1.2 & 26.7 $\pm$ 7.7 & -21.6 $\pm$ 3.8 & -5.1 $\pm$ 7.2 \\
\cmidrule{2-9}  & \multirow{2}[2]{*}{ResDeepCNN} & Concatenation & 76.2 $\pm$ 0.9 & 58.2 $\pm$ 1.7 & 68.0 $\pm$ 1.2 & 16.7 $\pm$ 6.9 & -7.1 $\pm$ 3.8 & -9.6 $\pm$ 6.5 \\
  &   & Addition & 75.3 $\pm$ 0.9 & 56.0 $\pm$ 1.7 & 65.4 $\pm$ 1.2 & 38.7 $\pm$ 7.6 & -19.2 $\pm$ 3.8 & -19.5 $\pm$ 7.0 \\
\midrule
\multicolumn{1}{c|}{\multirow{10}[4]{*}{Hybrid Fusion}} & \multirow{5}[2]{*}{DeepCNN} & Concatenation & 75.8 $\pm$ 1.0 & 57.5 $\pm$ 1.7 & 68.5 $\pm$ 1.2 & 21.2 $\pm$ 7.7 & 3.6 $\pm$ 4.4 & -24.8 $\pm$ 7.0 \\
  &   & Addition & 75.2 $\pm$ 1.0 & 57.0 $\pm$ 1.8 & 68.5 $\pm$ 1.2 & 11.3 $\pm$ 7.4 & 21.4 $\pm$ 5.2 & -32.7 $\pm$ 6.7 \\
  &   & Attention-on-Mov & 75.4 $\pm$ 1.0 & 56.7 $\pm$ 1.7 & 67.7 $\pm$ 1.2 & 15.3 $\pm$ 7.5 & 3.9 $\pm$ 4.9 & -19.3 $\pm$ 6.7 \\
  &   & Attention-on-Car & 74.3 $\pm$ 1.0 & 56.6 $\pm$ 1.7 & 67.8 $\pm$ 1.2 & -14.6 $\pm$ 7.5 & 14.7 $\pm$ 4.6 & -0.1 $\pm$ 6.5 \\
  &   & Bilinear & 74.9 $\pm$ 1.0 & 57.4 $\pm$ 1.8 & 68.3 $\pm$ 1.2 & -13.4 $\pm$ 7.9 & 8.6 $\pm$ 4.3 & 4.7 $\pm$ 7.2 \\
\cmidrule{2-9}  & \multirow{5}[2]{*}{ResDeepCNN} & Concatenation & 75.5 $\pm$ 1.0 & 57.1 $\pm$ 1.8 & 68.7 $\pm$ 1.2 & 19.8 $\pm$ 7.4 & 20.8 $\pm$ 5.0 & -40.6 $\pm$ 6.6 \\
  &   & Addition & 76.2 $\pm$ 0.9 & 58.4 $\pm$ 1.7 & 69.0 $\pm$ 1.2 & 19.6 $\pm$ 7.1 & -0.9 $\pm$ 4.1 & -18.7 $\pm$ 6.6 \\
  &   & Attention-on-Mov & 76.0 $\pm$ 1.0 & \boldmath{}\textbf{58.7 $\pm$ 1.8}\unboldmath{} & \boldmath{}\textbf{69.0 $\pm$ 1.2}\unboldmath{} & 18.5 $\pm$ 7.4 & 10.3 $\pm$ 4.9 & -28.8 $\pm$ 6.4 \\
  &   & Attention-on-Car & 75.4 $\pm$ 1.0 & 58.0 $\pm$ 1.7 & 68.6 $\pm$ 1.2 & 1.5 $\pm$ 7.5 & 12.9 $\pm$ 4.8 & -14.4 $\pm$ 6.5 \\
  &   & Bilinear & \boldmath{}\textbf{76.4 $\pm$ 0.9}\unboldmath{} & 58.7 $\pm$ 1.8 & 68.8 $\pm$ 1.2 & 37.8 $\pm$ 7.1 & -0.8 $\pm$ 4.1 & -36.9 $\pm$ 6.5 \\
\bottomrule
\end{tabular}%

    }
  \label{tab:mesa_hrv_win_20}%
\end{table}%

\label{apx:results_of_21_51}

\end{document}